\documentclass[10pt,twocolumn,letterpaper]{article}

\usepackage{iccv}
\usepackage{times}
\usepackage{epsfig}
\usepackage{graphicx}
\usepackage{amsmath}
\usepackage{amssymb}

\usepackage{graphicx}
\usepackage{amsmath}
\usepackage{amssymb}
\usepackage{booktabs}

\usepackage{times}
\usepackage{epsfig}

\usepackage{multirow}
\usepackage[table,xcdraw]{xcolor}
\usepackage{algorithmicx,algorithm}
\usepackage{color}
\usepackage{paralist}
\usepackage{arydshln}
\usepackage{multirow}

\usepackage[pagebackref=true,breaklinks=true,letterpaper=true,colorlinks,bookmarks=false]{hyperref}

\iccvfinalcopy % *** Uncomment this line for the final submission

\def\iccvPaperID{3793} % *** Enter the ICCV Paper ID here

\ificcvfinal\fi

\begin{document}

%%%%%%%%% TITLE
\title{Super-resolving Real-world Image Illumination Enhancement: \\
A New Dataset and A Conditional Diffusion Model}
%Using A Conditional Diffusion Model}
% \author{Yang Liu\footnote{Equal contribution}\\
% National University of Defense Technology\\
% {\tt\small lewis.yangliu@gmail.com}
% \and
% Yaofang Liu\footnote{Equal contribution}\\
% City University of Hong Kong\\
% {\tt\small yaofanliu2-c@my.cityu.edu.hk}
% \and
% Jinshan Pan\\
% Nanjing University of Science and Technology\\
% {\tt\small sdluran@gmail.com}
% \and
% Yuxiang Hui\\
% The Chinese University of Hong Kong\\
% {\tt\small 1155152121@link.cuhk.edu.hk}
% \and
% Fan Jia\\
% The Chinese University of Hong Kong\\
% {\tt\small jiafan@math.cuhk.edu.hk}
% \and
% Raymond H. Chan\\
% Lingnan University\\
% {\tt\small raymond.chan@ln.edu.hk}
% \and
% Tieyong Zeng\\
% The Chinese University of Hong Kong\\
% {\tt\small zeng@math.cuhk.edu.hk}
% }
\newcommand{\inst}[1]{\textsuperscript{#1}}
\author{Yang Liu\inst{1,} \thanks{Equal Contribution.}, Yaofang Liu\inst{2,}\footnotemark[1], Jinshan Pan\inst{3},  Yuxiang Hui\inst{1}, Fan Jia\inst{1}, Raymond H. Chan\inst{4}, Tieyong Zeng\inst{1} \\
\inst{1} The Chinese University of Hong Kong,
\inst{2} City University of Hong Kong\\
\inst{3} Nanjing University of Science and Technology,
\inst{4} Lingnan University
}

\maketitle
\ificcvfinal\thispagestyle{empty}\fi

%%%%%%%%% ABSTRACT
%%%%%%%%% ABSTRACT
\begin{abstract}
%
%The last decade has seen a vast quantity of well-lighted super-resolution datasets and algorithms to boost the computational imaging in daylight. 
%
Most existing super-resolution methods and datasets have been developed to improve the image quality in well-lighted conditions. However, these methods do not work well in real-world low-light conditions as the images captured in such conditions lose most important information and contain significant unknown noises.
%
%However, the assumed degradation models deviate from those in real-world low-light images where the low-contrast structural details are hardly distinguishable from seriously disturbing noises.
%
To solve this problem, we propose a SRRIIE dataset with an efficient conditional diffusion probabilistic models-based method. 
%
%In this paper, we propose a large scale SRRIIE dataset for super-resolving real-world illumination enhancement tasks. This dataset consisting of 15,000 paired low-high quality image to enable the training process of deep neural networks from Raw data space to sRGB color space.
%
The proposed dataset contains 4800 paired low-high quality images. To ensure that the dataset are able to model the real-world image degradation in low-illumination environments, we capture images using an ILDC camera and an optical zoom lens with exposure levels ranging from -6 EV to 0 EV and ISO levels ranging from 50 to 12800. 
%
%These images are captured using an ILDC camera and an optical zoom lens with exposure levels ranging from -6 EV to 0 EV and ISO values ranging from 0 to 12800.
%
%To benchmark the performance of existing SOTA methods, we comprehensively evaluate with various reconstruction and perceptual metrics and demonstrate the practicabilities of the SRRIIE dataset for deep neural solvers.
%
We comprehensively evaluate with various reconstruction and perceptual metrics and demonstrate the practicabilities of the SRRIIE dataset for deep learning-based methods.
%
%We reveal the limitations of existing methods in preserving the structures and sharpness from complicated noises. And we provide a simple and effective algorithm by incorporating conditional diffusion models for real Raw sensor data in an unfolding framework.
%
We show that most existing methods are less effective in preserving the structures and sharpness of restored images from complicated noises. To overcome this problem, we revise the condition for Raw sensor data and propose a novel time-melding condition for diffusion probabilistic model. 
Comprehensive quantitative and qualitative experimental results on the real-world benchmark datasets demonstrate the feasibility and effectivenesses of the proposed conditional diffusion probabilistic model on Raw sensor data. Code and dataset will be available at \url{https://github.com/Yaofang-Liu/Super-Resolving}.
\end{abstract}

%%%%%%%%% BODY TEXT
\vspace{-4mm}
\section{Introduction}
\label{sec: introduction}
\vspace{-1mm}

\begin{figure}[!t]\footnotesize
\centering
%\begin{center}
\begin{tabular}{ccc}
\multicolumn{2}{c}{\multirow{3}*[69.45pt]{\hspace{-1.4mm}\includegraphics[width=0.651\linewidth, height = 0.6915\linewidth]{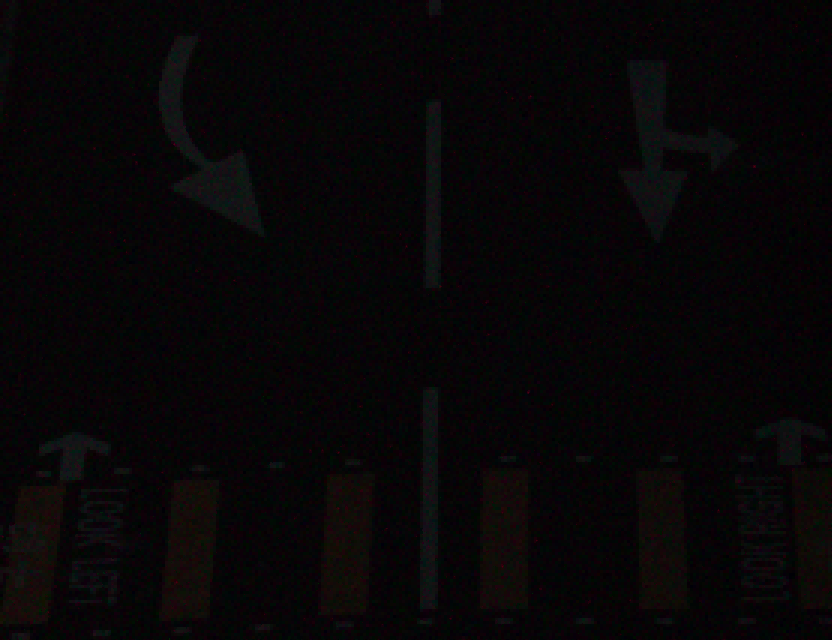}}} &\hspace{-3.3mm}\includegraphics[width=0.32\linewidth, height = 0.32\linewidth]{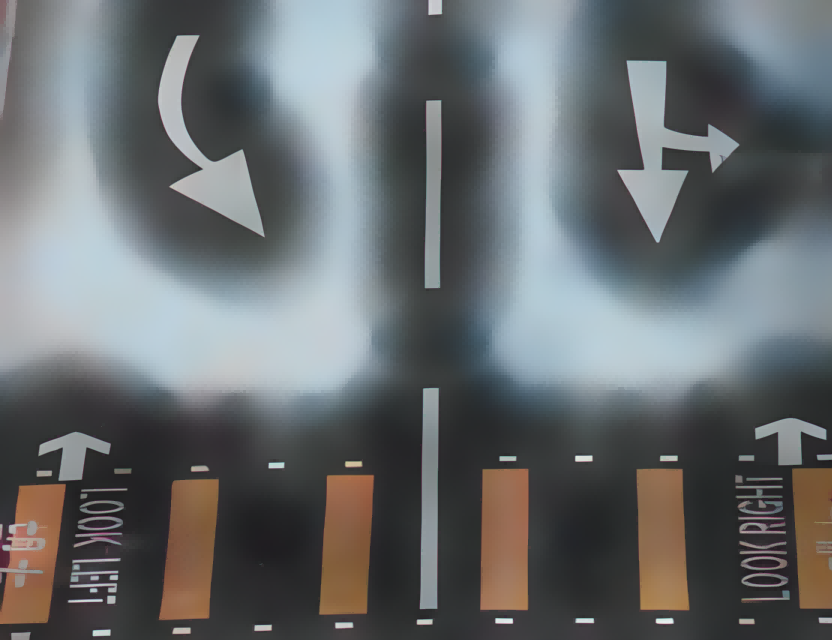}  \\
\multicolumn{2}{c}{}                   &\hspace{-4mm} (b) Real-ESRGAN \cite{wang2021realesrgan}  \\
\multicolumn{2}{c}{}                   &\hspace{-4.0mm} \includegraphics[width=0.32\linewidth, height = 0.32\linewidth]{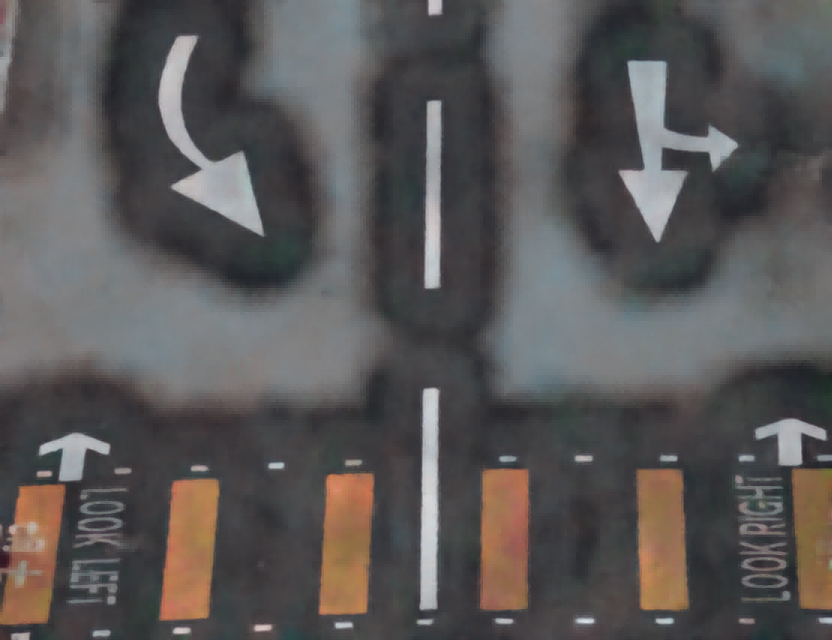} 
\\
\multicolumn{2}{c}{\hspace{-1.4mm}(a) Real-world scene (-5 EV, ISO 4000)}                  &\hspace{-4.0mm} (c) SwinIR \cite{liang2021swinir}  
\\
\hspace{-2.1mm}
\includegraphics[width=0.32\linewidth, height = 0.32\linewidth]{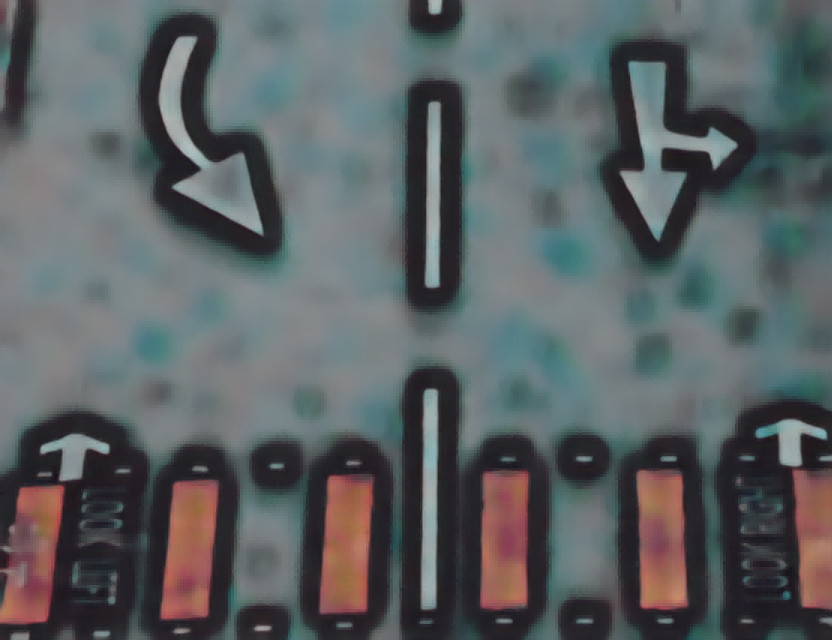}                  &\hspace{-4.0mm} \includegraphics[width=0.32\linewidth, height = 0.32\linewidth]{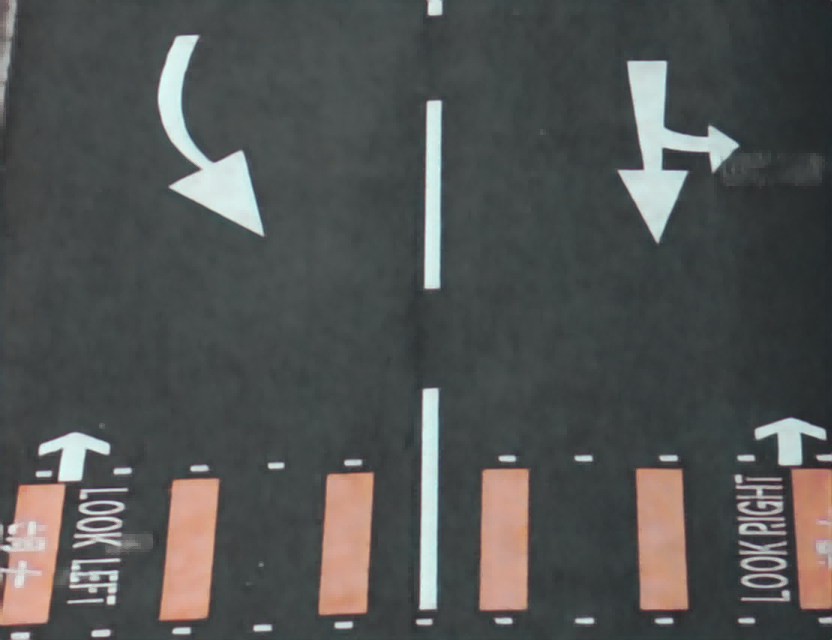}                 &\hspace{-4.0mm} \includegraphics[width=0.32\linewidth, height = 0.32\linewidth]{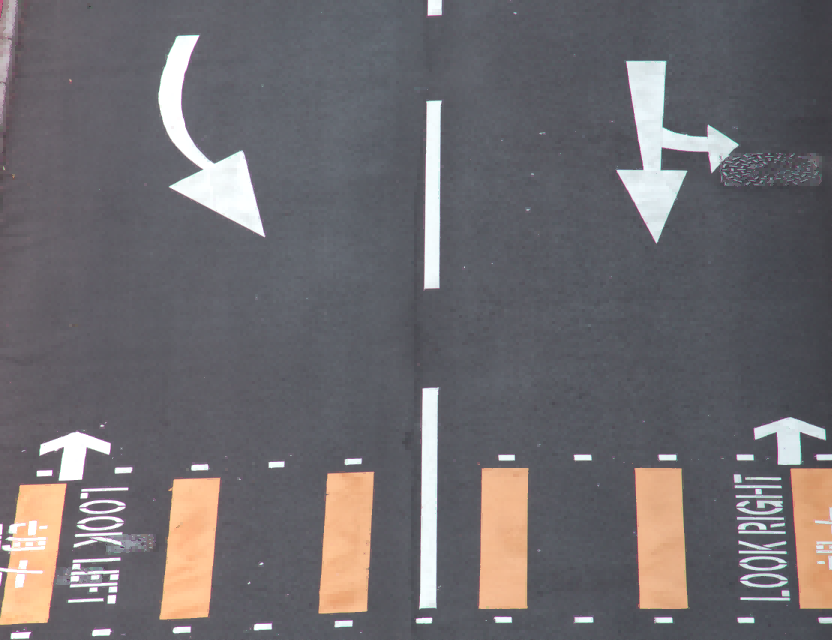}  
\\
\hspace{-2.1mm}
(d) USRNet \cite{zhang2020deep}                 &\hspace{-4.0mm} (e) Ours-Raw                &\hspace{-4.0mm} (f) Ground truth 
\end{tabular}
\vspace{2mm}
\caption{Visual comparisons ($\times 4$ SR) on the SRRIIE dataset for super-resolving real-world image illumination enhancement tasks. The low-quality image is captured with the camera settings of -5 EV and ISO 4000. The complicated noises make it difficult to restore truthful image information. Our conditional DPM-based method using Raw sensor data generates better structural details.}%
\label{fig: preface}
\vspace{-4mm}
\end{figure}

Super-resolving real-world illumination enhancement images captured in dark scenes has recently attracted increasing attention due to its potential applications in practical computational imaging, such as traffic detection and night photography. The objective of this task is to restore rich structural details and enhance unbiased colors as much as possible from the degraded images. A real-world image captured with camera settings of ISO 4000 and -5 EV is shown in Figure \ref{fig: preface} (a). In real-world low-light photography, it is necessary to increase the ISO level to capture with a safe shutter speed to avoid obtaining blurry images. However, high ISO levels will introduce severe noise in images. These real-world degradation processes make the joint image illumination enhancement and super-resolution problems highly ill-posed.
%
%this is a  problem as the information in degraded images is significantly less compared to normal-light high-resolution images. In this paper, we focus on super-resolving real-world shot-exposure clear images in the dark.

To restore images suffering from joint degradation problems, we could employ existing single image illumination enhancement methods \cite{liu2021retinex, wang2021low, Zamir2022MIRNetv2, 9369102} and single image super-resolution methods \cite{zhang2020deep, wang2021realesrgan, liang2021swinir, Zamir2022MIRNetv2} separately in a cascaded manner. 
%However, widely used image illumination datasets, such as LOL \cite{Chen2018Retinex}, SID \cite{chen2018learning}, and SICE \cite{Cai2018deep} datasets, only contain low-normal light images and are not sufficient for real-world high-resolution image restoration tasks. Furthermore, as discussed in Section \ref{ssec: analysis on sequential methods and joint methods}, this cascaded strategy does not always outperform the joint methods due to 
%
However, the intermediate output will suffer from significant information loss, e.g., the over-smoothing effects \cite{ledig2017photo} if we use a CNN for the first stage.
Moreover, the accumulation errors from the first stage will be further amplified in the second stage.
To address these problems, another approach is to directly fine-tune a super-resolution network \cite{zhang2020deep, wang2021realesrgan, liang2021swinir, Zamir2022MIRNetv2} to learn high-resolution images from low-light images in an end-to-end manner on the low-light datasets. 
However, widely used low-light datasets \cite{Chen2018Retinex, chen2018learning, Cai2018deep} does not contain real-world normal-light high-resolution images as the ground truth.
While the widely used super-resolution datasets \cite{martin2001database, zeyde2010single, bevilacqua2012low, agustsson2017ntire, cai2019toward, chen2019camera, zhang2019zoom, wei2020component} does not contain real-world degraded low-light images.
%
%Although these models perform well on well-lighted super-resolution datasets \cite{martin2001database, zeyde2010single, bevilacqua2012low, agustsson2017ntire}, they are less effective in real-world. To address the domain gap of synthetic datasets, several real-world datasets, such as RealSR \cite{cai2019toward}, City100 \cite{chen2019camera}, SR-RAW \cite{zhang2019zoom}, DRealSR \cite{wei2020component}, and 
%
To solve this problem, the RELLISUR \cite{aakerberg2021rellisur} datasets have been proposed using a DSLR cameras to capture paired sRGB data. However, the RELLISUR dataset neglects the noise model and is captured with low ISO levels from 100 to 400. We note that the low ISO levels and less noises simplify the real-world degradation model and will ease the learning process from low-quality to high-quality images.
%Ji et al. \cite{Ji_2020_CVPR_Workshops} introduce a noise injection module to deal with real-world super-resolution, and Zhang et al. \cite{zhang2021designing} propose a practical degradation model. However, the distributions of synthetic noises in daylight scenes in these datasets and methods are significantly different from those of complicated real-world noises in the dark. The simplified degradation model eases the learning process from low-quality to high-quality images; therefore, the designed methods still perform less effectively on real-world datasets due to the deviation from existing assumed SR models, as shown in Figure \ref{fig: preface} (b-d).

%

To address these problems, we introduce a novel image dataset called SRRIIE for super-resolving real-world image illumination enhancement tasks. The SRRIIE dataset poses a challenging task for image super-resolution due to the low light caused by a wide range of under-exposure levels from -6 EV to 0 EV, and real-world noises caused by large ISO levels rom 50 to 12800.
Due to the severe degradation in real-world imaging, existing methods performs less effectively to restore images that lose most important information and contain significant unknown noises.
%
%To overcome the limitations of ResNet-based and Transformer-based methods, which suffer from over-smoothing effects, and GAN-based methods, which introduce untruthful artifacts, 
%
We then propose a new conditional diffusion model that progressively generates truthful structural details from severely degraded Raw images. We demonstrate that Raw sensor data contains more lossless information than sRGB images, and we revise the condition for Raw sensor data to adapted to DPM. To improves the consistency and robustness of the reverse generation process and boost the restoration and enhancement results, we propose a novel time-melding condition by fusing temporal coherence information across relevant time points. We conduct comprehensive benchmark experiments on our SRRIIE dataset. The results show that our proposed method (Figure \ref{fig: preface} (e)) achieves promising results against existing methods (Figures \ref{fig: preface} (b)-(d)).
	
The main contributions of this work are as follows:
	%\vspace{1mm}
	\begin{compactitem}
		\item We have analyzed the real-world noises introduced by high ISO levels in real-world imaging and propose a large-scale SRRIIE dataset. This dataset comprises paired low-high quality image sequences and aims to facilitate the super-resolution of real-world illumination enhancement from Raw sensor data.
		\item We revise the condition for Raw sensor data to adapt to conditional DPM, thus lossless image structural details can be extracted. We further propose a novel time-melding condition implemented by fusing time-consistency information across relevant time points to effectively improve the reverse generation process.
		%conducted a comprehensive benchmark of existing methods on the proposed SRRIIE dataset using various reconstruction and perceptual metrics. Both cascaded and joint processing methods were evaluated to demonstrate the practicality of the dataset for deep neural solvers.
		%
		\item We present a new conditional diffusion model with the proposed two effective conditions for Raw sensor data, which progressively generates sharp image structural details from complex noises. Through both quantitative and qualitative assessments on our real-world SRRIIE datasets, we demonstrate the feasibility and effectiveness of our proposed dataset and method.
	\end{compactitem}

\vspace{-1mm}
\section{Related Work}
\label{sec: related work}
\vspace{-1mm}

We recently have witnessed significant advances in real-world benchmark datasets and  methods for single image illumination enhancement and/or super-resolution tasks. A comprehensive review is beyond the scope of this work. We present the most related ones briefly within proper contexts.

\vspace{-2mm}
{\flushleft \bf {Image illumination enhancement datasets and methods.}}
%Prevailing
Several unpaired low-light image datasets \cite{lee2013contrast, ma2015perceptual, guo2016lime} without ground truth normal-light images are used by traditional methods \cite{stark2000adaptive, lee2013contrast, coltuc2006exact, 2007brightness, guo2016lime, li2018structure, ren2020lr3m}.
However, these histogram-based and variational-based methods cannot be adequate to images with various lighting conditions.
%and thus do not apply to extensive scaled data. 
Wei et al. propose the paired LOL dataset \cite{Chen2018Retinex}. SID \cite{chen2018learning} introduces a Raw dataset containing short-long exposure night-time images. The SICE dataset \cite{Cai2018deep} contains multi-exposed image sequences and the reference images are synthesized by using the exposure fusion methods. Based on these large-scale datasets, EnlightenGAN \cite{EG} performs restoration via a generative network. Zero-DCE \cite{9369102} iteratively enhances a given input image by estimating a set of best-fitting light-enhancement curves in an unsupervised manner. However, these unsupervised learning methods encounter artifacts and color distortion problems. Most recent methods \cite{Chen2018Retinex, liu2021retinex, zhang2019kindling, Yang_2020_CVPR, wang2021low} could achieve favorable results w.r.t. reconstruction metrics, but they are less effective on real Raw low-light images with other degradation processes such as real-world noises or low resolution. 

\vspace{-2mm}
{\flushleft \bf {Image super-resolution datasets and methods.}}
The widely used datasets \cite{martin2001database, zeyde2010single, bevilacqua2012low, agustsson2017ntire} facilitate the training of deep image super-resolution methods.
Seminal ResNet-based methods \cite{kim2016accurate, tai2017image, lim2017enhanced, zhang2018residual, zhang2020deep} using these datasets and demonstrate the effectivenesses against traditional non-deep learning methods \cite{yang2010image, timofte2014a+}.
However, these pioneered works assume that the SR kernels in the degradation process are known and are fixed to the bicubic one, where the low-resolution images are downsample from the HR one with bicubic interpolation. 
To overcome the over-smoothing problem, GAN-based methods \cite{ledig2017photo, wang2018esrgan, wang2021real, zhang2021designing} present considerable potentials in generating rich details. However, unpleasant artifacts and untruthful structures are usually introduced.
%and lower the generated image quality.
%
In \cite{lugmayr2020srflow}, normalizing flow is proposed to generate the distributions of realistic structural details for high-resolution images.
Transformer-based methods \cite{chen2021pre, liang2021swinir, Fang_2022_CVPR} achieve remarkable performance using the attention mechanism.
To fix the domain gap between synthetic datasets and real-world datasets, both low-resolution and high-resolution images are captured with DSLR and smartphone cameras in recently proposed datasets including RealSR \cite{cai2019toward}, City100 \cite{chen2019camera}, SR-RAW \cite{zhang2019zoom}, and DRealSR \cite{wei2020component} datasets.
The most related dataset to ours is the RELLISUR \cite{aakerberg2021rellisur} dataset.
However, these datasets are captured in well-lighted condition with low ISO levels, e.g., 100-400. They neglect the noise model and rarely explore how real-world noises affect the design of deep image super-resolution solvers.
%
%
%We revealed in this works, . 
%Hence it is worth designing practical SR algorithms remains further investigating.
%
 
%The most related dataset to ours is the RELLISUR \cite{aakerberg2021rellisur} dataset which collects a joint low-light image super-resolution dataset on RGB color space.
%%
%However, the RELLISUR \cite{aakerberg2021rellisur} dataset neglects the noise model and is captured in well-lighted conditions with low ISO values, e.g., 100-400. Although they present the noises via histogram stretching, we note that it is usually necessary to increase the ISO value in real-world imaging otherwise we will obtain blurry images.
%%
%In contrast, we explore how real-world raw sensor noises will affect the design of deep image super-resolution solvers, of which the ISO values are ranging from 50 to 12800.

\vspace{-2mm}
{\flushleft \bf {Diffusion probabilistic models for image restoration.}}
Diffusion probabilistic models \cite{ho2020denoising,sohl2015deep,song2019generative,song2020score, dhariwal2021diffusion} forming a novel family of deep generative models \cite{yang2022diffusion} have showed state-of-the-art performance in image restoration \cite{kawar2022denoising,chung2022come,rombach2022high,saharia2022image}. 
These pioneered works reveal that the diffusion models have strong capability in generating high-fidelity and perceptually pleasing structures and details \cite{croitoru2022diffusion}. 
Existing methods \cite{ho2020denoising, kawar2022denoising} commonly adopt the Gaussian noises for the diffusion process.
Cold diffusion \cite{bansal2022cold} replaces the Gaussian noises with various degradation models and shows remarkable restoration results across multiple tasks. 
Conditional diffusion models achieves noticeable results in face image super-resolution \cite{li2022srdiff, saharia2022image, rombach2022high}.
%
%As the data distributions of face images are relatively simpler, it worth further investigating for natural image super-resolution.
%
Whang et al. \cite{whang2022deblurring} train a stochastic sampler for blind deblurring based on conditional diffusion models.
PatchDM \cite{ozdenizci2023} is proposed to restore hazy and rainy images.
%
%However, these works are mainly designed for simplified degradation models that eases the learning process. 
%
However, these methods rarely explore how real-world noises affect the design and performance of diffusion probabilistic models and how to accommodate to the real Raw sensor data.

Different from existing datasets and methods, our collected SRRIIE dataset captures real Raw sensor data (Section \ref{sec: SRRIIE dataset}). The low-resolution images suffer from complicated real-world noises. To solve this problem, we propose an effective conditional DPM-based method with two novel conditions (Section \ref{sec: proposed algorithm}) to progressively generate photo-realistic structural details from lossless Raw images.

	\begin{figure*}[!t]\footnotesize
\centering
	\vspace{-4mm}
	\begin{center}
		\begin{tabular}{c}
			\hspace{-2.6mm}
			\includegraphics[width = 1.0\linewidth]{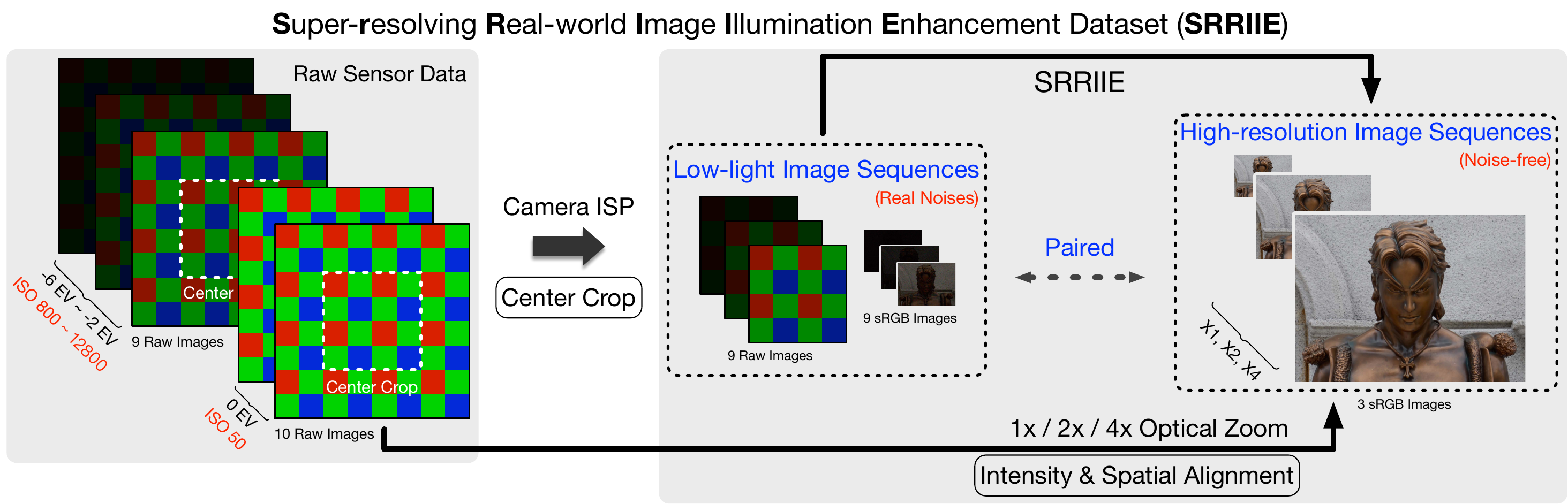}\\
		\end{tabular}
	\end{center}
	\vspace{-3mm}
	\caption{Our data collection and processing pipeline for one scene.
	%An overview of an image sequence for one scene in our SRRIIE dataset. 
			%
			%We capture paired raw image sequences of low-normal light in different scenes with the exposure value ranging from -6 EV to 0 EV and an optical zoom ratio of X1, X2 and X4.
			%
			An ILDC with an optical zoom lens is used to collect real Raw sensor data with a wide range of EV and ISO levels.
			Camera ISP is employed to convert Raw images to sRGB images and only the centered croppings are retained.
			Each low-light image sequence contain 9 low-light Raw/sRGB images with complicated real-world noises and low intensities.
			While each high-resolution image sequence contains 3 noise-free normal-light high-resolution images as the ground truth. 
			These paired images model the real-world degradation process for for super-resolving image illumination enhancement (SRRIIE) tasks.   
			%image sequence is captured per scene with bracketing exposure values ranging from -6 EV to 0 EV and 3 optical zoom levels of X1, X2, and X4. Different ISO values ranging from 50 to 12800 are used to account for real-world noise levels per scene.
			%
			%An IE and an SR processing branch are employed to process these real-world data to accommodate the joint illumination enhancement and super-resolution tasks.
			%
			%Finally, we obtain the low-quality images in raw data space and corresponding ground truth images in RGB color space.
			%the main steps in our procedure for ground truth image estimation. The respective sections for each step are shown.
		}
	\label{fig: data-framework}
	\vspace{-4mm}
\end{figure*}

%\begin{figure}[!t]\scriptsize
%	\centering
%	\vspace{0mm}
%	\vspace{0mm}
%	\begin{tabular}{cccccc}
%	\hspace{-2.6mm}
%	\includegraphics[width=0.33\linewidth]{figures/section3/ISO/1_x1_10.pdf} &\hspace{-4.5mm}
%	\includegraphics[width=0.33\linewidth]{figures/section3/ISO/1_x1_10_G_D_copy.png} &\hspace{-4.5mm}
%	\includegraphics[width=0.33\linewidth]{figures/section3/ISO/2_x1_1_copy.png} 
%	\\
%	\hspace{-2.6mm}
%	(a) Noise-free image &\hspace{-4.5mm} (b) Gaussian-Poisson noise &\hspace{-4.5mm} (c) ISO 100 in \cite{aakerberg2021rellisur}
%	\\
%	\hspace{-2.6mm}
%	\includegraphics[width=0.33\linewidth]{figures/section3/ISO/2_x1_2_copy.png} &\hspace{-4.5mm}
%	\includegraphics[width=0.33\linewidth]{figures/section3/ISO/2_x1_6_copy.png} &\hspace{-4.5mm}
%	\includegraphics[width=0.33\linewidth, height=0.3\linewidth]{figures/section3/ISO/ISO_noise.pdf} 
%	\\
%	\hspace{-2.6mm}
%	(d) ISO 800 &\hspace{-4.5mm} (e) ISO 3200 &\hspace{-4.5mm} (f) ISO 12800
%		\end{tabular}
%		%\end{center}
%	\vspace{2mm}
%	\caption{Visual distributions of different synthetic and real-world noisy images captured with varying ISO values. 
%	%The standard deviation $\sigma$ is estimated using the non-reference heteroscedastic Gaussian model \cite{hasinoff2014photon}.
%	The standard deviation $\sigma$ is measured using \cite{hasinoff2014photon}. (b-f) are cropped from (a).
%	}
%	\vspace{-4mm}
%	\label{fig: ISO-noise}
%\end{figure}

\begin{figure}[!t]\scriptsize
	\centering
	\vspace{0mm}
	\begin{tabular}{cccccc}
	\hspace{-2.1mm}
	\includegraphics[width=0.48\linewidth, height=0.37\linewidth]{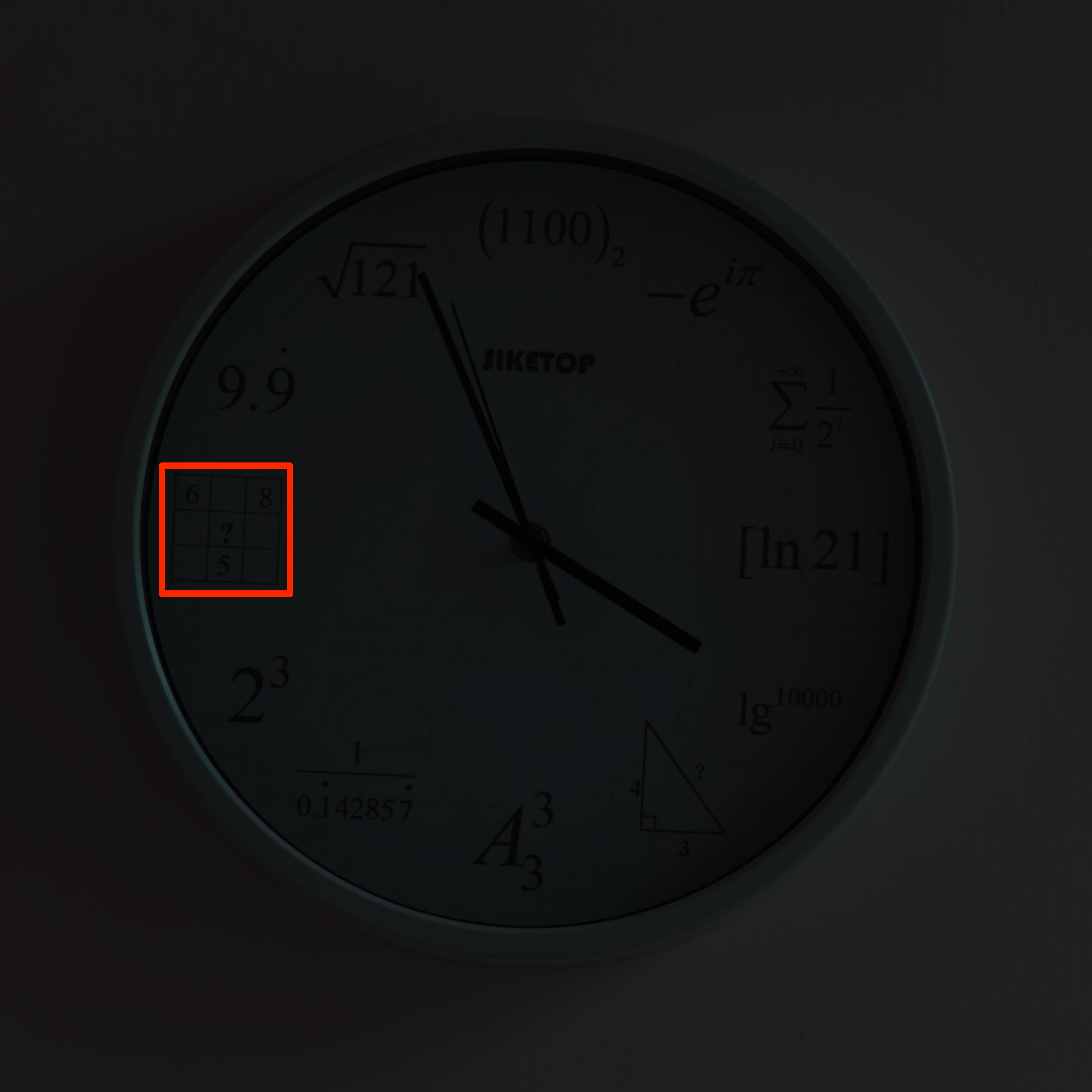} &\hspace{-4mm}
	\includegraphics[width=0.48\linewidth, height=0.37\linewidth]{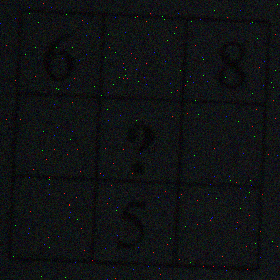}
	\\
	\hspace{-2.1mm}
	(a) Noise-free image &\hspace{-4mm} (b) Gaussian-Poisson noises
	\\
	\hspace{-2.1mm}
	\includegraphics[width=0.48\linewidth, height=0.37\linewidth]{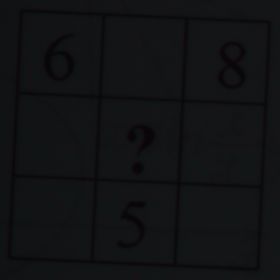} &\hspace{-4mm}
	\includegraphics[width=0.48\linewidth, height=0.37\linewidth]{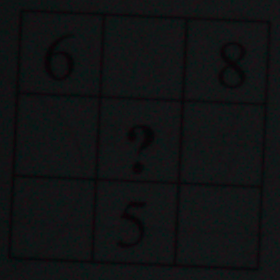}
	\\
	\hspace{-2.1mm}  
	 (c) ISO 100 in \cite{aakerberg2021rellisur} &\hspace{-4mm} (d) ISO 800
	\\
	\hspace{-2.1mm}
	\includegraphics[width=0.48\linewidth, height=0.37\linewidth]{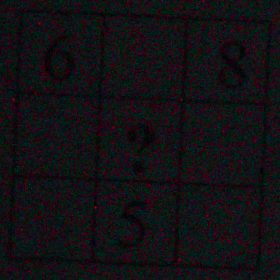} &\hspace{-4mm}
	\includegraphics[width=0.48\linewidth]{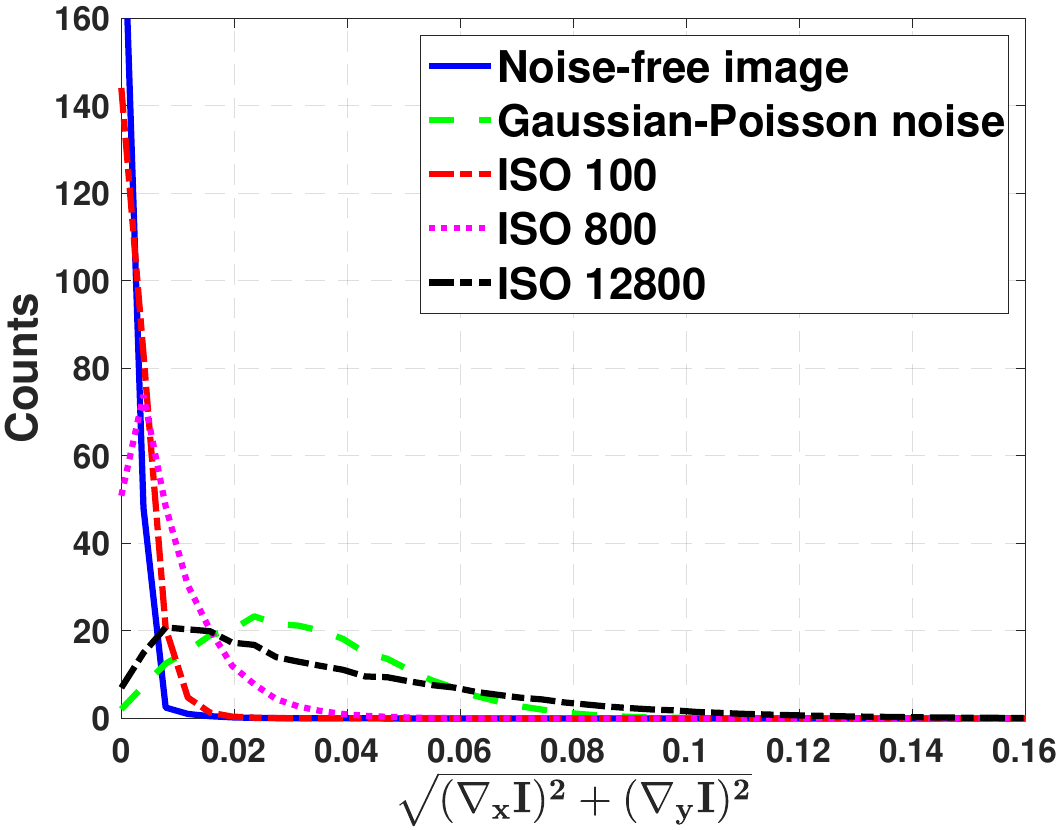} 
	\\
	\hspace{-2.1mm}
	 (e) ISO 12800 &\hspace{-4mm} (f) Image gradient distributions
		\end{tabular}
		%\end{center}
	\vspace{0mm}
	\caption{Simple cases to illustrate the motivation. (b)-(e) are the corresponding cropped and zoomed-in regions as denoted by the red box in (a). The synthetic Gaussian-Poisson noisy image (b) or the image captured with the low ISO level (c) deviates largely from real-world noisy images (d) - (e) in terms of visual comparison and statistical image gradient distributions (f).
	%Visual comparisons of different synthetic and real-world noisy images captured with varying ISO values. 
	%The standard deviation $\sigma$ is estimated using the non-reference heteroscedastic Gaussian model \cite{hasinoff2014photon}.
	%The standard deviation $\sigma$ is measured using \cite{hasinoff2014photon}. (b-f) are cropped from (a).
	%
	%We crop and zoom-in the corresponding  for Figures \ref{fig: ISO-noise} .
	}
	\vspace{-5mm}
	\label{fig: ISO-noise}
\end{figure}

\vspace{-1mm}
\section{SRRIIE Dataset}
\label{sec: SRRIIE dataset}
\vspace{-1mm}

% joint task, real-world, is highly ill-posed task, difficult
% sota methods performance not well in RGB, RGB is insufficient, to explore the upper bound

To model the real-world degradation process for Super-resolving image illumination enhancement tasks, we collect the real-world SRRIIE dataset that contains paired low-light images (Raw/sRGB) and high-resolution images (sRGB) to enables the learning process for deep neural networks.
%
%with paired low-quality input and high-quality ground truth image sequences for training and test.
%
%
%In an end-to-end manner to jointly perform real-world demosaicing, denoising, illumination enhancement, and super-resolution tasks. 
%A data pipeline is an end-to-end sequence of digital processes used to collect, modify, and deliver data.
Our data pipeline explores the complicated real-world degradation processes including real-world noises, low light and low resolution, which differentiate our data and method framework from existing data and methods.

\vspace{-1mm}
\subsection{Data collection and processing pipeline}
\label{ssec: Data pipeline}
\vspace{-1mm}

We use a Sony A7 IV camera equipped with a Tamron 28-200mm zoom lens to collect 400 paired image sequences from 400 different scenes for super-resolving real-world illumination enhancement tasks.
Figure \ref{fig: data-framework} illustrates the main collection and processing steps of our proposed SRRIIE dataset for one scene, where the Raw sensor data is collected to better preserve the real-world degradation process. We employ the camera ISP to convert Raw sensor data to corresponding sRGB images and retain the centered croppings.
Each image sequence contains paired low-light images suffering from real-world noises under varying ISO levels, and and the corresponding noise-free normal-light high-resolution images.
We illustrate the details as follows.
%The low-light image sequences  We employ an IE processing branch and a SR processing branch as follows to process each image sequence.

\vspace{-2mm}
{\flushleft \bf {Low-light image sequences.}}
We capture 9 Raw images for each low-light image sequence using the auto bracketing exposure mode of the camera in an under-exposure manner ranging from -6 EV to -2 EV.
The only variable camera parameter to achieve the real-world low-light sequences for each scene is the exposure time and the others are fixed. 
%We fix the other camera parameters , e.g., aperture, focus distance, white balance, etc. 
%
%Example results are presented in Figure \ref{fig: ISO-noise} (d-f).
%
We use a fixed ISO level for each scene, while the ISO level varies from 800 to 12800 according to different scene conditions. 
We note that the camera ISO parameter is one of the key settings in real-world low-light imaging to capture properly exposed image in poor lighting conditions. It is usually necessary to increase the ISO level to ensure the safe shutter speed otherwise we will capture a blurry image.
However, as higher ISO levels will inevitably degrade the image quality, the image structures and details are corrupted by complicated noises.
We note that a wide range of ISO levels with complicated real-world noises for super-resolving real-world illumination enhancement on real sensor data makes the restoration problems high ill-posed, which differentiates our data pipeline from existing datasets \cite{Chen2018Retinex, chen2018learning, zhang2019zoom, aakerberg2021rellisur}.
%that usually employ limited ISO settings

\vspace{-2.0mm}
{\flushleft \bf {High-resolution image sequences.}}
The high-resolution image sequences are achieved by 1x, 2x and 4x optical zoom using 0 EV as the size of an imaging object is approximately linear to the focal length of the camera..
%
%We repeat the above image acquisition process at the optical zoom ratio of X1, X2, and X4 to mimic the magnifications of image resolution since 
% to realized the X1, X2 and X4 super-resolution tasks.
%
For example, we use a focal length of 30 mm, 60 mm, and 120 mm for indoor scenes, and a focal length of 50 mm, 100 mm, and 200 mm for outdoor scenes, which respectively correspond to $\times$1, $\times$2, and $\times$4 image super-resolution tasks.
We capture 10 successive Raw images $\{I_n^m\}_{n=1}^{10}$ for each optical zoom level using a continuous capture mode of the camera and robustly estimate the mean image to obtain the high-quality noise-free ground truth images $\{J_m\}_{m=1}^3$: 
\vspace{-1.mm}
\begin{equation}
%\vspace{-1.mm}
	\label{eq: noise-free estimate}
	J_m = \Psi_{mean} (g_{crop}(f_{ISP} (\{I_n^m\}_{n=1}^{10}))), ~~ m = 1,2,3,
	\vspace{-1.mm}
\end{equation}
where $m$ denotes three optical zoom levels. $f_{ISP}$, $g_{crop}$ and $\Psi_{mean}$ is the camera ISP, center crop operator and robust mean image estimation method \cite{abdelhamed2018high}, respectively. 
We then perform intensity and spatial alignment to obtain the rectified ground truth images $\{\bar{J}_m\}_{m=1}^3$: 
\vspace{-1.mm}
\begin{equation}
%\vspace{-2.mm}
	\label{eq: intensity spatial align}
	\bar{J}_m = \Phi_{align}(J_m - \mu_m + \mu_1), ~~m = 1,2,3,
	\vspace{-1.mm}
\end{equation}
where $\{\mu_m\}_{m=1}^3$ are corresponding mean intensities of $\{J_m\}_{m=1}^{3}$. $\Phi_{align}$ is a spatial alignment function based on SIFT descriptors \cite{lowe2004distinctive} to deal with the distortions caused by optical zoom.
The only variable camera parameter to achieve the real-world high-resolution sequences for each scene is the focal length, and the others are fixed.

%

%

%
%We experimentally find operating on raw images achieves lower noise levels compared to that on RGB images measured using \cite{hasinoff2014photon} as presented in Figure \ref{fig: image averaging-noise}.
%
%Different from performing standard camera ISP on each image individually, we take an average operator for these 10 images in raw data space after demosaicing to merge 10 raw images into one ground truth noise-free image. 
%
%Figure \ref{fig: image averaging-noise} shows a scene to illustrate our ground truth estimation.
%
%Since noise is a random process, each image contains a random sample from the sensor’s noise distribution.
%
%
%Therefore, we obtain 9 low-light images and 1 normal-light images at a fixed optical zoom level.
%
%Therefore, we obtain 3 image sequences with a number of 10 in each optical zoom level for one scene.
%

In total, we capture 400 image sequences which contain 4800 images for 100 indoor scenes and 300 outdoor scenes. 
Each image sequence contains 12 images, i.e., 9 low-light images (Raw/sRGB) and 3 high-resolution images (sRGB).
These images are paired to model the real-world degradation process for Super-resolving image illumination enhancement tasks.
%These paired RAW-RGB images form an image sequence for joint illumination and super-resolution tasks as shown in Figure \ref{fig: data-framework}.
%
%The original image size is 4672 $\times$ 7008 and we use different center crop sizes for each image sequence.
%
%Considering the categories of image contents and data distributions of camera ISO values within 500 scenes,
We selectively split 400 image sequences into 300 and 100 sets for training and test, respectively.

During the data collection process for each scene, we use fixed camera parameters to maximize imaging authenticity, e.g., aperture, white balance, etc.
We use the aperture size ranging from f/8 to f/19 to balance the image quality and depth of field considering the depth of objects.
We mount the camera on a sturdy tripod and use a remote shooting manner. The camera stabilization and noise reduction functions of the camera are turned off. We capture the images by manual focus and central partial metering.
The collected images that suffer from undesired defects including external illumination changes, blurs and unaligned objects have been manually discarded. We also avoid other ethical issues.
%
%We also provide an RGB image dataset for evaluation.

% \usepackage{multirow}
\begin{table}[!t]\small
\centering
\caption{Image quality assessment for different image degradation processes in our SRRIIE dataset. 
%The sub-processes of illumination enhancement and super-resolution are evaluated separately.
}
\vspace{1.5mm}
\label{tab:Degradation Process}
\begin{tabular}{|ll|llll|}
\hline
\multicolumn{2}{|c|}{Degradation Process}                 & \multicolumn{1}{c|}{PSNR $\uparrow$} & \multicolumn{1}{c|}{SSIM $\uparrow$} & \multicolumn{1}{c|}{LPIPS $\downarrow$} \\ 
\hline
\hline
\multicolumn{1}{|l|}{\multirow{3}{*}{Exposure}} & -2.0 EV & \multicolumn{1}{c|}{10.74}                           & \multicolumn{1}{c|}{0.5241}                           & \multicolumn{1}{c|}{0.7543}                            \\
%\multicolumn{1}{|l|}{}                          & \multicolumn{1}{c|}{-2.5 EV} & \multicolumn{1}{c|}{10.17}                            & \multicolumn{1}{c|}{0.4695}                           & \multicolumn{1}{c|}{0.8205}                            \\
%\multicolumn{1}{|l|}{}                          & \multicolumn{1}{c|}{-3.0 EV} & \multicolumn{1}{c|}{9.75}                           & \multicolumn{1}{c|}{0.4247}                           & \multicolumn{1}{c|}{0.8776}                            \\
%\multicolumn{1}{|l|}{}                          & \multicolumn{1}{c|}{-3.5 EV} & \multicolumn{1}{c|}{9.43}                           & \multicolumn{1}{c|}{0.3892}                           & \multicolumn{1}{c|}{0.9232}                            \\
\multicolumn{1}{|l|}{}                          & \multicolumn{1}{c|}{-4.0 EV} & \multicolumn{1}{c|}{9.20}                           & \multicolumn{1}{c|}{0.3612}                           & \multicolumn{1}{c|}{0.9568}                            \\
%\multicolumn{1}{|l|}{}                          & \multicolumn{1}{c|}{-4.5 EV} & \multicolumn{1}{c|}{9.02}                           & \multicolumn{1}{c|}{0.3396}                           & \multicolumn{1}{c|}{0.9798}                            \\
%\multicolumn{1}{|l|}{}                          & \multicolumn{1}{c|}{-5.0 EV} & \multicolumn{1}{c|}{8.88}                           & \multicolumn{1}{c|}{0.3233}                           & \multicolumn{1}{c|}{0.9904}                            \\
%\multicolumn{1}{|l|}{}                          & \multicolumn{1}{c|}{-5.5 EV} & \multicolumn{1}{c|}{8.78}                           & \multicolumn{1}{c|}{0.3104}                           & \multicolumn{1}{c|}{1.0021}                            \\
\multicolumn{1}{|l|}{}                          & \multicolumn{1}{c|}{-6.0 EV} & \multicolumn{1}{c|}{8.72}                           & \multicolumn{1}{c|}{0.3011}                           & \multicolumn{1}{c|}{1.0095}                            \\ 
\hline
\end{tabular}
\vspace{-4mm}
\end{table}

\vspace{-1mm}
\subsection{Degradation analysis on the dataset}
\label{ssec: statistical analysis on the dataset}
\vspace{-1mm}

To analyze the challenge of the highly ill-posed real-world degradation process, we show an indoor example captured with 0 EV in Figure \ref{fig: ISO-noise}. We crop and zoom-in the corresponding regions denoted by the red box in Figure \ref{fig: ISO-noise} (a) for Figures \ref{fig: ISO-noise} (b)-(e).
The noise-free image (Figure \ref{fig: ISO-noise} (a)) is achieved by robustly averaging 10 images \cite{abdelhamed2018high} captured with ISO 50.
%
%preserving the structures and sharpness from complicated noises.
%
The synthetic Gaussian-Poisson noises are widely used to add to the noise-free image in existing methods. However, the synthetic noises deviate significantly from real noises as shown in Figure \ref{fig: ISO-noise} (b) ($\sigma_G = 0.04, \lambda_P = 0.001$). 
Several benchmark datasets such as the RELLISUR \cite{aakerberg2021rellisur} dataset use low ISO levels (Figure \ref{fig: ISO-noise} (c)), but we note that it will easily cause significant blurs without using a tripod due to long exposure time.
%
%and contains less noises as shown in Figure \ref{fig: ISO-noise} (c) shows a scene captured using different ISO values and the accompanying noises.
%
On the contrary, Figures \ref{fig: ISO-noise} (b)-(e) show the real-world noises captured using ISO 800 and 12800 to guarantee the safe shutter speed. As the ISO level increases, the heavier real noises severe degrade the image quality and make this problem highly ill-posed. 
Figure \ref{fig: ISO-noise} (f) further demonstrates that the low-light images differ widely in image gradient distribution.
%
%As a trade-off, the captured images will contain complicated noises discouraging effective restoration and enhancement. Figure \ref{fig: ISO-noise} also shows the noisy examples.
%
%On the other hand, our dataset contains a wide range of under-exposure and low signal deviations as summarized in Table \ref{tab:mu-sigma}.
%
Table \ref{tab:Degradation Process} summarizes the severe information loss compared to the ground truth in terms of PSNR, SSIM and LPIPS metrics due to the superposition of a wide range of under-exposure levels and high ISO levels used in our dataset. The severely degraded image quality makes the inverse problems much more difficult to learn with existing deep solvers.
%
%These differences distinguish our real raw sensor data from existing datasets as well as the RELLISUR \cite{aakerberg2021rellisur} dataset.
%
%The low contrast and small spread characteristics of our proposed real-world dataset make it challenging to learn with existing deep solvers.
%

%
%very challenging for existing methods

%This wide range of under exposure levels can help to improve the generalization abilities of models trained on the RELLISUR \cite{aakerberg2021rellisur} dataset

%obtain a variety of noise levels (the higher the ISO level, the higher the noise).

\begin{figure*}[!tp]\footnotesize
	\vspace{-3mm}
	%\hspace{-4mm}
	\begin{center}
		\begin{tabular}{c}
			\vspace{-0mm}
			\hspace{-3mm}
			\includegraphics[width = 0.99\linewidth]{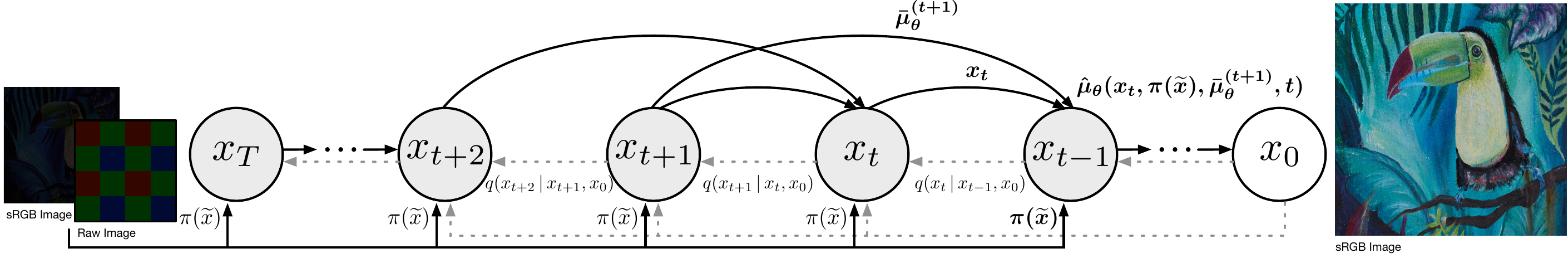}\\
		\end{tabular}
	\end{center}
	\vspace{-2mm}
	\caption{An overview of our conditional DPM with the forward diffusion (dashed line) and reverse process (solid line).
	Our method generates photo-realistic image structural details conditioned on Raw sensor data $\pi{(\widetilde{x})}$ (Eq. (\ref{eq: DPM7})). 
	The proposed time-melding condition (Eq. (\ref{eq: DPM7})) improves the reverse generation process (Eq. (\ref{eq: DPM8})) by capturing temporal coherence and consistency across relevant time points.
		}
	\label{fig: network-framework}
	\vspace{-4mm}
\end{figure*}

\vspace{-1mm}
\section{Revisiting DPM}
\label{sec: Revisit diffusion models}
\vspace{-1mm}
Although we could apply existing methods directly to our SRRIIE dataset, there are still significant challenges due to the highly ill-posed nature of the problem.
%
%as real-world image degradation process suffers from significant information loss. 
%
To address these issues and restore high-quality images from degraded low-quality counterparts, we propose a conditional diffusion model-based method. 
To aid the subsequent presentation of our method, we briefly review the diffusion probabilistic models (DPM).

DPM \cite{sohl2015deep, ho2020denoising} are generative models that learn a Markov Chain to gradually transform a Gaussian noise distribution into the target data distribution. To accelerate the sampling speed, DDIM \cite{songdenoising} introduces a generalized non-Markovian diffusion process which involves the following inference distributions indexed by a real vector $\sigma \in \mathbb{R}_{\geq 0}^T$: %sequentially corrupts the data x0 ∼ q(x0) at T diffusion time steps by injecting Gaussian noise according to a variance schedule β1 , . . . , βT.
\vspace{-1.5mm}
\begin{equation}
	\label{eq: DPM1}
	q_{\sigma}(x_{1:T}\, |\, x_0) := q_{\sigma}(x_T\, |\, x_0) \prod_{t=2}^T q_{\sigma}(x_{t-1}\, |\, x_t, x_0),
	\vspace{-1.5mm}
\end{equation}
where the marginals $q_{\sigma}(x_T\, |\, x_0)$ and $q_{\sigma}(x_{t-1}\, |\, x_t, x_0)$ are Gaussian distributions $\mathcal{N}(x_T; \sqrt{\alpha_T} x_0, (1-\sqrt{\alpha_T})\mathbf{I})$ and $\mathcal{N}(x_{t-1}; \mu_t(x_t,x_0), \sigma_t^2 \mathbf{I})$, respectively.
$\alpha_{1:T}\in (0,T]^T$ is a decreasing sequence which could be manually configured or learned jointly \cite{ho2020denoising}. 
Starting from a standard normal prior $p(x_T)=\mathcal{N}(x_T; \mathbf{0}, \mathbf{I})$, the reverse process involves the following Markov Chain:
\vspace{-1.5mm}
\begin{equation}
	\label{eq: DPM2}
	p_{\theta}(x_{0:T}) := p(x_T) \prod_{t=1}^T p_{\theta}^{(t)}(x_{t-1}\, |\, x_t),
	\vspace{-1.5mm}
\end{equation}
that employs learned Gaussian denoisers $\mu_{\theta}(x_t,t)$ to predict $\mu_t(x_t,x_0)$. The denoisers are trained to perform noise reduction while progressively preserve relevant signal information to restore cleaner images.
To train the denoisers, the variational inference objective \cite{ho2020denoising, dhariwal2021diffusion, songdenoising} is employed to optimize $\theta$ which can be simplified to:
%
%\vspace{-1.mm}
%\begin{equation}
%	\label{eq: non-Markovian_1}
%	q_{\sigma}(x_T\, |\, x_0) = \mathcal{N}(x_T; \sqrt{\alpha_T} x_0, (1-\sqrt{\alpha_T})\mathbf{I}),
%	\vspace{-1.mm}
%\end{equation}
%%
%\vspace{-1mm}
%\begin{equation}
%\label{eq: haze-model-data-term-2}
%q_{\sigma}(x_{t-1}\, |\, x_t, x_0) = \mathcal{N}(x_{t-1}; \mu_t(x_t,x_0), \sigma_t^2 \mathbf{I}),
%\vspace{-1mm}
%\end{equation} 
%q_{\sigma}(x_{t-1}|x_t, x_0) = \mathcal{N}(x_{t-1}; \sqrt{\alpha_{t-1}} x_0 \\
%+ \sqrt{1-\alpha_{t-1}-\sigma_t^2 \cdot \frac{x_t - \sqrt{\alpha_t} x_0}{\sqrt{1 - \alpha_t}}}, \sigma_t^2 \mathbf{I}),
%
%\vspace{-1.5mm}
%\begin{multline}
%	\label{eq: non-Markovian_1}
%	L_{\theta} = \mathbb{E}_{q_{\sigma}} [\underbrace{D_{KL}({q_{\sigma}}(x_T\, |\, x_0)\, ||\, p_{\theta}(x_T))}_{L_T} - \underbrace{\log p_{\theta}(x_0\, |\, x_1)}_{L_0} \\
%	+ \sum_{t=2}^T \underbrace{D_{KL} ({q_{\sigma}}(x_{t-1}\, |\, x_t,x_0)\, ||\, p_{\theta}^{(t)} (x_{t-1}\, |\, x_t))}_{L_{t-1}}],
%	\vspace{-1.5mm}
%\end{multline}
%
\vspace{-1mm}
\begin{equation}
\label{eq: DPM3}
L_{\theta}^{(t-1)} = \mathbb{E}_{x_0, \epsilon_t, t} \, [||\epsilon_t - \epsilon_{\theta}(\sqrt{\alpha_t}x_0 + \sqrt{1-\alpha_t}\epsilon_t, t)||^2],
\vspace{-1mm}
\end{equation} 
The alternative reparameterization of $\mu_{\theta}$ is used by $\epsilon_{\theta}$ with a re-weighted optimization strategy \cite{ho2020denoising, songdenoising}.
%
%\vspace{-1mm}
%\begin{equation}
%\label{eq: haze-model-data-term-2}
%L_{t-1} = \mathbb{E}_{x_0, \epsilon_t, t} \, [||\epsilon_t - \epsilon_{\theta}(\sqrt{\alpha_t}x_0 + \sqrt{1-\alpha_t}\epsilon_t, t)||^2],
%\vspace{-1mm}
%\end{equation} 
%
Formally, the one-step generation process of DDIM from $x_t$ to $x_{t-1}$ using the trained denoisers is performed by:
\vspace{-1.mm}
\begin{equation}
	\label{eq: DPM4}
	x_{t-1} = \sqrt{\alpha_{t-1}} \hat{\mu}_{\theta} (x_t, t) + \hat{r}_{\theta} (x_t, t) + \sigma_t \epsilon_t,
	\vspace{-1.mm}
\end{equation}
where $\hat{\mu}_{\theta} (x_t, t) = (x_t - \sqrt{1-\alpha_t} \cdot \epsilon_{\theta}(x_t, t)) / \sqrt{\alpha_t}$ is the predicted $x_0$, while $\hat{r}_{\theta} (x_t, t) = \sqrt{1-\alpha_{t-1}-\sigma_t^2} \cdot \epsilon_{\theta}(x_t, t)$ is a residual term directed to $x_t$. We set $\sigma_t=0$ to achieve the deterministic implicit sampling.

Conditional diffusion models \cite{dhariwal2021diffusion, rombach2022high, saharia2022photorealistic, ozdenizci2023} aim to generate samples of $x$ that closely match the data distribution conditioned on $\widetilde{x}$, without modifing the underlying diffusion process $q_{\sigma}(x_{1:T}\, |\, x_0)$. The condition $\widetilde{x}$ is usually injected into Eqs. (\ref{eq: DPM2}) and (\ref{eq: DPM3}) and the correspond revised Eq. (\ref{eq: DPM4}) is:
\vspace{-1.mm}
\begin{equation}
	\label{eq: DPM5}
	x_{t-1} = \sqrt{\alpha_{t-1}} \hat{\mu}_{\theta} (x_t, \widetilde{x},t) + \hat{r}_{\theta} (x_t, \widetilde{x}, t) + \sigma_t \epsilon_t,
	\vspace{-1.mm}
\end{equation}
where $x_t$ and $\widetilde{x}$ can be concated \cite{ozdenizci2023} or linearly added \cite{saharia2022photorealistic}.

\vspace{-1mm}
\section{Conditional DPM-Based Method}
\label{sec: proposed algorithm}
\vspace{-1mm}
Based on above analysis, we propose the conditional DPM-based method with two novel conditions. Figure \ref{fig: network-framework} shows an overview of our approach.
%for the progressive generation of photo-realistic image structural details.

\vspace{-1mm}
\subsection{Revised condition for Raw sensor data}
\label{ssec: Conditional diffusion models}
\vspace{-1mm}
The low-light image sequances of our SRRIIE dataset contains both Raw and sRGB images. To bridge the domain gap, we employ an end-to-end U-Net \cite{ronneberger2015u} similar to \cite{chen2018learning} to learn a mapping from Raw images to sRGB images. We modify the last convolutional layer with a integrated gamma correction function to adjust the brightness and contrast of learned deep features.
The reverse process defined in Eqs. (\ref{eq: DPM2}) and (\ref{eq: DPM3}) can then be revised with the condition $\pi{(\widetilde{x})}$. Correspondingly, we update Eq. (\ref{eq: DPM5}) use the revised $\epsilon_{\theta}(x_t, \pi{(\widetilde{x})}, t)$:
%
%\vspace{-1.mm}
%\begin{equation}
%	\label{eq: DPM6}
%	p_{\theta}(x_{0:T}\, |\, \pi{(\widetilde{x})}) = p(x_T) \prod_{t=1}^T p_{\theta}^{(t)}(x_{t-1}\, |\, x_t, \pi{(\widetilde{x})}),
%	\vspace{-1.mm}
%\end{equation}
\vspace{-1.mm}
\begin{equation}
	\label{eq: DPM6}
	x_{t-1} = \sqrt{\alpha_{t-1}} \hat{\mu}_{\theta} (x_t, \pi{(\widetilde{x})},t) + \hat{r}_{\theta} (x_t, \pi{(\widetilde{x})}, t) + \sigma_t \epsilon_t,
	\vspace{-1.mm}
\end{equation}
where $\pi$ represents an identity mapping if $\widetilde{x}$ is a sRGB image otherwise $\pi$ denotes the learnable U-Net.
 $x_t$ and $\pi{(\widetilde{x})}$ are concatenated in a channel-wise manner.

\vspace{-1mm}
\subsection{Time-melding condition}
\label{ssec: Time-melding condition}
\vspace{-1mm}
%%
%recursively, 
%
We note that the involved Markov Chain in the reverse process defined in Eq. (\ref{eq: DPM2}) contains progressively generated $x_{0:T}$ from sequenced time points $\{1,...t,...T\}$. 
The intermediate individuals $x_{t-1}$ could be obtained from $x_t$ via a one-step process defined in Eq. (\ref{eq: DPM6}) using the learned denoisers.
However, we observe that there exists uncertainties in its estimations $x_{0:T}$ of the underlying probability distributions $p_{\theta}^{(t)}(x_{t-1}\, |\, x_t, \pi{(\widetilde{x})})$.
For example, the generated intermediate sequences $x_{t-1}, x_{t}, x_{t+1}$ may contain inconsistent structural details.
Therefore we propose to define the following time-melding condition:
%
%$\hat{\mu}_{\theta} (x_{t+1}, t+1) = (x_{t+1} - \sqrt{1-\alpha_{t+1}} \cdot \epsilon_{\theta}(x_{t+1}, t+1)) / \sqrt{\alpha_{t+1}}$ as the time-melding condition 
%
\vspace{-1.5mm}
\begin{equation}
	\label{eq: DPM7}
	\bar{\mu}_{\theta}^{(t)} = \hat{\mu}_{\theta} (x_t, \pi{(\widetilde{x})}, \bar{\mu}_{\theta}^{(t+1)}, t),
	\vspace{-1.5mm}
\end{equation}
where we implement this temporal fusion for $x_t$, $\pi{(\widetilde{x})}$ and $\bar{\mu}_{\theta}^{(t+1)}$ by a channel-wise concatenated operator such that the conditions will re-weighted adaptively by the denoiser $\hat{\mu}_{\theta}(\cdot,t)$.
We then use to time-melding condition to revise Eq. (\ref{eq: DPM6}) and $x_{t-1}$ can be predicted as:
\vspace{-1.5mm}
\begin{multline}
	\label{eq: DPM8}
	x_{t-1} = \sqrt{\alpha_{t-1}} \hat{\mu}_{\theta} (x_t, \pi{(\widetilde{x})}, \bar{\mu}_{\theta}^{(t+1)}, t) \\
	+ \hat{r}_{\theta} (x_t, \pi{(\widetilde{x})}, \bar{\mu}_{\theta}^{(t+1)}, t) + \sigma_t \epsilon_t,
	\vspace{-1.5mm}
\end{multline}
The time-melding conditions $\{{\bar{\mu}_{\theta} ^{(t)}}\}_{t=1}^T$ are concatenated recursively according to Eq. (\ref{eq: DPM7}). We set ${\bar{\mu}_{\theta}^{(T)}} = x_0$.
The goal of the time-melding condition is to capture the temporal coherence across wide yet relevant time points as a single image $x_t$ may not provide enough information. 
This technique takes advantage of the observation that the information lost in one time point may be available in another time point, thus facilitates preserving as much detail and visual fidelity as possible for the progressively generation process.
By fusing valuable information from different time points, the ill-conditioning issues of the real-world degradation process are mitigated, which improves the consistency and robustness of the reverse generation process.
\begin{table*}[h!t]\footnotesize
\centering
\caption{Quantitative evaluations for the state-of-the-art methods on our SRRIIE dataset in terms of PSNR, SSIM, LPIPS \cite{zhang2018perceptual}, DISTS \cite{ding2020iqa} and FID \cite{Seitzer2020FID} metrics. We fine-tune the SOTA methods using officially released pre-trained models until they converge. Our method achieves favorable results w.r.t. reconstruction and perceptual metrics against the state-of-the-art methods for Super-resolving real-world illumination enhancement. The green and blue color denotes the best and second-best performance of the evaluated methods, respectively.}
\label{tab:comparison-sota}
\vspace{2mm}
\begin{tabular}{|l|ccccc|ccccc|}
\hline
\multirow{2}{*}{Method} & \multicolumn{5}{c|}{$\times$2}                                                                                                & \multicolumn{5}{c|}{$\times$4}                                                                                                \\ \cline{2-11} 
                        & \multicolumn{1}{c|}{PSNR $\uparrow$} & \multicolumn{1}{c|}{SSIM $\uparrow$} & \multicolumn{1}{c|}{LPIPS $\downarrow$} & \multicolumn{1}{c|}{DISTS $\downarrow$} & FID $\downarrow$ & \multicolumn{1}{c|}{PSNR $\uparrow$} & \multicolumn{1}{c|}{SSIM $\uparrow$} & \multicolumn{1}{c|}{LPIPS $\downarrow$} & \multicolumn{1}{c|}{DISTS $\downarrow$} & FID $\downarrow$ \\ 
                        \hline
                        \hline
\begin{tabular}[c]{@{}l@{}}MIRNet-v2 \cite{Zamir2022MIRNetv2}\\ $\rightarrow$ SwinIR \cite{liang2021swinir}\end{tabular}    & \multicolumn{1}{c|}{\cellcolor[HTML]{96FFFB}21.74}     & \multicolumn{1}{c|}{0.8446}     & \multicolumn{1}{c|}{0.4451}      & \multicolumn{1}{c|}{0.2498}      &156.31      & \multicolumn{1}{c|}{\cellcolor[HTML]{96FFFB}21.44}     & \multicolumn{1}{c|}{\cellcolor[HTML]{96FFFB}0.8619}     & \multicolumn{1}{c|}{0.5050}      & \multicolumn{1}{c|}{0.2600}      &160.33      \\
\hdashline
USRNet \cite{zhang2020deep}                  & \multicolumn{1}{c|}{18.09}     & \multicolumn{1}{c|}{0.8243}     & \multicolumn{1}{c|}{0.4291}      & \multicolumn{1}{c|}{0.3029}      &181.18      & \multicolumn{1}{c|}{18.00}     & \multicolumn{1}{c|}{0.8412}     & \multicolumn{1}{c|}{0.4333}      & \multicolumn{1}{c|}{0.3221}      &190.65      \\
RealSR \cite{Ji_2020_CVPR_Workshops}                  & \multicolumn{1}{c|}{-}     & \multicolumn{1}{c|}{-}     & \multicolumn{1}{c|}{-}      & \multicolumn{1}{c|}{-}      &-      & \multicolumn{1}{c|}{18.56}     & \multicolumn{1}{c|}{0.7961}     & \multicolumn{1}{c|}{0.4221}      & \multicolumn{1}{c|}{0.2476}      &153.27      \\
%MIRNet-v2 \cite{Zamir2022MIRNetv2}               & \multicolumn{1}{c|}{}     & \multicolumn{1}{c|}{}     & \multicolumn{1}{c|}{}      & \multicolumn{1}{c|}{}      &      & \multicolumn{1}{c|}{}     & \multicolumn{1}{c|}{}     & \multicolumn{1}{c|}{}      & \multicolumn{1}{c|}{}      &      \\
Real-ESRGAN \cite{wang2021realesrgan}             & \multicolumn{1}{c|}{21.72}     & \multicolumn{1}{c|}{\cellcolor[HTML]{96FFFB}0.8641}     & \multicolumn{1}{c|}{\cellcolor[HTML]{9AFF99}0.2592}      & \multicolumn{1}{c|}{\cellcolor[HTML]{9AFF99}0.1735}      &\cellcolor[HTML]{96FFFB}{108.43}      & \multicolumn{1}{c|}{19.94}     & \multicolumn{1}{c|}{0.8530}     & \multicolumn{1}{c|}{\cellcolor[HTML]{9AFF99}0.3224}      & \multicolumn{1}{c|}{\cellcolor[HTML]{9AFF99}0.2086}      &{\cellcolor[HTML]{96FFFB}138.97}      \\
SwinIR \cite{liang2021swinir}                  & \multicolumn{1}{c|}{21.25}     & \multicolumn{1}{c|}{0.8522}     & \multicolumn{1}{c|}{0.3946}      & \multicolumn{1}{c|}{0.2344}      &136.95      & \multicolumn{1}{c|}{20.70}     & \multicolumn{1}{c|}{0.8609}     & \multicolumn{1}{c|}{0.4858}      & \multicolumn{1}{c|}{0.2554}      &147.47      \\
Ours                    & \multicolumn{1}{c|}{\cellcolor[HTML]{9AFF99}24.24}     & \multicolumn{1}{c|}{\cellcolor[HTML]{9AFF99}0.8658}     & \multicolumn{1}{c|}{\cellcolor[HTML]{96FFFB}0.3493}      & \multicolumn{1}{c|}{\cellcolor[HTML]{96FFFB}0.1764}      &{\cellcolor[HTML]{9AFF99}107.09}      & \multicolumn{1}{c|}{\cellcolor[HTML]{9AFF99}21.62}     & \multicolumn{1}{c|}{\cellcolor[HTML]{9AFF99}0.8640}     & \multicolumn{1}{c|}{\cellcolor[HTML]{96FFFB}0.4326}      & \multicolumn{1}{c|}{\cellcolor[HTML]{96FFFB}0.1896}      &{\cellcolor[HTML]{9AFF99}125.13}      \\ \hline
\end{tabular}
\end{table*}

\begin{figure*}[h!tp]\footnotesize
\centering
\begin{tabular}{ccccccc}
\hspace{-1.3mm}
\includegraphics[width=0.245\linewidth, height=0.245\linewidth]{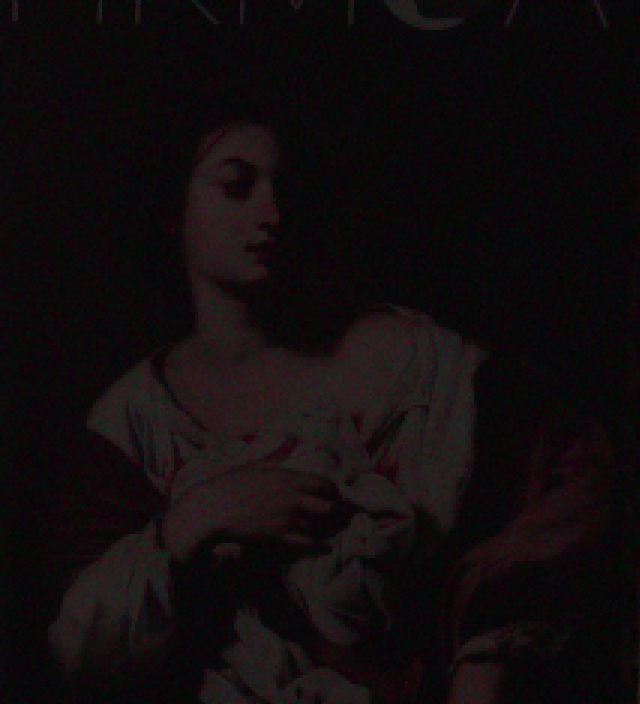} &\hspace{-4.5mm}
\includegraphics[width=0.245\linewidth, height=0.245\linewidth]{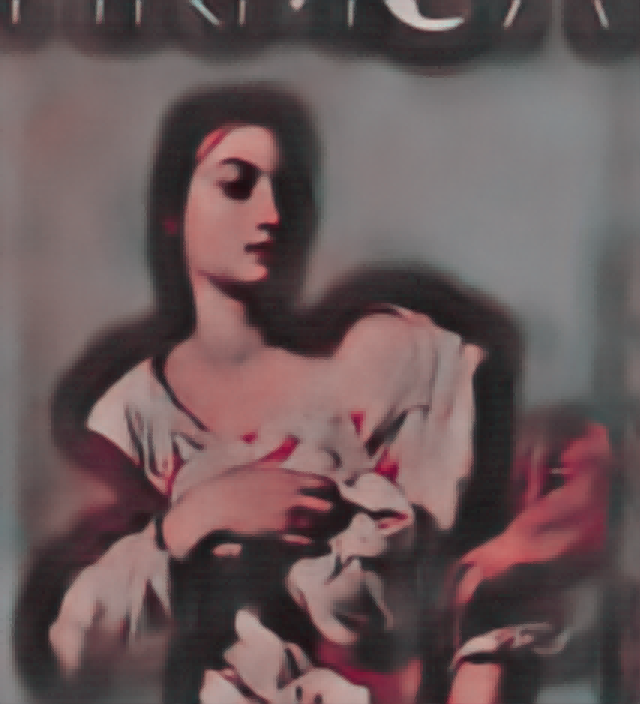}  &\hspace{-4.5mm}
\includegraphics[width=0.245\linewidth, height=0.245\linewidth]{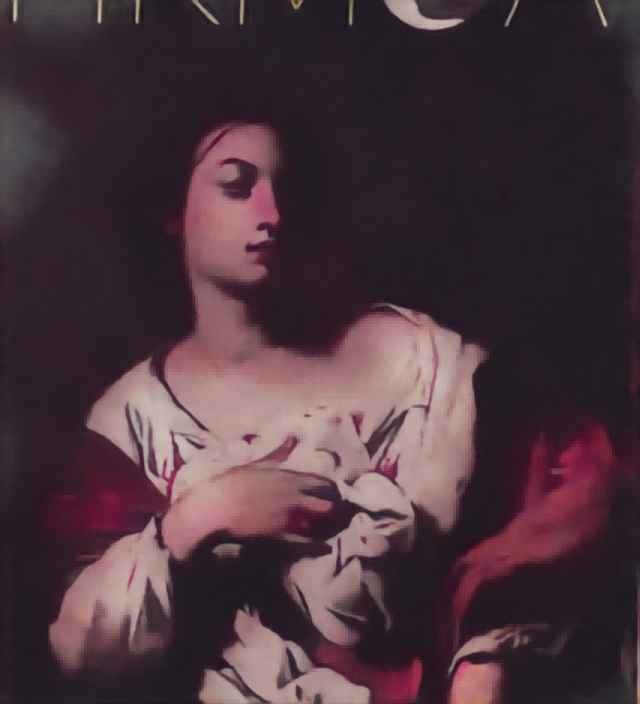} &\hspace{-4.5mm}
\includegraphics[width=0.245\linewidth, height=0.245\linewidth]{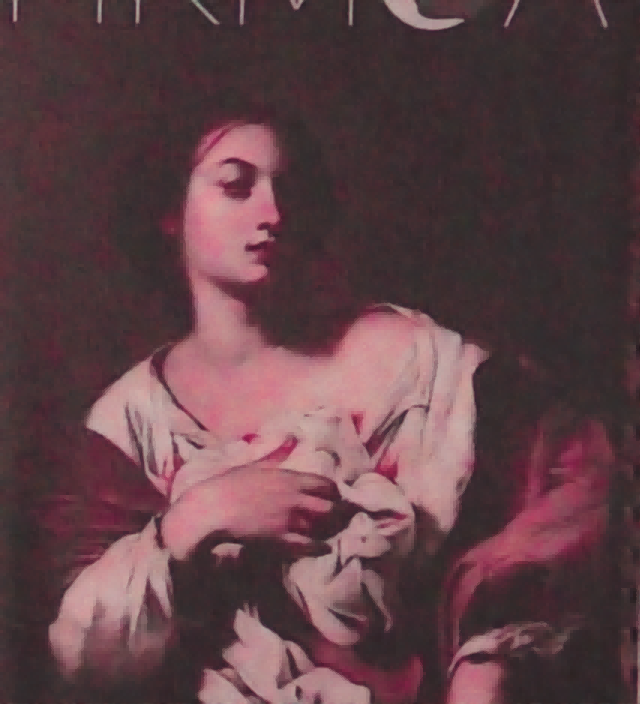} 
\\
\hspace{-1.3mm}
Input image &\hspace{-4.5mm} USRNet \cite{zhang2020deep} &\hspace{-4.5mm} MIRNet \cite{Zamir2022MIRNetv2} $\rightarrow$ SwinIR \cite{liang2021swinir} &\hspace{-4.5mm} MIRNet \cite{Zamir2022MIRNetv2}
\\
\hspace{-1.3mm}
\includegraphics[width=0.245\linewidth, height=0.245\linewidth]{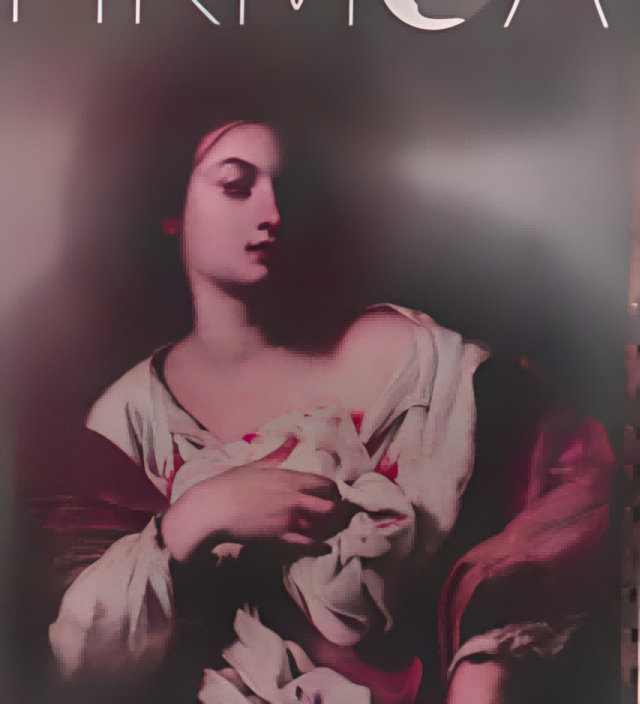} &\hspace{-4.5mm}
\includegraphics[width=0.245\linewidth, height=0.245\linewidth]{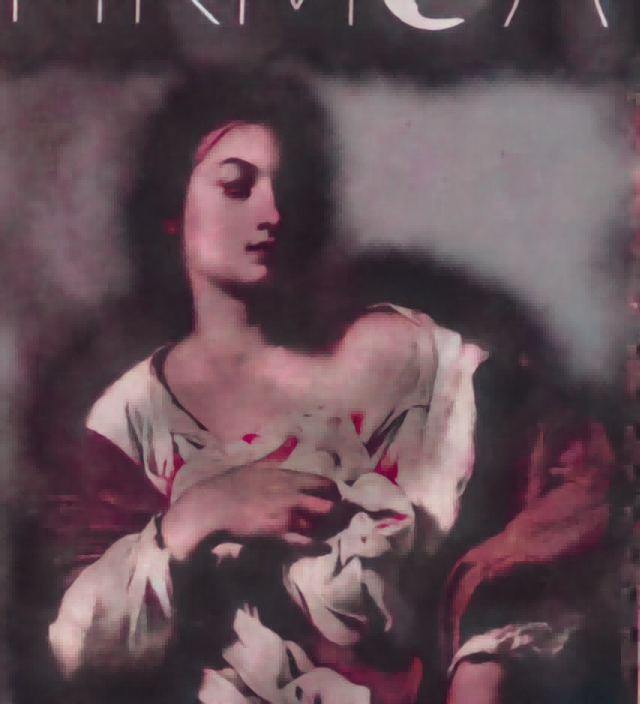} &\hspace{-4.5mm}
\includegraphics[width=0.245\linewidth, height=0.245\linewidth]{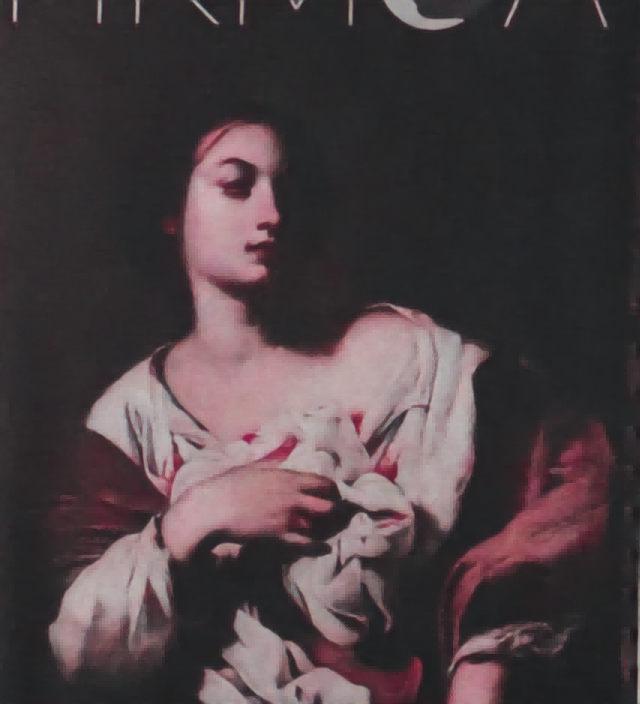} &\hspace{-4.5mm}
\includegraphics[width=0.245\linewidth, height=0.245\linewidth]{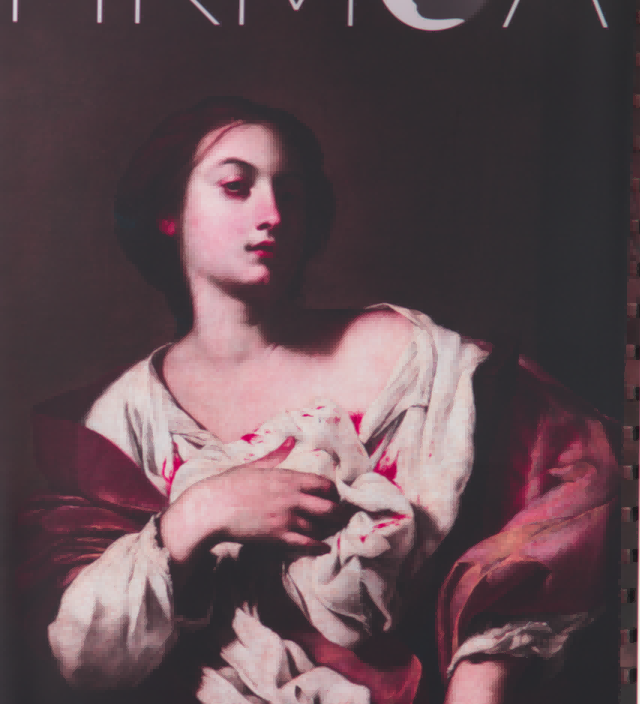}
\\
\hspace{-1.3mm}
Real-ESRGAN \cite{wang2021realesrgan} &\hspace{-4.5mm} SwinIR \cite{liang2021swinir} &\hspace{-4.5mm} Ours &\hspace{-4.5mm} Ground Truth
\end{tabular}
%\end{center}
\vspace{0mm}
\caption{Qualitative visual comparisons for $\times$2 SR (-3 EV, ISO 3200) on our SRRIIE dataset. The compared methods restore noticeable color distortions and artifacts while our method generates photo-realistic results with nature face structures.
%\emph{Zoom in for best view.}
}
\vspace{-2mm}
\label{fig: comparisons-sota-SRRIIE-x2}
\end{figure*}

\vspace{-1mm}
\section{Experimental Results}
\label{sec: experimental results}
\vspace{-1mm}
We evaluate our method against state-of-the-art illumination enhancement methods and super-resolution methods on our SRRIIE dataset.
%and use the publicly available benchmark RELLISUR \cite{aakerberg2021rellisur} dataset to validate the cross-dataset generalization.
%
We perform comprehensive experiments, and only show parts of representative results in the main paper. Please visit our supplemental material for more extensive results.
We will release our SRRIIE dataset, source codes, and trained diffusion models to the public.
	
	%
	%We empirically set the weight parameter of three stages to be 1.0.
	%

\vspace{-1mm}
\subsection{Implementation details}
\label{ssec: implementation details}
\vspace{-1mm}

%\begin{figure*}[!tp]\footnotesize
%	\centering
%	\vspace{0mm}
%	\begin{tabular}{cccccc}
%	\hspace{-2.6mm}
%	\includegraphics[width=0.195\linewidth]{figures/ablation-study/sequential-joint/106_x1_9.png} &\hspace{-4.5mm}
%	\includegraphics[width=0.195\linewidth]{figures/ablation-study/sequential-joint/0189_mirnet_IE.png} &\hspace{-4.5mm}
%	\includegraphics[width=0.195\linewidth]{figures/ablation-study/sequential-joint/0189_SwinIR_SR.png} &\hspace{-4.5mm}
%	\includegraphics[width=0.195\linewidth]{figures/ablation-study/sequential-joint/0189_SwinIR_joint.png} &\hspace{-4.5mm}
%	\includegraphics[width=0.195\linewidth]{figures/ablation-study/sequential-joint/106_x4_10.png}
%	\\
%	\hspace{-2.6mm}
%	(a) Input image &\hspace{-4.5mm} (b) MIRNet-v2 \cite{Zamir2022MIRNetv2} (IE $\rightarrow$) &\hspace{-4.5mm} (c) SwinIR \cite{liang2021swinir} ($\rightarrow$ SR) &\hspace{-4.5mm} (d) SwinIR \cite{liang2021swinir} (Joint) &\hspace{-4.5mm} (e) Ground truth
%	\\
%	\end{tabular}
%		%\end{center}
%	\vspace{0mm}
%	\caption{Qualitative visual evaluations (-2 EV, ISO 5000) for cascaded methods versus joint methods on x4 SR of SRRIIE dataset.
%	}
%	\vspace{-4mm}
%	\label{fig: cascaded method and joint method}
%\end{figure*}

\begin{figure*}[!tp]\footnotesize
	\centering
	\vspace{0mm}
	\begin{tabular}{cccccc}
	\hspace{-2.6mm}
	\includegraphics[width=0.195\linewidth]{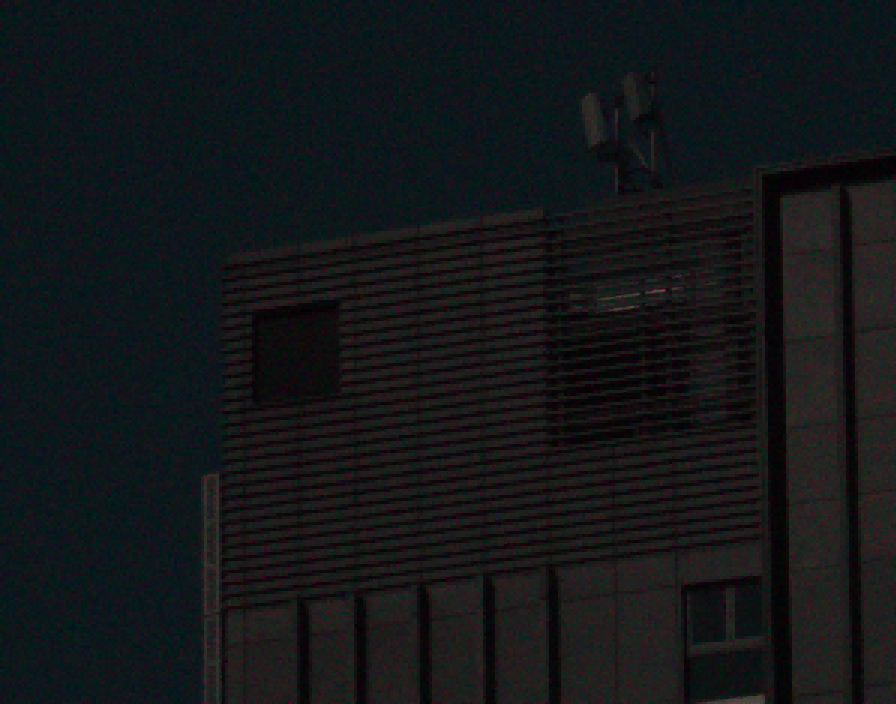} &\hspace{-4mm}
	\includegraphics[width=0.195\linewidth]{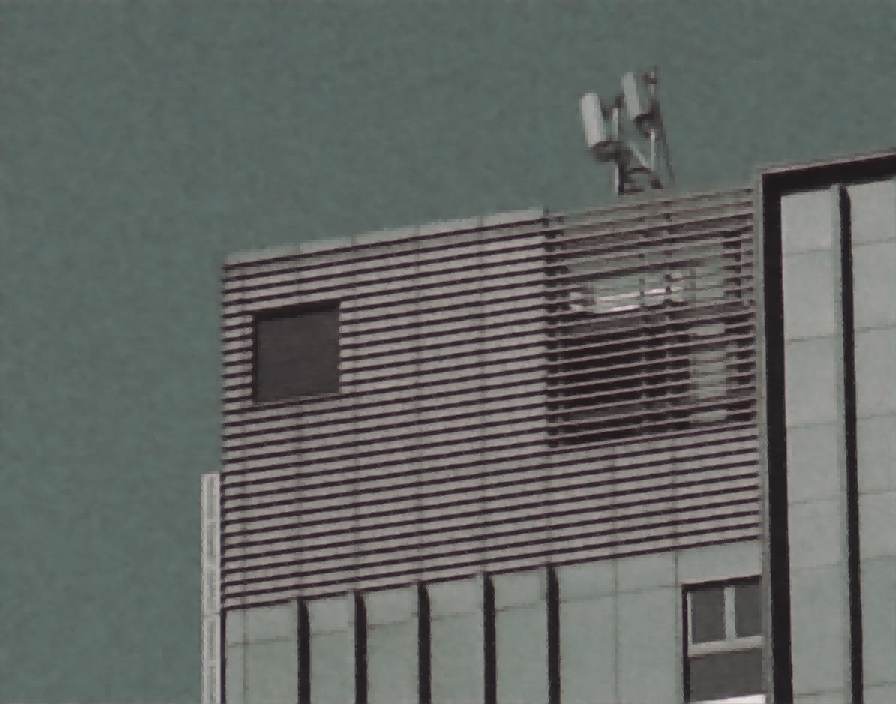} &\hspace{-4mm}
	\includegraphics[width=0.195\linewidth]{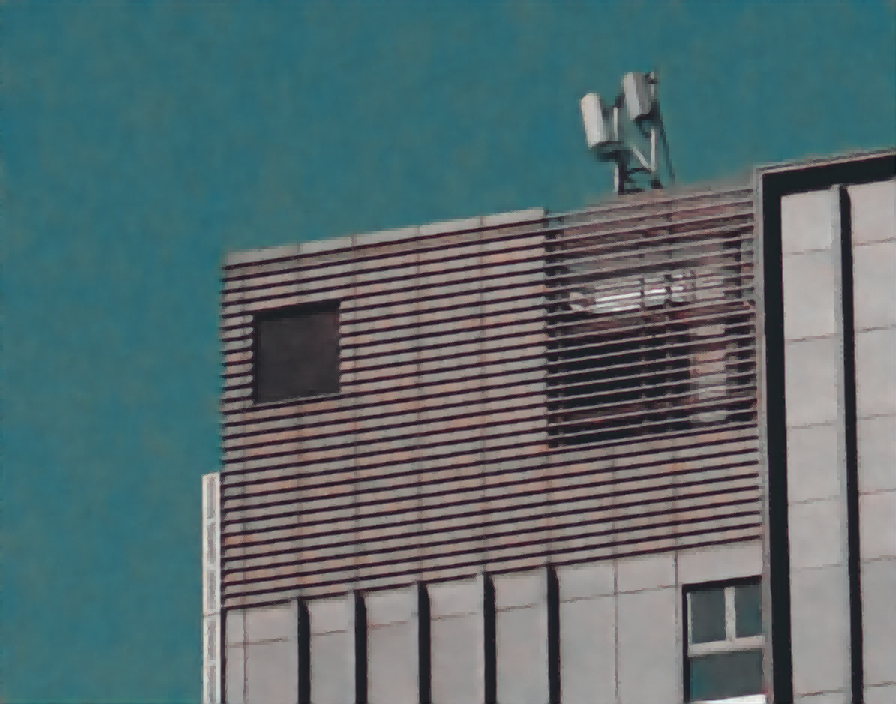} &\hspace{-4mm}
	\includegraphics[width=0.195\linewidth]{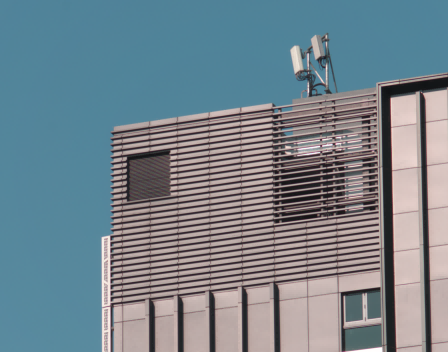}
	\\
	\hspace{-2.6mm}
	(a) Input image &\hspace{-4mm} (b) w/o TMC &\hspace{-4mm} (c) w/ TMC &\hspace{-4mm} (d) Ground truth
	\\
	\end{tabular}
		%\end{center}
	\vspace{1mm}
	\caption{Qualitative visual evaluations on the proposed TMC (Eq. (\ref{eq: DPM7})). The generated image with TMC exhibits more realistic textures and truthful colors. In contrast, the resulting image without TMC suffers from color distortion and contains more remain noises in the sky.
	}
	\vspace{-4mm}
	\label{fig: ablation-TMC}
\end{figure*}

\begin{table}[!tp]\small
  \centering
\caption{Effectiveness of the proposed TMC (Eq. (\ref{eq: DPM7})) on the validation set. The proposed TMC could effectively improve the reverse generation process to obtain better image quality.
%Quantitative evaluation results comparing the performance of the TMC conditioned and non-conditioned models, as measured by PSNR, SSIM, LPIPS, and DISTS
}
\vspace{2mm}
\begin{tabular}{|l|c|c|c|c|c|}
   \hline
    Method  & PSNR $\uparrow$ & SSIM $\uparrow$ & LPIPS $\downarrow$ & DISTS $\downarrow$ \\
    \hline
    % \hline
    % w/o TMC & \multirow{2}{*}{20000} & & \\
    % w/ TMC & & &  \\
    \hline
    %w/o TMC &20.93 &0.8641 & 0.4713 & 0.2882 \\
    %w/ TMC &22.04  &0.8682 & 0.3810 & 0.2547\\
    w/o TMC &20.83 &0.8644 & 0.4417 & 0.2794 \\
    w/ TMC &22.64  &0.8720 & 0.3671 & 0.2506\\
    \hline
  \end{tabular}
  \vspace{-2mm}
  \label{tab:ablation-TMC-table}
\end{table}

% \usepackage{multirow}

%
%
We employ a U-Net like in \cite{songdenoising, bansal2022cold} as the denoiser in our solution, which has 7 residual blocks, and the number of filters are set to 64, 128, 256, 512, respectively. In the learning process, we use the ADAM optimizer~\cite{kingma2014adam} with parameters $\beta_1 = 0.9$, $\beta_2 = 0.999$, and $\epsilon = 10^{-8}$.
The batch size is set to be 4. The learning rate is initialized as $2 \times 10^{-4}$ which is updated by SGDR \cite{loshchilov2016sgdr}. 
%T_0=5000, T_mult=2, eta_min=0, last_epoch=-1, and verbose=False. 
The parameters are initialized randomly. 
We set $T = 50$ for the DPM.
During training, the input Raw images and ground truth sRGB images are randomly cropped in a paired manner. We pack each 2 $\times$ 2 block in the Raw Bayer mosaic image into 4 channels denoting for the RGGB image.
We randomly crop non-overlapping 50,000 256 $\times$ 256 (512 $\times$ 512 and 1024 $\times$ 1024 for $\times$2 and $\times$4 HR) paired patches for training. The entire network is trained using the PyTorch framework.
 %into 256 $\times$ 256 patches
 %
900 and 900 full-resolution images are used for the validation and test phase, where each high-resolution ground truth image corresponds to 9 low-light low-resolution input images.

\vspace{-1mm}
\subsection{Comparisons with the state-of-the-arts}
\label{ssec: comparisons-sota}
\vspace{-1mm}
{\flushleft \bf {Evaluation Metrics.}}
Our proposed SRRIIE dataset contains 4800 paired low-high quality images.
We employ PSNR and SSIM on the Y channel in the YCbCr color space to evaluate the reconstruction accuracy. 
We employ LPIPS \cite{zhang2018perceptual}, DISTS \cite{ding2020iqa} and FID \cite{Seitzer2020FID} in sRGB color space to evaluate the human perception of image quality,  
%and as well as a non-reference metric NIQE \cite{Mittal2013} in the RGB color space.
%
which have been validated effectively especially in generation tasks of image structures and details.
Higher PSNR and SSIM values indicate higher reconstruction quality, while lower LPIPS, DISTS, and FID values indicate higher perceptual quality.
All the methods are strictly evaluated with the same settings on the 900 full-resolution test images.

\vspace{-2.5mm}
{\flushleft \bf {Quantitative and qualitative results.}}
Table~\ref{tab:comparison-sota} shows the quantitative evaluations on different benchmark datasets, where the results of the state-of-the-art methods are obtained using the corresponding publicly available codes and models for fair comparisons.
	The proposed method achieves favorable restored images compared to state-of-the-art methods in terms of reconstruction and perceptual metrics.
	All of the compared methods are strictly evaluated with the same settings.
	We fine-tune the SOTA methods using their officially released pre-trained models to achieve better results instead of training from scratch.
Figure \ref{fig: comparisons-sota-SRRIIE-x2} shows a challenging portrait image captured with the under-exposure level of -3 EV and ISO 3200. The complicated noises and low light discourage existing methods as the assumed degradation models deviate from those in this real-world low-light noisy case.
The ResNet-based methods \cite{zhang2020deep} suffer from over-smoothing problems.
GAN-based methods \cite{wang2021realesrgan} generate unpleasant artifacts and untruthful colors.
Attention-based methods \cite{Zamir2022MIRNetv2, liang2021swinir} perform less effectively in this real-world low-contrast noisy case.
In contrast, our method generate perceptually pleasing results and preserve more structural details.

\vspace{-1mm}
\section{Analysis and Discussions}
\label{sec: analysis and discussions}
\vspace{-1mm}

\vspace{-1mm}
\subsection{Effectiveness of the time-melding condition}
\label{ssec: Effectiveness of the time-melding condition}
\vspace{-1mm}
%
%
%Cascaded methods and joint methods are two alternative solutions to super-solve image illumination enhancement tasks.
%%
%We further analyze their effectivenesses and ineffectivenesses.
%%
%As shown in Figure \ref{fig: cascaded method and joint method} (b-c), we first use MIRNet-v2 \cite{Zamir2022MIRNetv2} to enhance the input low-light image, then use SwinIR \cite{liang2021swinir} to super-solve the enhanced output image by a scale of 4.
%%
%Due to the biased colors and over-smoothing effects introduced by the first stage, the SR model in the second stage performs less effectively in correcting color distortions and generating lost detailed textures.
%%
%On the contrast, the joint method in Figure \ref{fig: cascaded method and joint method} (d) restores more truthful colors especially the hairs.
%%
%The quantitative results in Table \ref{tab:comparison-sota} show that the cascaded method MIRNet-v2 \cite{Zamir2022MIRNetv2} $\rightarrow$ SwinIR \cite{liang2021swinir} achieves higher PSNR values, but performs less effectively in terms of other metrics. 
%%
%This is mainly due to the over-smoothing effects in the first stage by reducing noises.
%%
%Considering the biased estimations and information loss issues of cascaded methods, we suggest joint methods to be the preferential direction in the future research.

Our proposed time-melding condition (TMC) fuses valuable information across relevant time points. The recursively strategy facilitates preserving detail and visual fidelity from multiple frames for the progressively generation process in Eq.(\ref{eq: DPM8}).
	To demonstrate its effectiveness, we disable Eq. (\ref{eq: DPM7}) in the proposed method and retrain with the same settings for fair comparisons.
	Figure \ref{fig: ablation-TMC} (c) shows the result using TMC, where the structural details and colors are restored well. In contrast, the result without TMC contains more remain noises and color distortion as shown in Figure \ref{fig: ablation-TMC} (b).
	The quantitative comparisons in Table \ref{tab:ablation-TMC-table}  demonstrates that using TMC achieves better restoration results in terms of PSNR, SSIM, LPIPS and DISTS metrics on the validation set of our SRRIIE dataset.
	%
	%Overall, the knowledge distillation network does not need any user-specified label and is trained in an unpaired learnning manner.
	%
	Our conditional DPM-based method benefits from the time-melding condition, thus further improves the consistency and robustness of the reverse generation process to generate better image quality.

\vspace{-1mm}
\subsection{Effectiveness of the Raw sensor data}
\label{ssec: effectiveness of RAW restoration}
\begin{figure}[!tp]\footnotesize
	\centering
	\vspace{0mm}
	\begin{tabular}{cccccc}
	\hspace{-2.6mm}
	\includegraphics[width=0.46\linewidth]{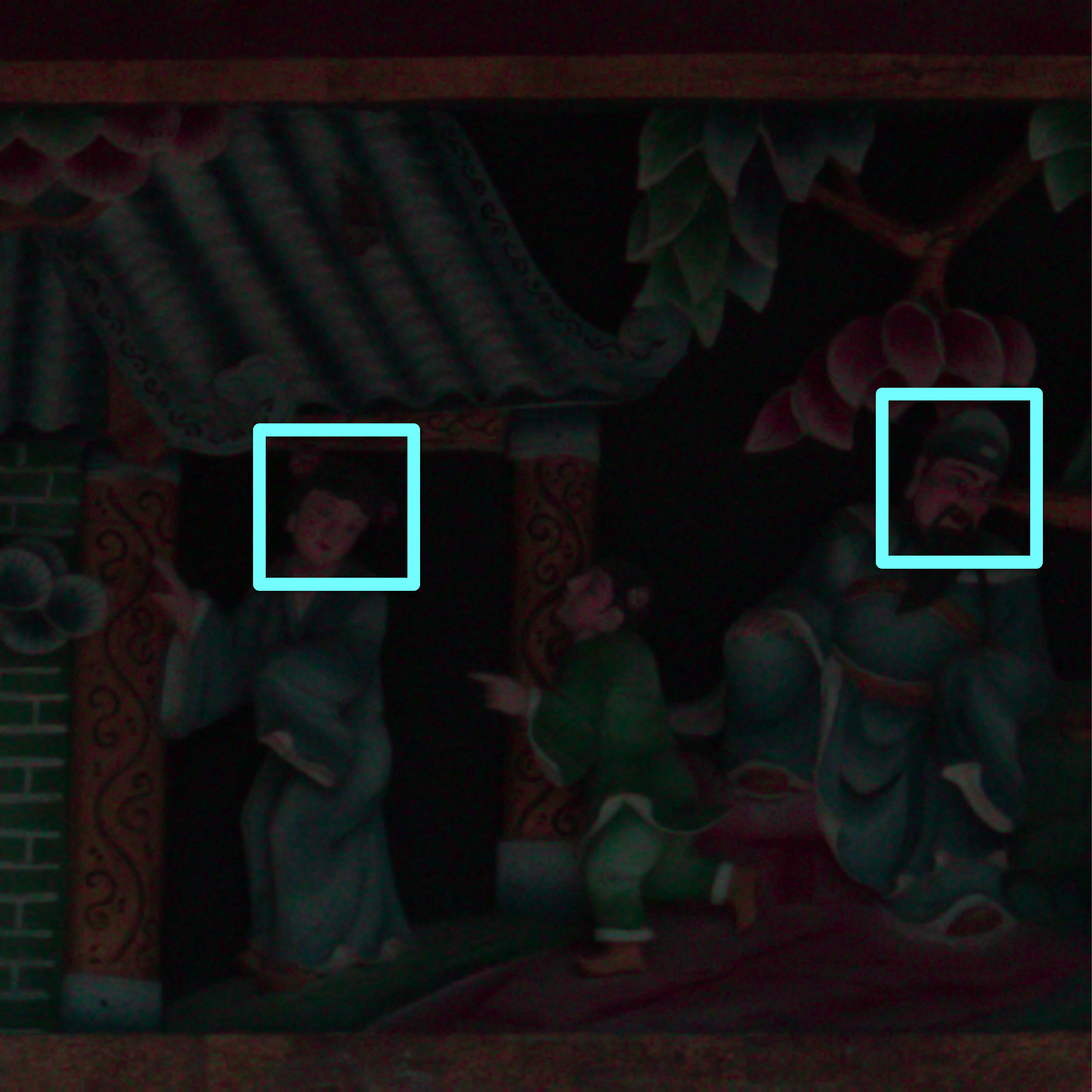} &\hspace{-4.5mm}
	\includegraphics[width=0.46\linewidth]{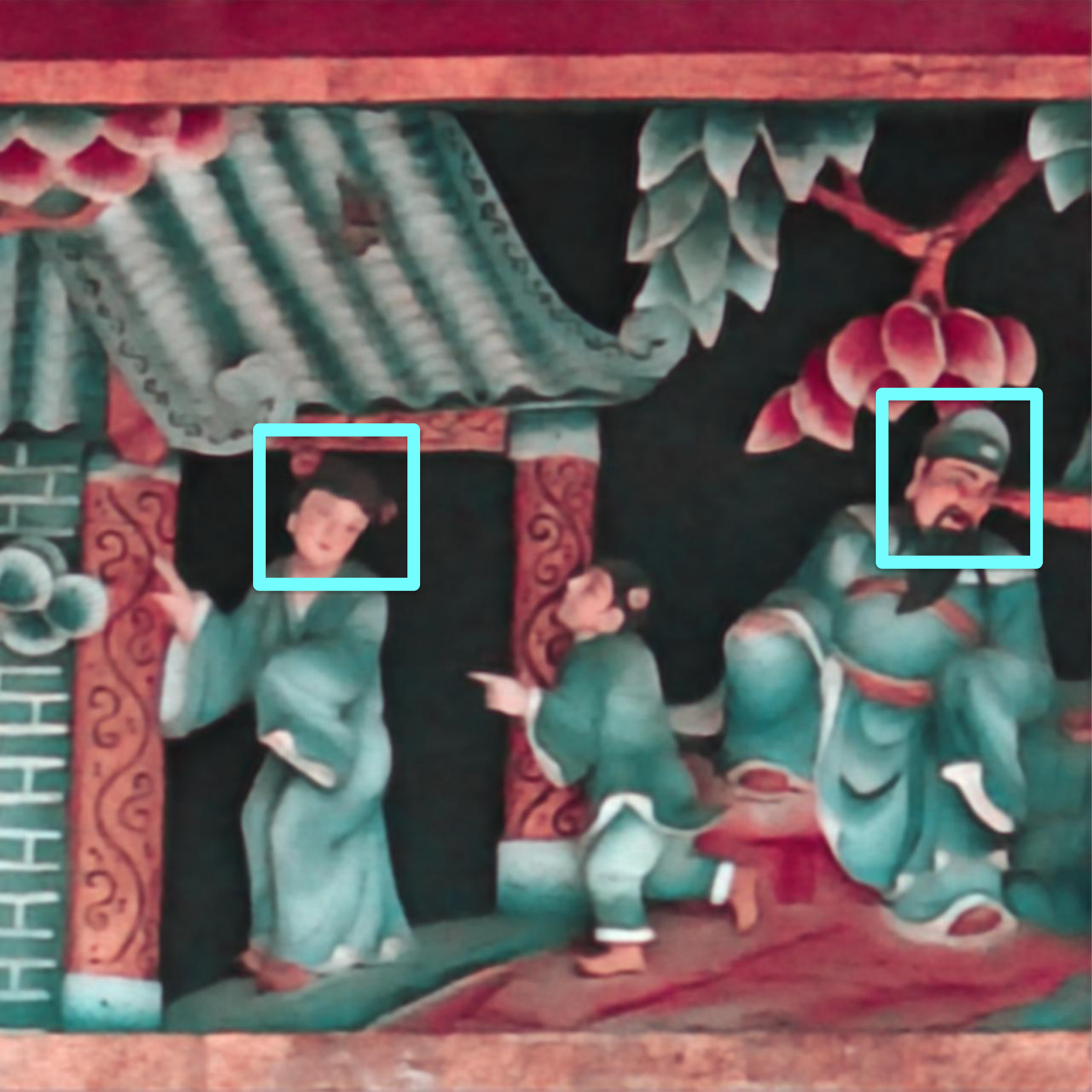} 
	\\
	\hspace{-2.6mm}
	(a) Input image &\hspace{-4.5mm} (b) Ours-sRGB
	\\
	\hspace{-2.6mm}
	\includegraphics[width=0.46\linewidth]{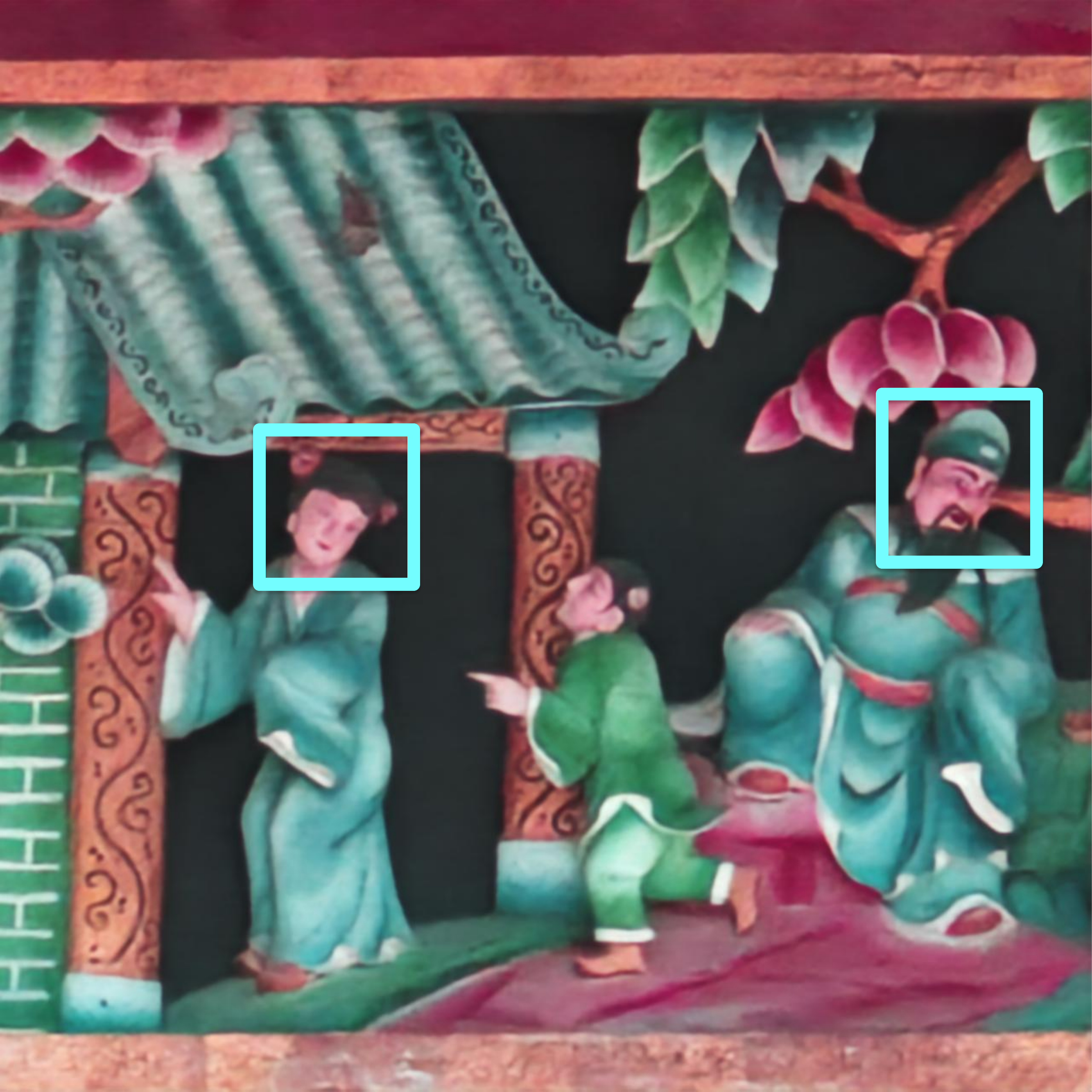} &\hspace{-4.5mm}
	\includegraphics[width=0.46\linewidth]{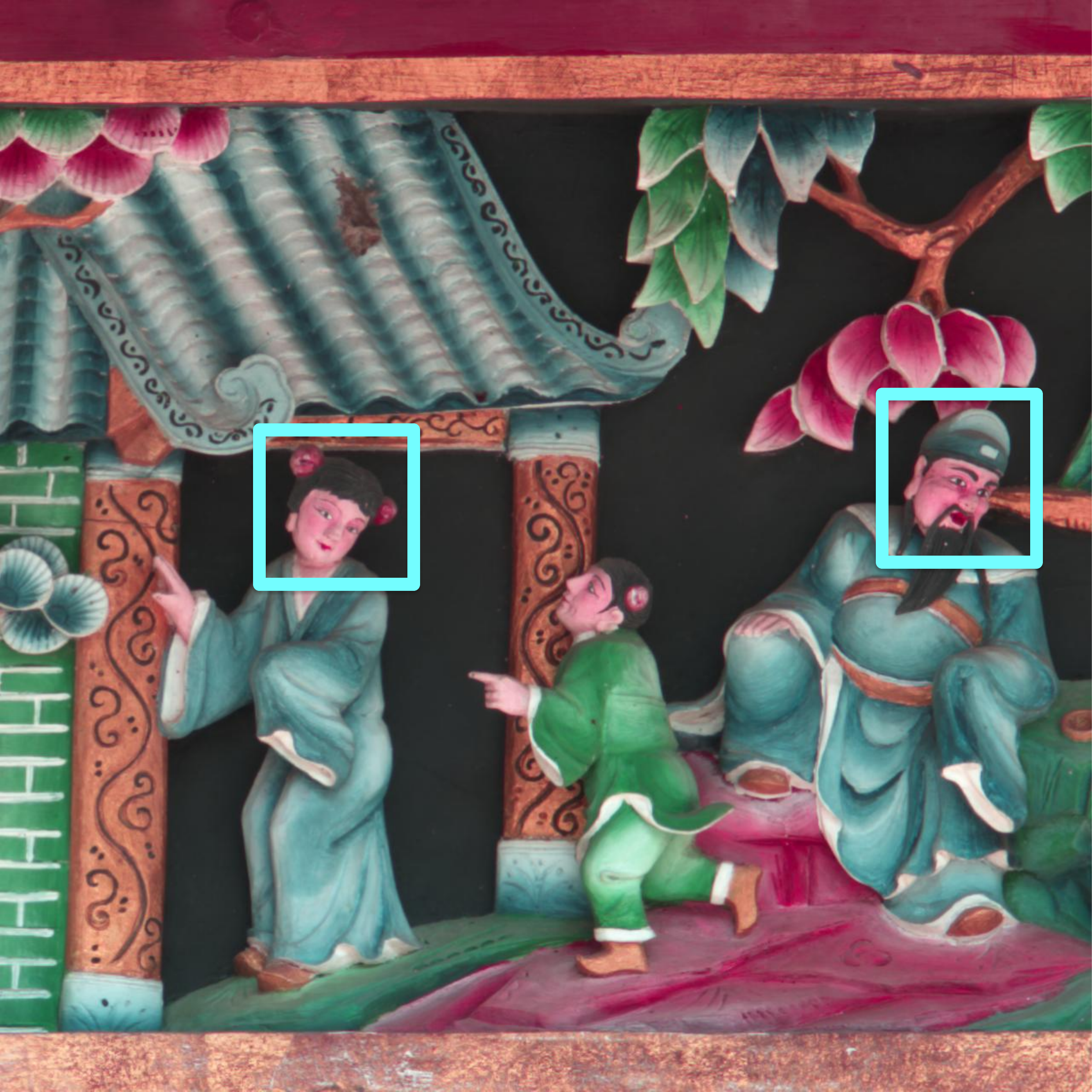} 
	\\
	\hspace{-2.6mm}
	(c) Ours-Raw &\hspace{-4.5mm} (d) Ground truth
		\end{tabular}
		%\end{center}
	\vspace{2mm}
	\caption{The method trained on real Raw sensor data could restore more truthful and clear face details compared to the method trained on sRGB images.
	}
	\vspace{-2mm}
	\label{fig: effectiveness of Raw restoration}
\end{figure}

\vspace{-1mm}
We provide both Raw sensor data and sRGB data for our SRRIIE dataset.
%since most of the existing methods are designed on sRGB color space. 
%
To demonstrate the effectivenesses of the real Raw sensor data, we retrain our diffusion model and disable $\pi$ in Eq. (\ref{eq: DPM6}). 
%the Raw processing module with the widely used sRGB processing module for the network inputs. 
%
Figure \ref{fig: effectiveness of Raw restoration} shows the qualitative visual results, where our method trained on Raw images achieves clearer and sharper face structures.
The quantitative results summarized in Table \ref{tab: effectiveness of Raw restoration} demonstrate that real Raw sensor data could consistently improve the restoration performances against the sRGB data in terms of both reconstruction and perceptual metrics.
%Effectiveness of the Raw sensor data%
Raw sensor data contains valuable information that is necessary for accurate restoration and enhancement, making it beneficial for image restoration. In contrast, the compression process used for low-bit sRGB images causes the loss of essential structures and details, leading to more ill-posed problems. Therefore, these experiments indicate that Raw sensor data is a better option for super-resolving real-world image illumination enhancement tasks as it maintains the lossless information required for restoration and enhancement.

	\begin{table}[!tp]\footnotesize
\centering
\caption{Quantitative evaluations of the methods trained on real Raw sensor data and 8-bit sRGB data of our SRRIIE dataset.}
\vspace{2mm}
\label{tab: effectiveness of Raw restoration}
\begin{tabular}{|c|c|c|c|c|c|}
\hline
Scale               & Method   & PSNR $\uparrow$ & SSIM $\uparrow$ & LPIPS $\downarrow$ & DISTS $\downarrow$ \\ 
\hline
\hline
\multirow{2}{*}{$\times$2} & Ours-sRGB &23.98      &0.8457      &0.4572       &0.2366      \\
                    & Ours-Raw &24.50      &0.8654      &0.3652       &0.1910      \\ \hline
\multirow{2}{*}{$\times$4} & Ours-sRGB &22.64      &0.8570      &0.5470       &0.2478      \\
                   & Ours-Raw &22.80      &0.8690      &0.4750       &0.2210      \\ \hline
\end{tabular}
\vspace{-6mm}
\end{table}

%\vspace{-1mm}
%\subsection{Effectiveness of diffusion model}
%\label{ssec: effectiveness of diffusion model}
%\vspace{-1mm}

%\begin{figure}[h!tp]\footnotesize
%	\centering
%	\vspace{0mm}
%	\begin{tabular}{cccccc}
%	\hspace{-2.6mm}
%	\includegraphics[width=0.33\linewidth]{figures/SRRIIE-dataset/0171.png} &\hspace{-4.5mm}
%	\includegraphics[width=0.33\linewidth]{figures/SRRIIE-dataset/0171.png} &\hspace{-4.5mm}
%	\includegraphics[width=0.33\linewidth]{figures/SRRIIE-dataset/0171.png} 
%	\\
%	\hspace{-2.6mm}
%	Input image &\hspace{-4.5mm} BSRGAN \cite{zhang2021designing} &\hspace{-4.5mm} SwinIR \cite{liang2021swinir}
%	\\
%	\hspace{-2.6mm}
%	\includegraphics[width=0.33\linewidth]{figures/SRRIIE-dataset/0171.png} &\hspace{-4.5mm}
%	\includegraphics[width=0.33\linewidth]{figures/SRRIIE-dataset/0171.png} &\hspace{-4.5mm}
%	\includegraphics[width=0.33\linewidth]{figures/SRRIIE-dataset/0171.png} 
%	\\
%	\hspace{-2.6mm}
%	MIRNet-v2 \cite{Zamir2022MIRNetv2} &\hspace{-4.5mm} Our diffusion model &\hspace{-4.5mm} Ground truth
%		\end{tabular}
%		%\end{center}
%	\vspace{0.5mm}
%	\caption{Effectiveness of diffusion model.
%	}
%	\vspace{-2mm}
%	\label{fig: effectiveness of diffusion model}
%\end{figure}

\vspace{-1mm}
\subsection{Cross-dataset generalization}
\label{ssec: cross-dataset generalization}
\vspace{-1mm}
Our SRRIIE dataset is captured using a Sony A7 IV camera, with exposure levels ranging from -6 EV to 0 EV and ISO levels ranging from 50 to 12800. To assess the generalizability across different sensors and datasets, we employ cross-validation by training the model on one dataset and testing it on the other dataset. The RELLISUR dataset is captured using a Canon camera, with exposure levels ranging from -5 EV to -2.5 EV and ISO levels ranging from 100 to 400. In Table \ref{tab:cross-dataset generalization}, we summarize the results of the model trained on RELLISUR and SRRIIE datasets, respectively. The results show that the model trained on SRRIIE dataset still performs well on RELLISUR dataset, whereas the model trained on RELLISUR dataset performs less effectively on SRRIIE dataset. The SRRIIE dataset contains real-world degradation processes in low-light conditions with complicated noises, making it more challenging.
\begin{table}[!tp]\small
\centering
\caption{Cross-dataset generalization test in terms of PSNR/SSIM. We train our network on one dataset then test on the other dataset. 
%Our MIRNet shows good generalization for all possible cases.
}
\vspace{2mm}
\label{tab:cross-dataset generalization}
\begin{tabular}{|l|c|cc|}
\hline
\multirow{3}{*}{Method} & \multirow{3}{*}{Trained} & \multicolumn{2}{c|}{Tested}                                                                                                                                                   \\ \cline{3-4} 
                        &                          & \multicolumn{1}{c|}{\multirow{2}{*}{\begin{tabular}[c]{@{}c@{}}RELLISUR \\ (Canon)\end{tabular}}} & \multirow{2}{*}{\begin{tabular}[c]{@{}c@{}}SRRIIE\\ (Sony)\end{tabular}} \\
                        &                          & \multicolumn{1}{c|}{}                                                                            &                                                                            \\ 
                        \hline
                        \hline
\multirow{2}{*}{Ours-sRGB}   & RELLISUR                 & \multicolumn{1}{c|}{24.64/0.8167}                                                                            &19.26/0.5551                                                                            \\
                        & SRRIIE                 & \multicolumn{1}{c|}{20.65/0.7601}                                                                            &23.98/0.8457                                                                            \\ \hline
\end{tabular}
\vspace{-6mm}
\end{table}

%\begin{figure}[!t]\footnotesize
%	\centering
%	\vspace{0mm}
%	\begin{tabular}{cccccc}
%	\hspace{-2mm}
%	\includegraphics[width=0.24\linewidth]{figures/limitations/input.png} &\hspace{-4.5mm}
%	\includegraphics[width=0.24\linewidth]{figures/limitations/SwinIR.png} &\hspace{-4.5mm}
%	\includegraphics[width=0.24\linewidth]{figures/limitations/ours-raw.png} &\hspace{-4.5mm}
%	\includegraphics[width=0.24\linewidth]{figures/limitations/gt.png}
%	\\
%	\hspace{-2mm}
%	(a) Input image &\hspace{-4.5mm} (b) SwinIR \cite{liang2021swinir}  &\hspace{-4.5mm} (c) Ours &\hspace{-4.5mm} (d) Ground truth
%	\\
%		\end{tabular}
%		%\end{center}
%	\vspace{2mm}
%	\caption{The evaluated methods perform less effectively in this example image captured with the camera settings of  -6 EV and ISO 12800, where the image structures do not maintain and the image contents suffer from color distortions in this X2 SR task.
%	}
%	\vspace{-4mm}
%	\label{fig: limitations}
%\end{figure}

\vspace{-1mm}
\section{Conclusions}
\label{sec: conclusions}
\vspace{-1mm}
In this paper, we have addressed the need for a high-quality Raw image dataset for super-resolving real-world illumination enhancement tasks.
Towards this goal, we collect 4800 paired low-high quality images with a wide range of under exposure levels from -6 EV to 0 EV and ISO levels from 50 to 12800.
We conduct comprehensive benchmark experiments for existing methods on the proposed SRRIIE dataset with various reconstruction and perceptual metrics.
As existing methods perform less effectively in real-world restoration problems with complicated noises, we develope a conditional DPM-based method to progressively generate truthful structural details from real Raw sensor data.
Our proposed method achieves promising results in both quantitative and qualitative experiments compared to state-of-the-art methods. We hope that our SRRIIE dataset as well as the proposed conditional diffusion process model will advance super-resolving real-world illumination enhancement tasks.
\clearpage
	{\small
		\bibliographystyle{ieee_fullname}
		\bibliography{egbib}

\begin{thebibliography}{10}\itemsep=-1pt

\bibitem{aakerberg2021rellisur}
Andreas Aakerberg, Kamal Nasrollahi, and Thomas~B Moeslund.
\newblock Rellisur: A real low-light image super-resolution dataset.
\newblock In {\em NeurIPS Datasets and Benchmarks}, 2021.

\bibitem{abdelhamed2018high}
Abdelrahman Abdelhamed, Stephen Lin, and Michael~S Brown.
\newblock A high-quality denoising dataset for smartphone cameras.
\newblock In {\em CVPR}, pages 1692--1700, 2018.

\bibitem{agustsson2017ntire}
Eirikur Agustsson and Radu Timofte.
\newblock Ntire 2017 challenge on single image super-resolution: Dataset and study.
\newblock In {\em CVPR workshops}, pages 126--135, 2017.

\bibitem{ba2016layer}
Jimmy~Lei Ba, Jamie~Ryan Kiros, and Geoffrey~E Hinton.
\newblock Layer normalization.
\newblock {\em arXiv preprint}, 2016.

\bibitem{bansal2022cold}
Arpit Bansal, Eitan Borgnia, Hong-Min Chu, Jie~S Li, Hamid Kazemi, Furong Huang, Micah Goldblum, Jonas Geiping, and Tom Goldstein.
\newblock Cold diffusion: Inverting arbitrary image transforms without noise.
\newblock {\em arXiv preprint}, 2022.

\bibitem{bevilacqua2012low}
Marco Bevilacqua, Aline Roumy, Christine Guillemot, and Marie~Line Alberi-Morel.
\newblock Low-complexity single-image super-resolution based on nonnegative neighbor embedding.
\newblock 2012.

\bibitem{Cai2018deep}
Jianrui Cai, Shuhang Gu, and Lei Zhang.
\newblock Learning a deep single image contrast enhancer from multi-exposure images.
\newblock {\em IEEE TIP}, 27(4):2049--2062, 2018.

\bibitem{cai2019toward}
Jianrui Cai, Hui Zeng, Hongwei Yong, Zisheng Cao, and Lei Zhang.
\newblock Toward real-world single image super-resolution: A new benchmark and a new model.
\newblock In {\em ICCV}, pages 3086--3095, 2019.

\bibitem{chen2018learning}
Chen Chen, Qifeng Chen, Jia Xu, and Vladlen Koltun.
\newblock Learning to see in the dark.
\newblock In {\em CVPR}, pages 3291--3300, 2018.

\bibitem{chen2019camera}
Chang Chen, Zhiwei Xiong, Xinmei Tian, Zheng-Jun Zha, and Feng Wu.
\newblock Camera lens super-resolution.
\newblock In {\em CVPR}, pages 1652--1660, 2019.

\bibitem{chen2021pre}
Hanting Chen, Yunhe Wang, Tianyu Guo, Chang Xu, Yiping Deng, Zhenhua Liu, Siwei Ma, Chunjing Xu, Chao Xu, and Wen Gao.
\newblock Pre-trained image processing transformer.
\newblock In {\em CVPR}, pages 12299--12310, 2021.

\bibitem{chung2022come}
Hyungjin Chung, Byeongsu Sim, and Jong~Chul Ye.
\newblock Come-closer-diffuse-faster: Accelerating conditional diffusion models for inverse problems through stochastic contraction.
\newblock In {\em CVPR}, pages 12413--12422, 2022.

\bibitem{coltuc2006exact}
Dinu Coltuc, Philippe Bolon, and J-M Chassery.
\newblock Exact histogram specification.
\newblock {\em IEEE TIP}, 15(5):1143--1152, 2006.

\bibitem{croitoru2022diffusion}
Florinel-Alin Croitoru, Vlad Hondru, Radu~Tudor Ionescu, and Mubarak Shah.
\newblock Diffusion models in vision: A survey.
\newblock {\em arXiv preprint arXiv:2209.04747}, 2022.

\bibitem{dai2019transformer}
Zihang Dai, Zhilin Yang, Yiming Yang, Jaime Carbonell, Quoc~V Le, and Ruslan Salakhutdinov.
\newblock Transformer-xl: Attentive language models beyond a fixed-length context.
\newblock {\em arXiv preprint}, 2019.

\bibitem{dhariwal2021diffusion}
Prafulla Dhariwal and Alexander Nichol.
\newblock Diffusion models beat gans on image synthesis.
\newblock {\em NeurIPS}, 34:8780--8794, 2021.

\bibitem{ding2020iqa}
Keyan Ding, Kede Ma, Shiqi Wang, and Eero~P. Simoncelli.
\newblock Image quality assessment: Unifying structure and texture similarity.
\newblock {\em CoRR}, abs/2004.07728, 2020.

\bibitem{dumoulin2016guide}
Vincent Dumoulin and Francesco Visin.
\newblock A guide to convolution arithmetic for deep learning.
\newblock {\em arXiv preprint}, 2016.

\bibitem{Fang_2022_CVPR}
Jinsheng Fang, Hanjiang Lin, Xinyu Chen, and Kun Zeng.
\newblock A hybrid network of cnn and transformer for lightweight image super-resolution.
\newblock In {\em CVPR Workshops}, pages 1103--1112, June 2022.

\bibitem{guo2016lime}
Xiaojie Guo, Yu Li, and Haibin Ling.
\newblock {LIME}: Low-light image enhancement via illumination map estimation.
\newblock {\em IEEE TIP}, 26(2):982--993, 2016.

\bibitem{he2016deep}
Kaiming He, Xiangyu Zhang, Shaoqing Ren, and Jian Sun.
\newblock Deep residual learning for image recognition.
\newblock In {\em CVPR}, pages 770--778, 2016.

\bibitem{hendrycks2016gaussian}
Dan Hendrycks and Kevin Gimpel.
\newblock Gaussian error linear units (gelus).
\newblock {\em arXiv preprint}, 2016.

\bibitem{ho2020denoising}
Jonathan Ho, Ajay Jain, and Pieter Abbeel.
\newblock Denoising diffusion probabilistic models.
\newblock {\em NeurIPS}, 33:6840--6851, 2020.

\bibitem{2007brightness}
Haidi Ibrahim and Nicholas Sia~Pik Kong.
\newblock Brightness preserving dynamic histogram equalization for image contrast enhancement.
\newblock {\em IEEE Transactions on Consumer Electronics}, 53(4):1752--1758, 2007.

\bibitem{Ji_2020_CVPR_Workshops}
Xiaozhong Ji, Yun Cao, Ying Tai, Chengjie Wang, Jilin Li, and Feiyue Huang.
\newblock Real-world super-resolution via kernel estimation and noise injection.
\newblock In {\em CVPR Workshops}, June 2020.

\bibitem{EG}
Yifan Jiang, Xinyu Gong, Ding Liu, Yu Cheng, Chen Fang, Xiaohui Shen, Jianchao Yang, Pan Zhou, and Zhangyang Wang.
\newblock {EnlightenGAN}: Deep light enhancement without paired supervision.
\newblock {\em IEEE TIP}, PP:1--1, 01 2021.

\bibitem{kawar2022denoising}
Bahjat Kawar, Michael Elad, Stefano Ermon, and Jiaming Song.
\newblock Denoising diffusion restoration models.
\newblock {\em arXiv preprint arXiv:2201.11793}, 2022.

\bibitem{kim2016accurate}
Jiwon Kim, Jung~Kwon Lee, and Kyoung~Mu Lee.
\newblock Accurate image super-resolution using very deep convolutional networks.
\newblock In {\em CVPR}, pages 1646--1654, 2016.

\bibitem{kingma2014adam}
Diederik~P Kingma and Jimmy Ba.
\newblock Adam: A method for stochastic optimization.
\newblock {\em arXiv preprint arXiv:1412.6980}, 2014.

\bibitem{ledig2017photo}
Christian Ledig, Lucas Theis, Ferenc Husz{\'a}r, Jose Caballero, Andrew Cunningham, Alejandro Acosta, Andrew Aitken, Alykhan Tejani, Johannes Totz, Zehan Wang, et~al.
\newblock Photo-realistic single image super-resolution using a generative adversarial network.
\newblock In {\em CVPR}, pages 4681--4690, 2017.

\bibitem{lee2013contrast}
Chulwoo Lee, Chul Lee, and Chang-Su Kim.
\newblock Contrast enhancement based on layered difference representation of {2D} histograms.
\newblock {\em IEEE TIP}, 22(12):5372--5384, 2013.

\bibitem{9369102}
Chongyi Li, Chunle Guo, and Chen~Change Loy.
\newblock Learning to enhance low-light image via zero-reference deep curve estimation.
\newblock {\em IEEE TPAMI}, 44(8):4225--4238, 2022.

\bibitem{li2022srdiff}
Haoying Li, Yifan Yang, Meng Chang, Shiqi Chen, Huajun Feng, Zhihai Xu, Qi Li, and Yueting Chen.
\newblock Srdiff: Single image super-resolution with diffusion probabilistic models.
\newblock {\em Neurocomputing}, 479:47--59, 2022.

\bibitem{li2018structure}
Mading Li, Jiaying Liu, Wenhan Yang, Xiaoyan Sun, and Zongming Guo.
\newblock Structure-revealing low-light image enhancement via robust retinex model.
\newblock {\em IEEE TIP}, 27(6):2828--2841, 2018.

\bibitem{liang2021swinir}
Jingyun Liang, Jiezhang Cao, Guolei Sun, Kai Zhang, Luc Van~Gool, and Radu Timofte.
\newblock Swinir: Image restoration using swin transformer.
\newblock In {\em ICCV Workshops}, pages 1833--1844, 2021.

\bibitem{lim2017enhanced}
Bee Lim, Sanghyun Son, Heewon Kim, Seungjun Nah, and Kyoung Mu~Lee.
\newblock Enhanced deep residual networks for single image super-resolution.
\newblock In {\em CVPR Workshops}, pages 136--144, 2017.

\bibitem{liu2021retinex}
Risheng Liu, Long Ma, Jiaao Zhang, Xin Fan, and Zhongxuan Luo.
\newblock Retinex-inspired unrolling with cooperative prior architecture search for low-light image enhancement.
\newblock In {\em CVPR}, pages 10561--10570, 2021.

\bibitem{liu2022convnet}
Zhuang Liu, Hanzi Mao, Chao-Yuan Wu, Christoph Feichtenhofer, Trevor Darrell, and Saining Xie.
\newblock A convnet for the 2020s.
\newblock In {\em CVPR}, pages 11976--11986, 2022.

\bibitem{loshchilov2016sgdr}
Ilya Loshchilov and Frank Hutter.
\newblock Sgdr: Stochastic gradient descent with warm restarts.
\newblock {\em arXiv preprint}, 2016.

\bibitem{lowe2004distinctive}
David~G Lowe.
\newblock Distinctive image features from scale-invariant keypoints.
\newblock {\em IJCV}, 60:91--110, 2004.

\bibitem{lugmayr2020srflow}
Andreas Lugmayr, Martin Danelljan, Luc~Van Gool, and Radu Timofte.
\newblock Srflow: Learning the super-resolution space with normalizing flow.
\newblock In {\em ECCV}, pages 715--732. Springer, 2020.

\bibitem{ma2015perceptual}
Kede Ma, Kai Zeng, and Zhou Wang.
\newblock Perceptual quality assessment for multi-exposure image fusion.
\newblock {\em IEEE TIP}, 24(11):3345--3356, 2015.

\bibitem{martin2001database}
David Martin, Charless Fowlkes, Doron Tal, and Jitendra Malik.
\newblock A database of human segmented natural images and its application to evaluating segmentation algorithms and measuring ecological statistics.
\newblock In {\em ICCV}, volume~2, pages 416--423, 2001.

\bibitem{ozdenizci2023}
Ozan \"{O}zdenizci and Robert Legenstein.
\newblock Restoring vision in adverse weather conditions with patch-based denoising diffusion models.
\newblock {\em IEEE TPAMI}, pages 1--12, 2023.

\bibitem{ren2020lr3m}
Xutong Ren, Wenhan Yang, Wen-Huang Cheng, and Jiaying Liu.
\newblock {LR3M}: Robust low-light enhancement via low-rank regularized retinex model.
\newblock {\em IEEE TIP}, 29:5862--5876, 2020.

\bibitem{rombach2022high}
Robin Rombach, Andreas Blattmann, Dominik Lorenz, Patrick Esser, and Bj{\"o}rn Ommer.
\newblock High-resolution image synthesis with latent diffusion models.
\newblock In {\em CVPR}, pages 10684--10695, 2022.

\bibitem{ronneberger2015u}
Olaf Ronneberger, Philipp Fischer, and Thomas Brox.
\newblock U-net: Convolutional networks for biomedical image segmentation.
\newblock In {\em MICCAI}, pages 234--241, 2015.

\bibitem{saharia2022photorealistic}
Chitwan Saharia, William Chan, Saurabh Saxena, Lala Li, Jay Whang, Emily Denton, Seyed Kamyar~Seyed Ghasemipour, Burcu~Karagol Ayan, S~Sara Mahdavi, Rapha~Gontijo Lopes, et~al.
\newblock Photorealistic text-to-image diffusion models with deep language understanding.
\newblock {\em arXiv preprint}, 2022.

\bibitem{saharia2022image}
Chitwan Saharia, Jonathan Ho, William Chan, Tim Salimans, David~J Fleet, and Mohammad Norouzi.
\newblock Image super-resolution via iterative refinement.
\newblock {\em IEEE TPAMI}, 2022.

\bibitem{Seitzer2020FID}
Maximilian Seitzer.
\newblock {pytorch-fid: FID Score for PyTorch}, 2020.

\bibitem{sohl2015deep}
Jascha Sohl-Dickstein, Eric Weiss, Niru Maheswaranathan, and Surya Ganguli.
\newblock Deep unsupervised learning using nonequilibrium thermodynamics.
\newblock In {\em ICML}, pages 2256--2265, 2015.

\bibitem{songdenoising}
Jiaming Song, Chenlin Meng, and Stefano Ermon.
\newblock Denoising diffusion implicit models.
\newblock In {\em ICLR}, 2021.

\bibitem{song2019generative}
Yang Song and Stefano Ermon.
\newblock Generative modeling by estimating gradients of the data distribution.
\newblock {\em NeurIPS}, 32, 2019.

\bibitem{song2020score}
Yang Song, Jascha Sohl-Dickstein, Diederik~P Kingma, Abhishek Kumar, Stefano Ermon, and Ben Poole.
\newblock Score-based generative modeling through stochastic differential equations.
\newblock {\em arXiv preprint arXiv:2011.13456}, 2020.

\bibitem{stark2000adaptive}
J~Alex Stark.
\newblock Adaptive image contrast enhancement using generalizations of histogram equalization.
\newblock {\em IEEE TIP}, 9(5):889--896, 2000.

\bibitem{tai2017image}
Ying Tai, Jian Yang, and Xiaoming Liu.
\newblock Image super-resolution via deep recursive residual network.
\newblock In {\em CVPR}, pages 3147--3155, 2017.

\bibitem{timofte2014a+}
Radu Timofte, Vincent De~Smet, and Luc Van~Gool.
\newblock A+: Adjusted anchored neighborhood regression for fast super-resolution.
\newblock In {\em ACCV}, pages 111--126, 2014.

\bibitem{vaswani2017attention}
Ashish Vaswani, Noam Shazeer, Niki Parmar, Jakob Uszkoreit, Llion Jones, Aidan~N Gomez, {\L}ukasz Kaiser, and Illia Polosukhin.
\newblock Attention is all you need.
\newblock {\em NeurIPS}, 30, 2017.

\bibitem{wang2021realesrgan}
Xintao Wang, Liangbin Xie, Chao Dong, and Ying Shan.
\newblock Real-esrgan: Training real-world blind super-resolution with pure synthetic data.
\newblock In {\em ICCV Workshops}.

\bibitem{wang2021real}
Xintao Wang, Liangbin Xie, Chao Dong, and Ying Shan.
\newblock Real-esrgan: Training real-world blind super-resolution with pure synthetic data.
\newblock In {\em ICCV}, pages 1905--1914, 2021.

\bibitem{wang2018esrgan}
Xintao Wang, Ke Yu, Shixiang Wu, Jinjin Gu, Yihao Liu, Chao Dong, Yu Qiao, and Chen Change~Loy.
\newblock Esrgan: Enhanced super-resolution generative adversarial networks.
\newblock In {\em ECCV Workshops}, 2018.

\bibitem{wang2021low}
Yufei Wang, Renjie Wan, Wenhan Yang, Haoliang Li, Lap-Pui Chau, and Alex Kot.
\newblock Low-light image enhancement with normalizing flow.
\newblock In {\em AAAI}, volume~36, pages 2604--2612, 2022.

\bibitem{Chen2018Retinex}
Chen Wei, Wenjing Wang, Wenhan Yang, and Jiaying Liu.
\newblock Deep retinex decomposition for low-light enhancement.
\newblock In {\em BMVC}, 2018.

\bibitem{wei2020component}
Pengxu Wei, Ziwei Xie, Hannan Lu, Zongyuan Zhan, Qixiang Ye, Wangmeng Zuo, and Liang Lin.
\newblock Component divide-and-conquer for real-world image super-resolution.
\newblock In {\em ECCV}, pages 101--117, 2020.

\bibitem{whang2022deblurring}
Jay Whang, Mauricio Delbracio, Hossein Talebi, Chitwan Saharia, Alexandros~G Dimakis, and Peyman Milanfar.
\newblock Deblurring via stochastic refinement.
\newblock In {\em CVPR}, pages 16293--16303, 2022.

\bibitem{yang2010image}
Jianchao Yang, John Wright, Thomas~S Huang, and Yi Ma.
\newblock Image super-resolution via sparse representation.
\newblock {\em IEEE TIP}, 19(11):2861--2873, 2010.

\bibitem{yang2022diffusion}
Ling Yang, Zhilong Zhang, Yang Song, Shenda Hong, Runsheng Xu, Yue Zhao, Yingxia Shao, Wentao Zhang, Bin Cui, and Ming-Hsuan Yang.
\newblock Diffusion models: A comprehensive survey of methods and applications.
\newblock {\em arXiv preprint arXiv:2209.00796}, 2022.

\bibitem{Yang_2020_CVPR}
Wenhan Yang, Shiqi Wang, Yuming Fang, Yue Wang, and Jiaying Liu.
\newblock From fidelity to perceptual quality: A semi-supervised approach for low-light image enhancement.
\newblock In {\em CVPR}, June 2020.

\bibitem{Zamir2022MIRNetv2}
Syed~Waqas Zamir, Aditya Arora, Salman Khan, Munawar Hayat, Fahad~Shahbaz Khan, Ming-Hsuan Yang, and Ling Shao.
\newblock Learning enriched features for fast image restoration and enhancement.
\newblock {\em IEEE TPAMI}, 2022.

\bibitem{zeyde2010single}
Roman Zeyde, Michael Elad, and Matan Protter.
\newblock On single image scale-up using sparse-representations.
\newblock In {\em ICCS}, pages 711--730, 2010.

\bibitem{zhang2021designing}
Kai Zhang, Jingyun Liang, Luc Van~Gool, and Radu Timofte.
\newblock Designing a practical degradation model for deep blind image super-resolution.
\newblock In {\em ICCV}, pages 4791--4800, 2021.

\bibitem{zhang2020deep}
Kai Zhang, Luc Van~Gool, and Radu Timofte.
\newblock Deep unfolding network for image super-resolution.
\newblock In {\em CVPR}, pages 3217--3226, 2020.

\bibitem{zhang2018perceptual}
Richard Zhang, Phillip Isola, Alexei~A Efros, Eli Shechtman, and Oliver Wang.
\newblock The unreasonable effectiveness of deep features as a perceptual metric.
\newblock In {\em CVPR}, 2018.

\bibitem{zhang2019zoom}
Xuaner Zhang, Qifeng Chen, Ren Ng, and Vladlen Koltun.
\newblock Zoom to learn, learn to zoom.
\newblock In {\em CVPR}, pages 3762--3770, 2019.

\bibitem{zhang2018residual}
Yulun Zhang, Yapeng Tian, Yu Kong, Bineng Zhong, and Yun Fu.
\newblock Residual dense network for image super-resolution.
\newblock In {\em CVPR}, pages 2472--2481, 2018.

\bibitem{zhang2019kindling}
Yonghua Zhang, Jiawan Zhang, and Xiaojie Guo.
\newblock Kindling the darkness: A practical low-light image enhancer.
\newblock In {\em ACM Multimedia}, pages 1632--1640, 2019.

\end{thebibliography}
	}

\clearpage 
\appendix

\begin{appendix}

%%%%%%%%% ABSTRACT
\section*{Appendix}
\label{sec: Overview}
This supplementary document provides a more comprehensive analysis of our proposed dataset and method. We present the training approach in Section \ref{sec: training approach} and the detailed network architecture in Section \ref{sec: Network Architecture}. In Section \ref{sec: Limitation Analysis}, we analyze the limitations of our proposed method. While in Section \ref{sec: Analysis on Cascaded Methods and Joint Methods}, we analyze cascaded methods and joint methods. We also provide more statistical analysis of our proposed SRRIIE dataset in Section \ref{sec: statistical analysis on the dataset}. Finally, we demonstrate more qualitative visual results on various scenes and compare our method with state-of-the-art algorithms in Section \ref{sec: More Experimental Results}. The results shown in this supplementary material are best viewed on a high-resolution display.

\section{Details of the Training Approach}
\label{sec: training approach}
The variational inference objective \cite{ho2020denoising, dhariwal2021diffusion, songdenoising} is employed to train the Gaussian denoisers utilized in Eq. (4) of the manuscript within diffusion probabilistic models:
%\vspace{-1.5mm}
% \begin{equation}
% 	\label{eq: non-Markovian_1}
% 	L_{\theta} = \mathbb{E}_{q_{\sigma}} [\underbrace{D_{KL}({q_{\sigma}}(x_T\, |\, x_0)\, ||\, p_{\theta}(x_T))}_{L_{\theta}^T} - \underbrace{\log p_{\theta}(x_0\, |\, x_1)}_{L_{\theta}^0} + \sum_{t=2}^T \underbrace{D_{KL} ({q_{\sigma}}(x_{t-1}\, |\, x_t,x_0)\, ||\, p_{\theta}^{(t)} (x_{t-1}\, |\, x_t))}_{L_{\theta}^{t-1}}],
% 	\vspace{-1.5mm}
% \end{equation}
\begin{align}
	\label{eq: non-Markovian_1}
	L_{\theta} = \mathbb{E}_{q_{\sigma}} [&\underbrace{D_{KL}({q_{\sigma}}(x_T\, |\, x_0)\, ||\, p_{\theta}(x_T))}_{L_{\theta}^T} - \underbrace{\log p_{\theta}(x_0\, |\, x_1)}_{L_{\theta}^0} \nonumber \\
	&+ \sum_{t=2}^T \underbrace{D_{KL} ({q_{\sigma}}(x_{t-1}\, |\, x_t,x_0)\, ||\, p_{\theta}^{(t)} (x_{t-1}\, |\, x_t))}_{L_{\theta}^{t-1}}], \vspace{-1.5mm}
\end{align}
where $L_{\theta}^{t-1}$ could be simplified using the alternative reparameterization that replace $\mu_{\theta}$ with $\epsilon_{\theta}$ and employs a re-weighted optimization strategy \cite{ho2020denoising, songdenoising}:
\begin{equation}
\label{eq: DPM3}
L_{\theta}^{(t-1)} = \mathbb{E}_{x_0, \epsilon_t, t} \, [||\epsilon_t - \epsilon_{\theta}(\sqrt{\alpha_t}x_0 + \sqrt{1-\alpha_t}\epsilon_t, t)||^2],
\vspace{-1mm}
\end{equation} 
where $\epsilon_t \sim \mathcal{N}(\mathbf{0}, \mathbf{I})$. We introduce two novel conditions $\pi{(\widetilde{x})}$ and $\bar{\mu}_{\theta}^{(t)}$ for the reverse generation process described in Eq. (10) of the manuscript. Using these conditions, we develop new Gaussian denoisers, denoted as $\hat{\mu}{\theta} (x_t, \pi{(\widetilde{x})}, \bar{\mu}{\theta}^{(t+1)}, t)$, to estimate the value of $x{t-1}$. To update Eq. (\ref{eq: DPM3}), we apply a reparameterization technique similar to that used for $\hat{\epsilon}_{\theta} (x_t, \pi{(\widetilde{x})}, \bar{\mu}_{\theta}^{(t+1)})$:
%
% \begin{equation}
% \label{eq: DPM3}
% L_{\theta}^{(t-1)} = \mathbb{E}_{x_0, \pi{(\widetilde{x})}, \epsilon_t, t} \, [||\epsilon_t - \hat{\epsilon}_{\theta}(\sqrt{\alpha_t}x_0 + \sqrt{1-\alpha_t}\epsilon_t, \pi{(\widetilde{x})}, \bar{\mu}_{\theta}^{(t+1)}, t)||^2],
% \vspace{-1mm}
% \end{equation} 
\begin{align}
\label{eq: DPM3}
L_{\theta}^{(t-1)} = \mathbb{E}_{x_0, \pi{(\widetilde{x})}, \epsilon_t, t} \, [&||\epsilon_t - \hat{\epsilon}_{\theta}(\sqrt{\alpha_t}x_0 + \sqrt{1-\alpha_t}\epsilon_t, \nonumber \\
&\pi{(\widetilde{x})}, \bar{\mu}_{\theta}^{(t+1)}, t)||^2], \vspace{-1mm}
\end{align}
In Section 3.1 of the manuscript, we provide 400 paired low-high quality images for super-resolving real-world image illumination enhancement tasks. Each sequence consists of 9 low-light images, serving as low-quality images, and 3 high-resolution images, serving as the ground truth high-quality images. We use the 9 low-light images to represent the condition $\pi{(\widetilde{x})}$, while the 3 high-resolution images defined in Eq. (2) represent $x_0$. We jointly train the denoisers $\{\hat{\epsilon}_{\theta}(\sqrt{\alpha_t}x_0 + \sqrt{1-\alpha_t}\epsilon_t, \pi{(\widetilde{x})}, \bar{\mu}_{\theta}^{(t+1)}, t)\}_{t=0}^T$ with modern deep learning techniques.

\section{Network Architecture}
\label{sec: Network Architecture}

We describe the detailed network architecture for $\{\hat{\epsilon}_{\theta}(\sqrt{\alpha_t}x_0 + \sqrt{1-\alpha_t}\epsilon_t, \pi{(\widetilde{x})}, \bar{\mu}_{\theta}^{(t+1)}, t)\}_{t=0}^T$. The deep networks share the same network architecture weights parameters across different sampling $t$ during training and test.
As summarized in Table \ref{tab:U-Net architecture}, the proposed denoiser is based on a U-Net architecture motivated by prior works \cite{ho2020denoising, songdenoising, bansal2022cold}. This network incorporates downsampling and upsampling paths using ConvNext blocks \cite{liu2022convnet} and residual blocks \cite{he2016deep}, while employing a sinusoidal positional embedding operation \cite{dai2019transformer} for time embedding. 
As summarized in Table \ref{tab:ConvNextBlock architecture}, the ConvNext block architecture includes a time multi-layer perceptron (MLP) and a convolutional network. And the ConvNext block architecture features the Gaussian error linear unit (GELU) activation function \cite{hendrycks2016gaussian} and layer normalization (LayerNorm) \cite{ba2016layer}. 
As summarized in Table \ref{tab:ResidualBlock architecture}, the residual block architecture utilizes pre-layer normalization \cite{ba2016layer} and linear attention \cite{vaswani2017attention} components. The downsampling path uses convolutional layers, while the upsampling path employs transposed convolutional layers \cite{dumoulin2016guide}. 
The network processing Raw images before the diffusion model shares a similar architecture to the denoising U-Net, except for three variations: 4 initial input channels, 3 final output channels, and channels 16 times smaller than the denoising U-Net, without incorporating time embedding.

\begin{table*}[htbp]
    \centering
        \caption{Proposed denosing U-Net architecture.}
        \vspace{1mm}
    \label{tab:U-Net architecture}
    \begin{tabular}{lcccc}
        \toprule
        
        \textbf{Layer} & \textbf{Input Channels} & \textbf{Output Channels} & \textbf{Kernel Size} & \textbf{Activation} \\
        \midrule
        \multicolumn{5}{c}{\textbf{Time Embedding Block}} \\
         \midrule
        SinusoidalPosEmb & - & 64 & - & - \\
        MLP & 64, 256  & 256, 64 & - & GELU\\
        \midrule
        \multicolumn{5}{c}{\textbf{Downsampling Blocks}} \\
        
        \midrule
        % ConvNextBlock & 128 & 256 & 1, 7, 3, 3 & LayerNorm, GELU, GELU \\
        % \midrule
        % ConvNextBlock & 256 & 256 & 1, 7, 3, 3 & LayerNorm, GELU, GELU \\
        % \midrule
        % ResidualBlock(PreNorm) & 256 & 256 & 1, 1 & LayerNorm \\
        % LinearAttention & 256 & 384, 128, 256 & 1, 1, 1, 1 & - \\
        % \midrule
        % Conv & 256 & 256 & 4, 4 & - \\
        %  \midrule
        ConvNextBlock $\times$ 2 & 9, 64 & 64, 64 & 1, 3, 7 & GELU\\
        ResidualBlock& 64 & 64 & 1 & - \\
        Conv & 64 & 64 & 4 & - \\
        \midrule
        ConvNextBlock $\times$ 2 & 64, 128 & 128, 128 & 1, 3, 7 & GELU \\
        ResidualBlock& 128 & 128 & 1 & - \\
        Conv & 128 & 128 & 4 & - \\
        \midrule
        ConvNextBlock $\times$ 2 & 128, 256 & 256, 256 & 1, 3, 7 & GELU \\
        ResidualBlock& 256 & 256 & 1 & - \\
        Conv & 256 & 256 & 4 & - \\
        \midrule
        ConvNextBlock $\times$ 2 & 256, 512 & 512, 512 & 1, 3, 7 & GELU \\
        ResidualBlock& 512 & 512 & 1 & - \\
        \midrule  
        \multicolumn{5}{c}{\textbf{Middle Block}} \\
        \midrule
        ConvNextBlock  & 512 & 512 &   3, 7 & GELU \\
        ResidualBlock& 512 & 512 & 1 & - \\
        ConvNextBlock & 512 & 512 &   3, 7 & GELU \\
        \midrule
        \multicolumn{5}{c}{\textbf{Upsampling Blocks}} \\
        \midrule
        ConvNextBlock $\times$ 2 & 1024, 256 & 256, 256 & 1, 3, 7 & GELU \\
        ResidualBlock& 256 & 256 & 1 & - \\
        Transposed Conv & 256 & 256 & 4 & - \\
        \midrule        
        ConvNextBlock $\times$ 2 & 512, 128 & 128, 128 & 1, 3, 7 & GELU \\
        ResidualBlock& 128 & 128 & 1 & - \\
        Transposed Conv & 128 & 128 & 4 & - \\
        \midrule
        ConvNextBlock $\times$ 2 & 256, 64 & 64, 64 & 1, 3, 7 & GELU \\
        ResidualBlock& 64 & 64 & 1 & - \\
        Transposed Conv & 64 & 64 & 4 & - \\
        \midrule
         \multicolumn{5}{c}{\textbf{Output Block}} \\
        \midrule
        ConvNextBlock & 64 & 64 & 3, 7 & - \\
        Conv & 64 & 3 & 1 & - \\
         % \midrule
        \bottomrule
        
    \end{tabular}
\end{table*}

\begin{table}[htbp]
    \centering
    \caption{Proposed ConvNext block architecture. The number of channels denoted by $N$ differs among blocks.}
    \vspace{1mm}
    \label{tab:ConvNextBlock architecture}
    \begin{tabular}{lcc}
        \toprule
        \textbf{Layer} & \textbf{Channels} & \textbf{Kernel Size} \\
        \midrule
         \multicolumn{3}{c}{\textbf{Time MLP}} \\
        \midrule
        GELU & - & - \\
        Linear & $N$ $\to$ 4$\times$$N$ & - \\
        \midrule
         \multicolumn{3}{c}{\textbf{Conv Net}} \\
        \midrule
        Conv & 4$\times$$N$ $\to$ 4$\times$$N$ &  7 \\
        LayerNorm & - & - \\
        Conv & 4$\times$$N$ $\to$ 8$\times$$N$ &  3 \\
        GELU & - & - \\
        Conv & 8$\times$$N$ $\to$ 4$\times$$N$ & 3 \\
        Conv/Identity & 4$\times$$N$ $\to$ $N$/- & 1/- \\
        \bottomrule
    \end{tabular}
\end{table}

\begin{table}[htbp]
    \centering
    \caption{Proposed residual block architecture. The number of channels denoted by $N$ differs among blocks.}
    \vspace{1mm}
    \label{tab:ResidualBlock architecture}
    \begin{tabular}{lcc}
        \toprule
        \textbf{Layer} & \textbf{Channels} & \textbf{Kernel Size} \\
        \midrule
        PreNorm & - & - \\
        \midrule
         \multicolumn{3}{c}{\textbf{Linear Attention}} \\
        \midrule
        Conv (to\_qkv) & 2$\times$$N$ $\to$ 3$\times$$N$ & 1\\
        Conv (to\_out) & $N$ $\to$ 2$\times$$N$ & 1\\
        \midrule
        LayerNorm & - & - \\
        \bottomrule
    \end{tabular}
\end{table}

\section{Limitation Analysis on the Proposed Method}
\label{sec: Limitation Analysis}
Our proposed conditional diffusion probabilistic model employs default network architectures used in state-of-the-art diffusion model works \cite{songdenoising, bansal2022cold}. However, its main limitation is the longer inference time compared to one-step forward deep learning-based methods. For instance, DDPM \cite{ho2020denoising} requires 1000 rounds of computationally expensive Langevin dynamics sampling. To address the slow speed issue, DDIM \cite{songdenoising} accelerates sampling with effective deterministic implicit sampling, while Stable Diffusion \cite{rombach2022high} trains a more efficient diffusion model by embedding data into a latent space. Although these methods perform well, our proposed model takes one minute to restore a $512\times512$ image with 50 sample steps on a single NVIDIA A6000 GPU, whereas Real-ESRGAN \cite{wang2021realesrgan} only takes 0.7 seconds. Thus, there is still a need for further research to develop more efficient algorithms for diffusion probabilistic models that can be applied to a wider range of practical problems.

\section{Analysis on Cascaded Methods and Joint Methods}
\label{sec: Analysis on Cascaded Methods and Joint Methods}
\begin{figure*}[!htp]\footnotesize
	\centering
	\vspace{0mm}
	\begin{tabular}{cccccc}
	\hspace{-2.6mm}
	\includegraphics[width=0.195\linewidth]{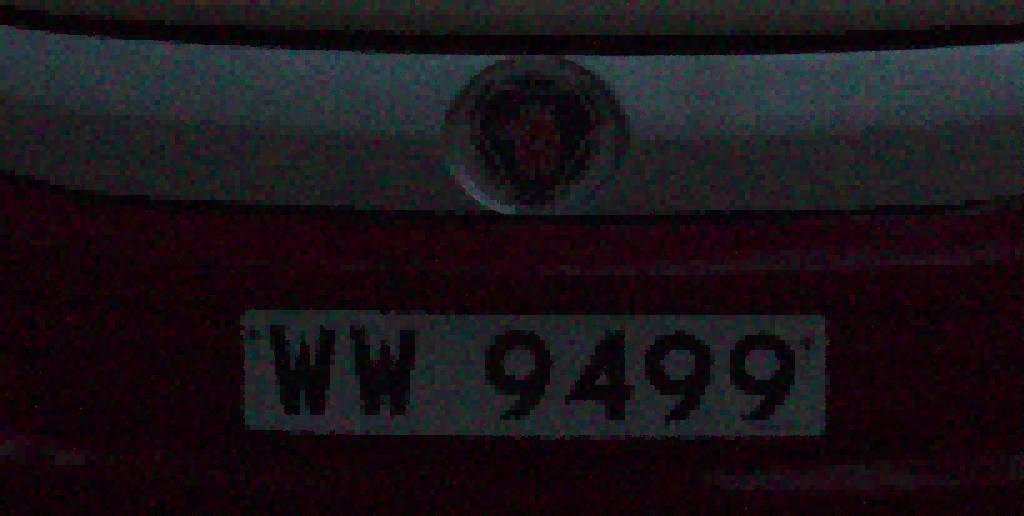} &\hspace{-4.5mm}
	\includegraphics[width=0.195\linewidth]{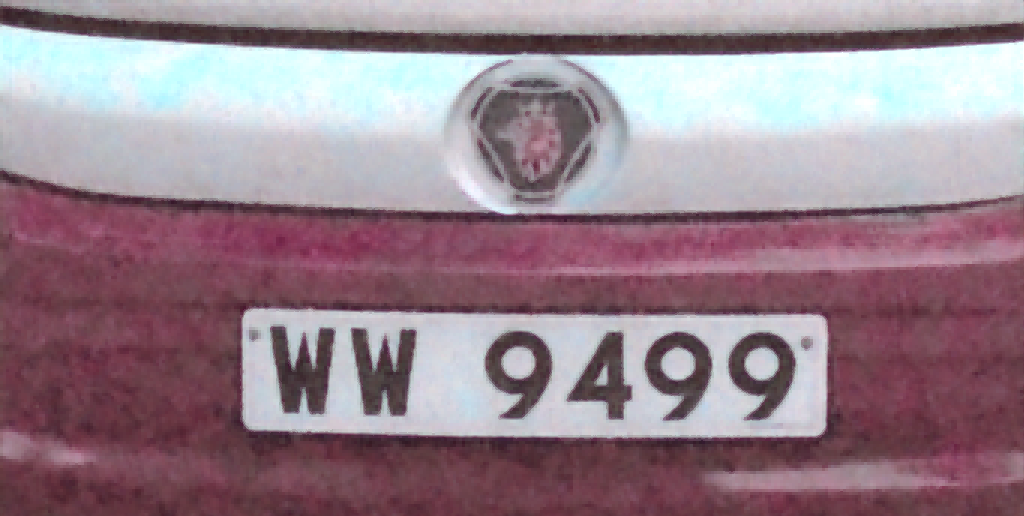} &\hspace{-4.5mm}
	\includegraphics[width=0.195\linewidth]{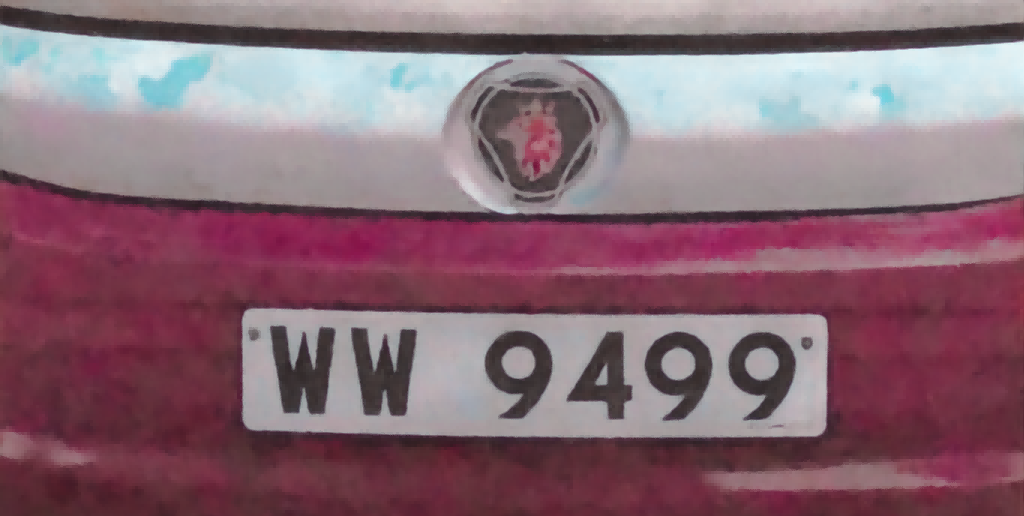} &\hspace{-4.5mm}
	\includegraphics[width=0.195\linewidth]{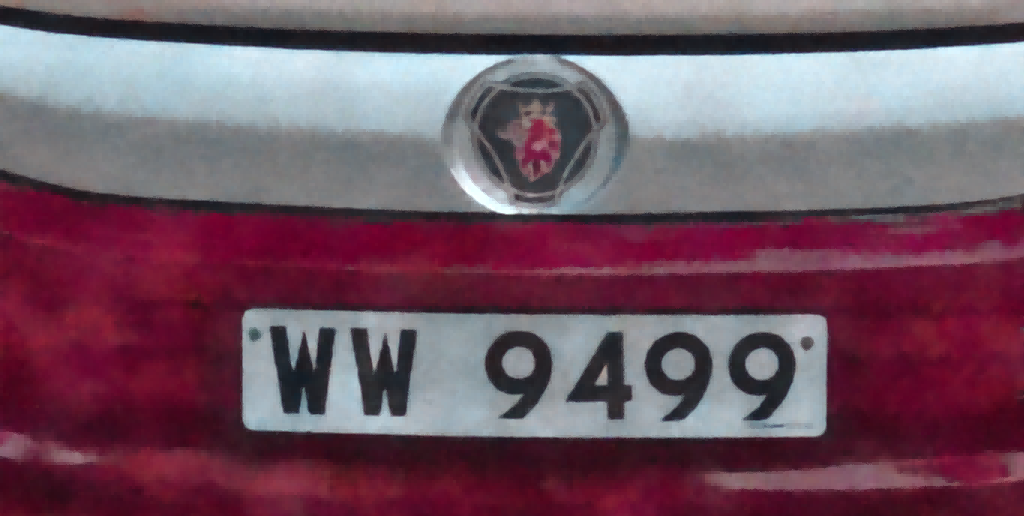} &\hspace{-4.5mm}
	\includegraphics[width=0.195\linewidth]{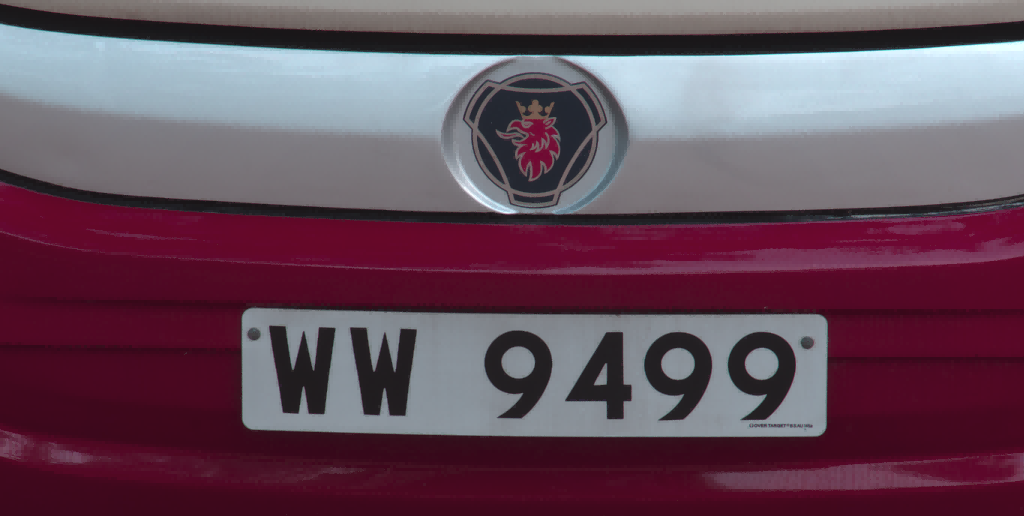}
	\\
	\hspace{-2.6mm}
	\includegraphics[width=0.195\linewidth]{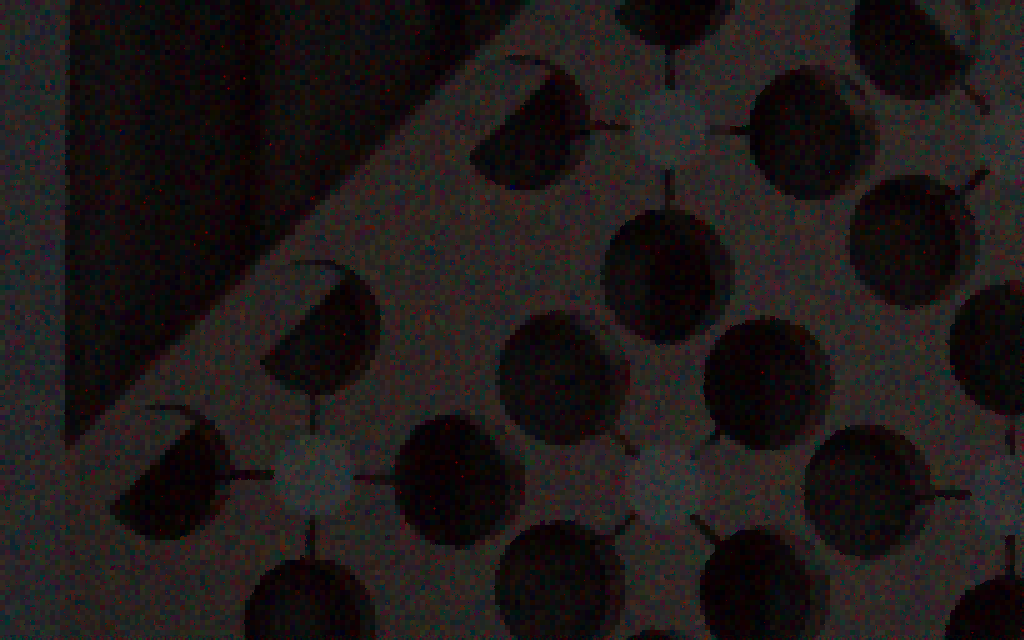} &\hspace{-4.5mm}
	\includegraphics[width=0.195\linewidth]{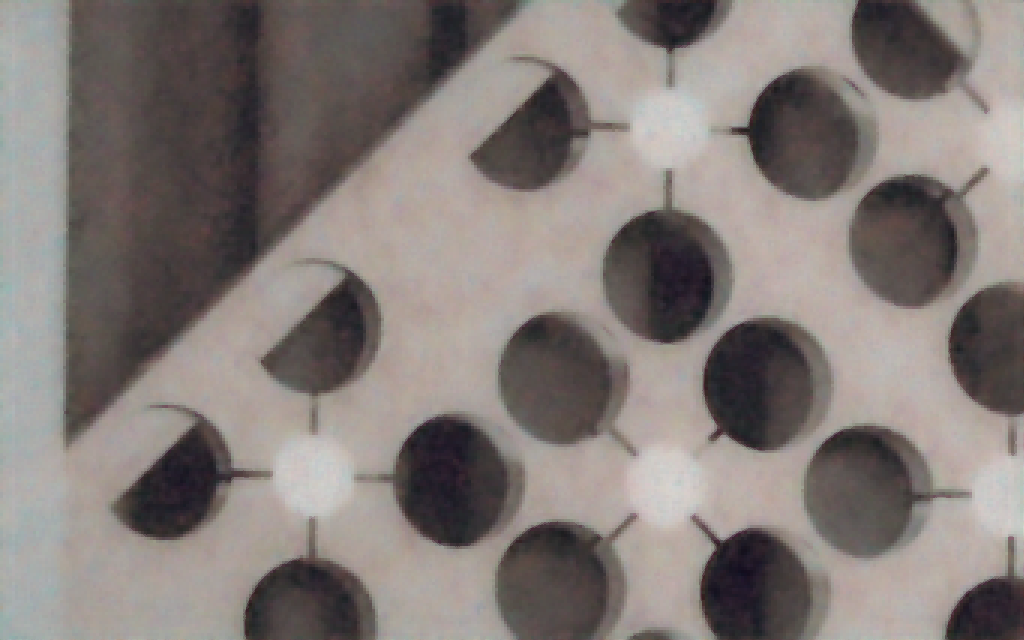} &\hspace{-4.5mm}
	\includegraphics[width=0.195\linewidth]{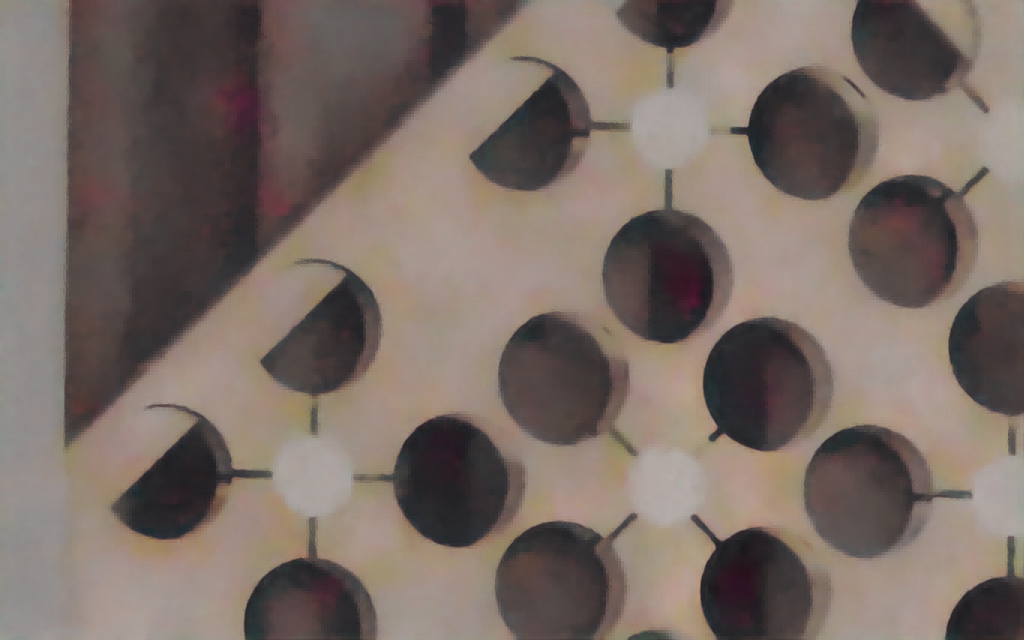} &\hspace{-4.5mm}
	\includegraphics[width=0.195\linewidth]{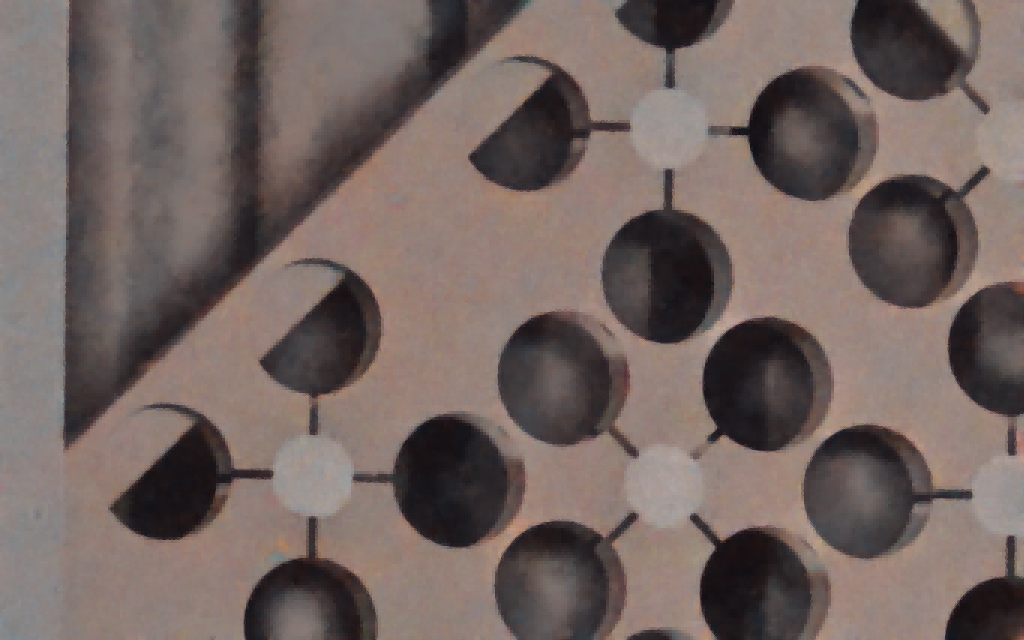} &\hspace{-4.5mm}
	\includegraphics[width=0.195\linewidth]{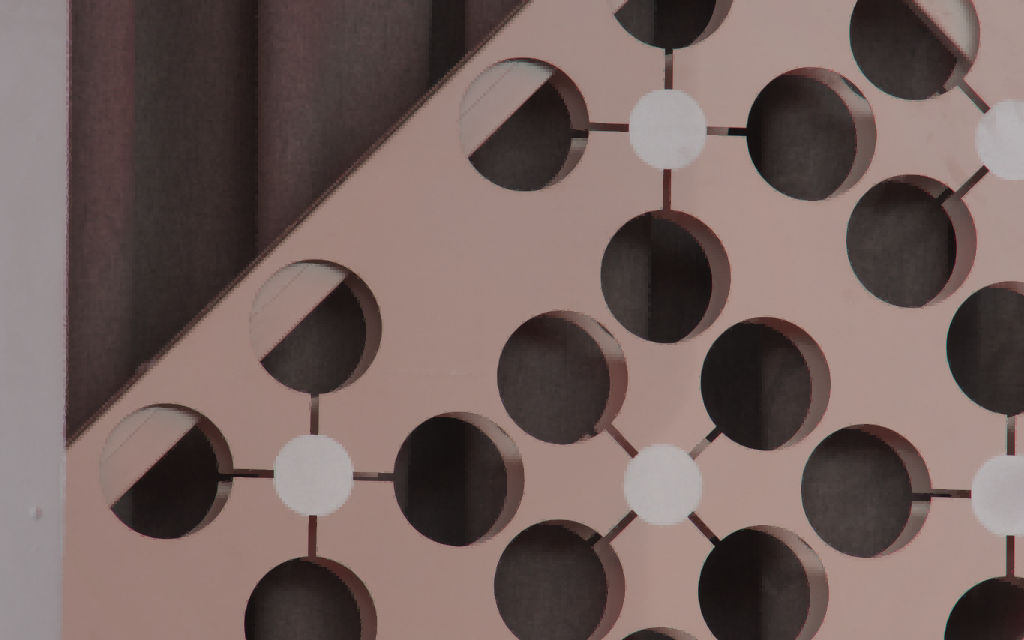}
	\\
	\hspace{-2.6mm}
	(a) Input image &\hspace{-4.5mm} (b) MIRNet-v2 \cite{Zamir2022MIRNetv2} (IE $\rightarrow$) &\hspace{-4.5mm} (c) SwinIR \cite{liang2021swinir} ($\rightarrow$ SR) &\hspace{-4.5mm} (d) SwinIR \cite{liang2021swinir} (Joint) &\hspace{-4.5mm} (e) Ground truth
	\\
	\end{tabular}
		%\end{center}
	\vspace{0mm}
	\caption{
	Visual comparisons (×4 SR) on the SRRIIE dataset for cascaded methods and joint methods. The low-quality input images are captured with -2 EV and high ISO levels, resulting in complicated real-world noises.
	The intermediate output images of the illumination enhancement model in the first stage contain artifacts and color distortions, which are further amplified in the super-resolution model in the second stage, as seen in (b) and (c). In contrast, the joint method generates more natural images, as shown in (d).
	}
	\vspace{-4mm}
	\label{fig: cascaded method and joint method}
\end{figure*}

As analyzed in Section 1 of the manuscript, we could directly use existing single image illumination enhancement methods and super-resolution methods separately in a cascaded manner for super-resolving real-world image illumination enhancement tasks. Figure \ref{fig: cascaded method and joint method} shows the example results. We first use MIRNet-v2 \cite{Zamir2022MIRNetv2} method (Figure \ref{fig: cascaded method and joint method} (b)) to enhance the input low-light image, then use SwinIR \cite{liang2021swinir} method (Figure \ref{fig: cascaded method and joint method} (c)) to super-resolve the enhanced output image by a scale of 4.
However, the artifacts and color distortions in Figure \ref{fig: cascaded method and joint method} (b) are further amplified in Figure \ref{fig: cascaded method and joint method} (c). In contrast, applying SwinIR \cite{liang2021swinir} method directly to super-resolve image illumination enhancement tasks in a joint manner can avoid these problems, as shown in Figure \ref{fig: cascaded method and joint method} (d).
Furthermore, the quantitative results in Table 2 of the manuscript indicate that the cascaded method MIRNet-v2 \cite{Zamir2022MIRNetv2} $\rightarrow$ SwinIR \cite{liang2021swinir} achieves higher PSNR values compared to using SwinIR \cite{liang2021swinir} in a joint manner, but performs less effectively in terms of other metrics.
These experiments demonstrate that cascaded methods do not always outperform joint methods. The intermediate output of the illumination enhancement model in the first stage may suffer from significant information loss or accumulated incorrect predicted information, resulting in a more ill-posed image restoration problem. Consequently, the super-resolution model in the second stage performs less effectively.
%
%This is mainly due to the over-smoothing effects in the first stage by reducing noises.
%
%Considering the biased estimations and information loss issues of cascaded methods, we suggest joint methods to be the preferential direction in the future research.

%\section{Real-world Scenes in the proposed SRRIIE Dataset}
%\label{sec: SRRIIE dataset}

%\vspace{-1mm}
\section{Statistical Analysis on the Proposed SRRIIE dataset}
\label{sec: statistical analysis on the dataset}
\begin{figure*}[!tp]
	\centering
	\vspace{0mm}
	\begin{tabular}{cccccc}
	\hspace{-2.6mm}
	\includegraphics[width=0.195\linewidth]{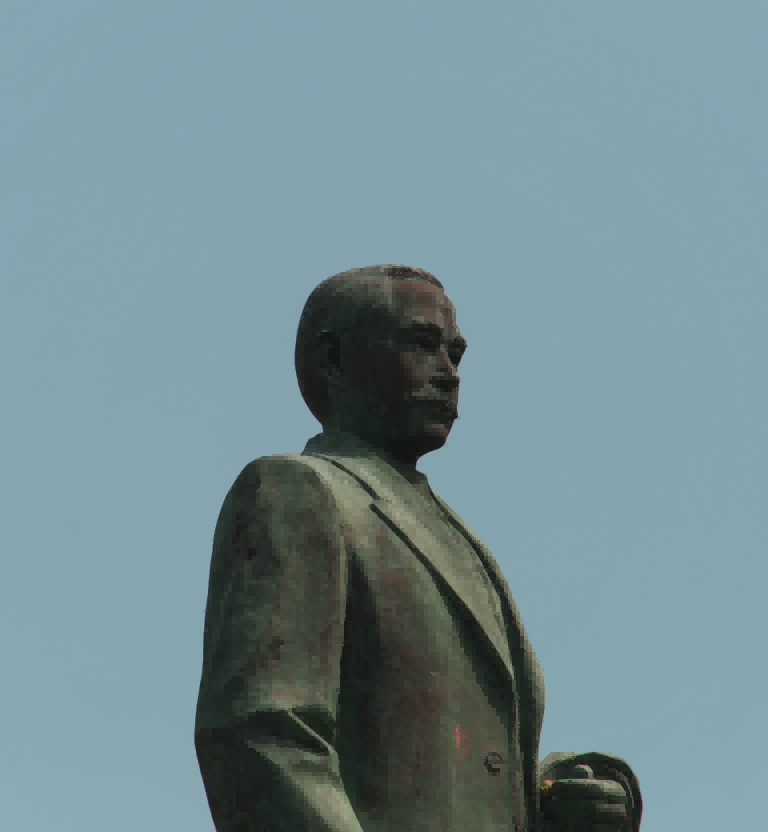} &\hspace{-4mm}
	\includegraphics[width=0.195\linewidth]{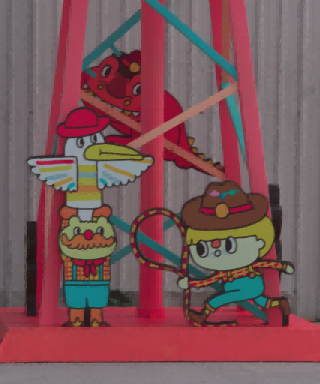} &\hspace{-4mm}
	\includegraphics[width=0.195\linewidth]{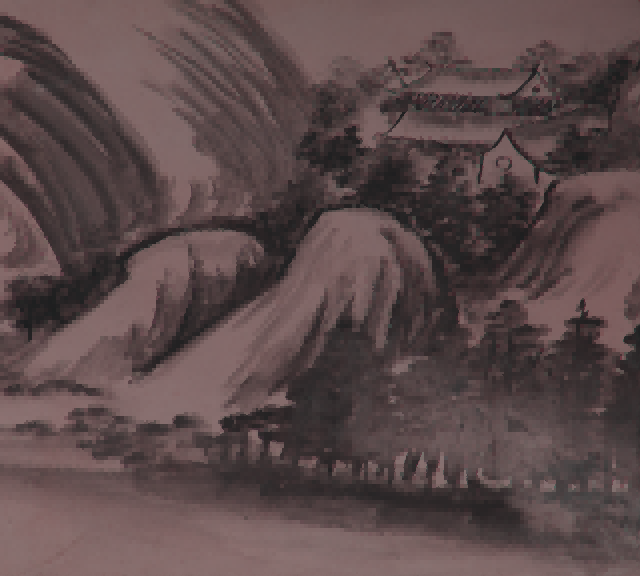} &\hspace{-4mm}
	\includegraphics[width=0.195\linewidth]{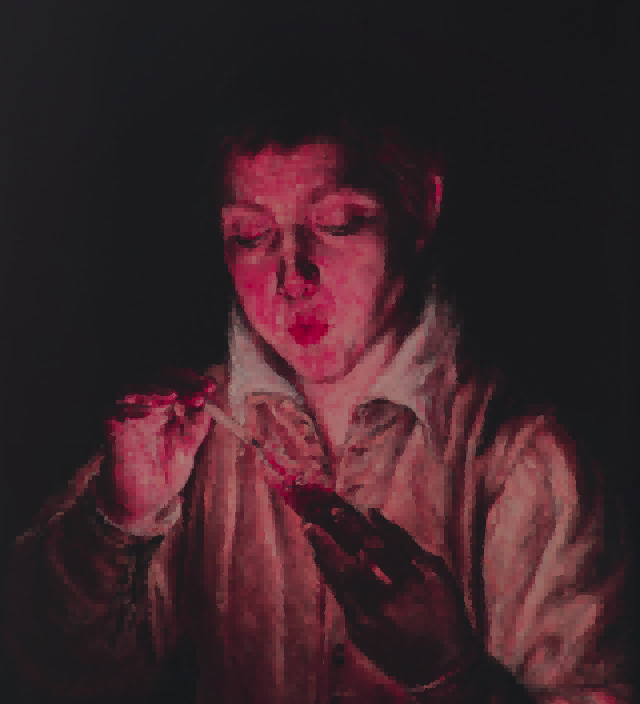} &\hspace{-4mm}
	\includegraphics[width=0.195\linewidth]{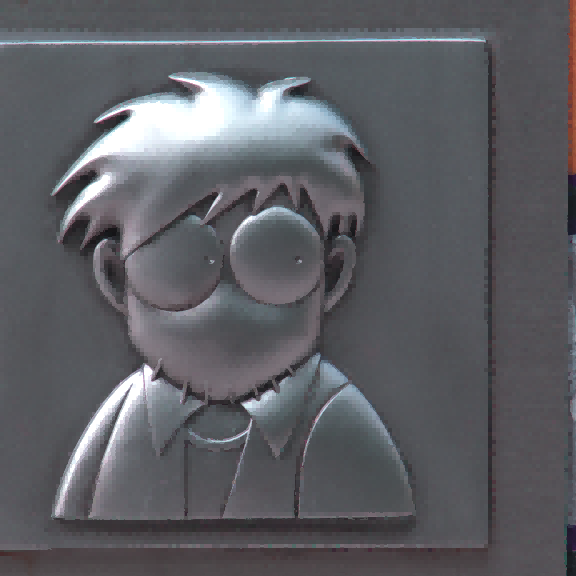}
	\\
	\hspace{-2.6mm}
	\includegraphics[width=0.195\linewidth]{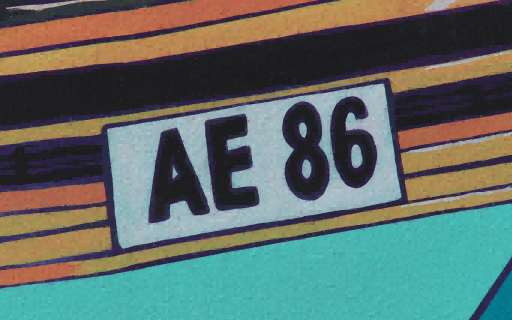} &\hspace{-4mm}
	\includegraphics[width=0.195\linewidth]{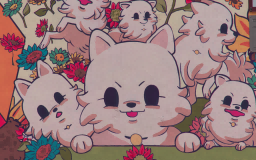} &\hspace{-4mm}
	\includegraphics[width=0.195\linewidth]{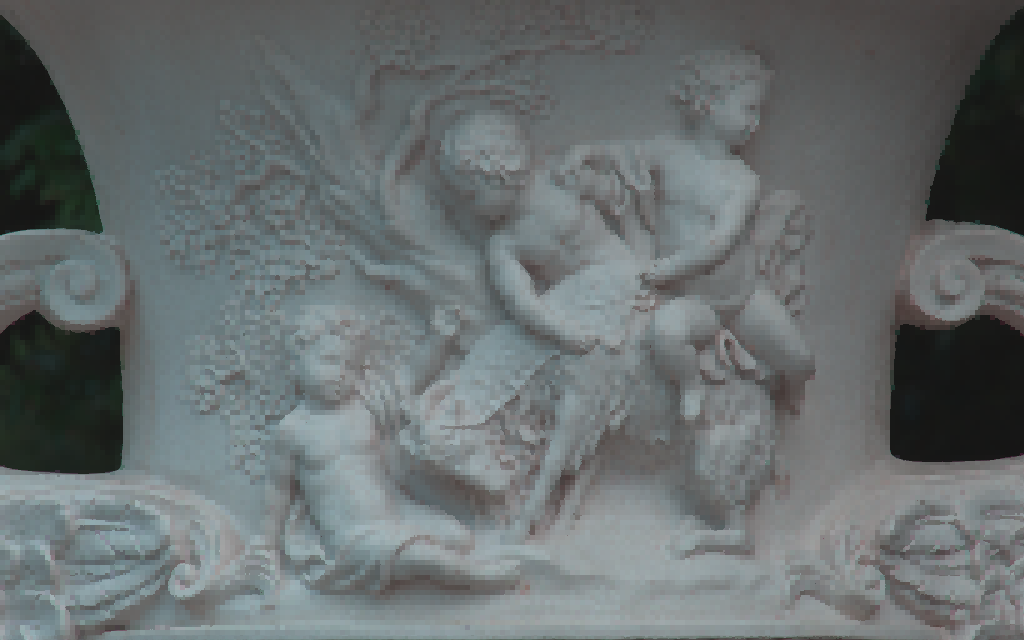} &\hspace{-4mm}
	\includegraphics[width=0.195\linewidth]{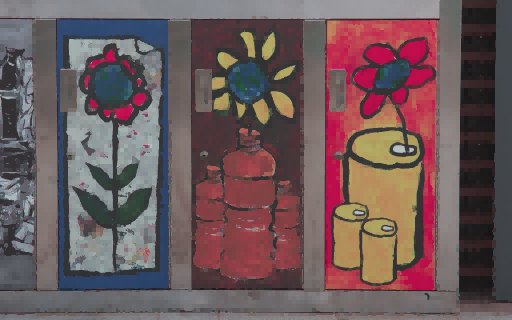} &\hspace{-4mm} 
	\includegraphics[width=0.195\linewidth]{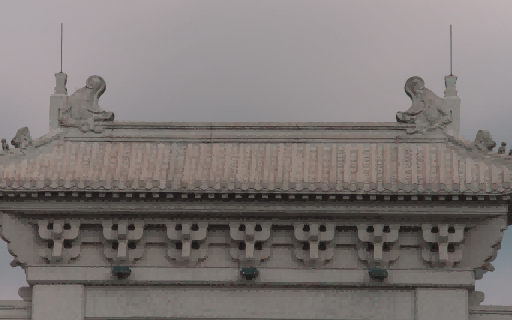}
	\\
	\hspace{-2.6mm}
	\includegraphics[width=0.195\linewidth]{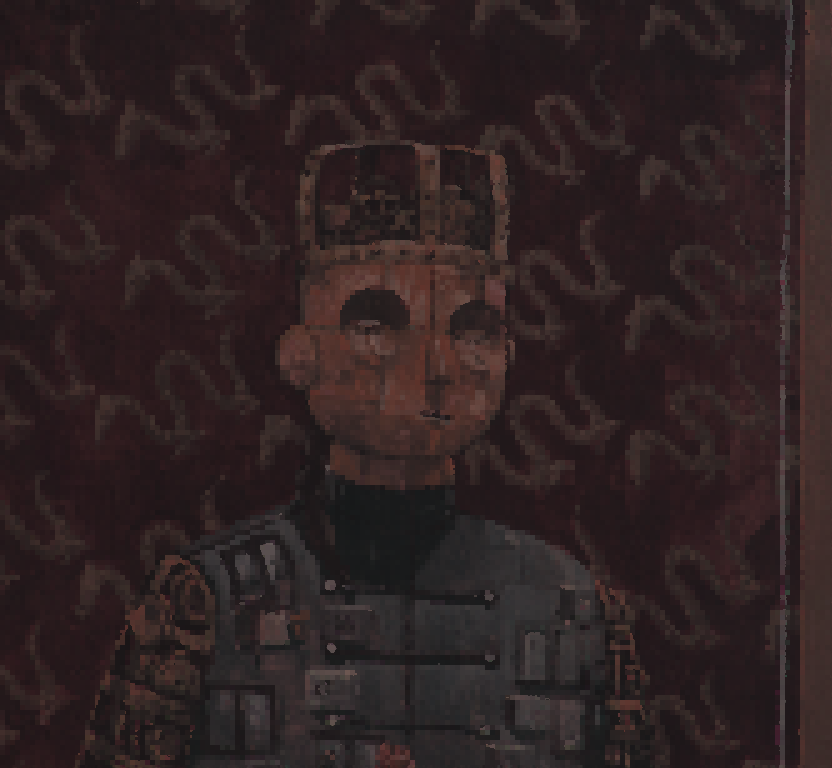} &\hspace{-4mm}
	\includegraphics[width=0.195\linewidth]{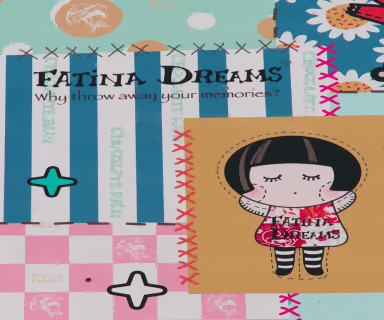} &\hspace{-4mm}
	\includegraphics[width=0.195\linewidth]{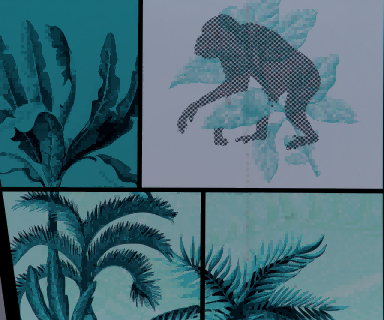} &\hspace{-4mm}
	\includegraphics[width=0.195\linewidth]{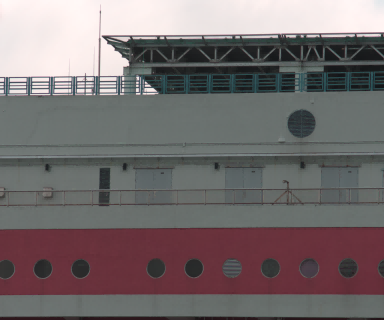} &\hspace{-4mm}
	\includegraphics[width=0.195\linewidth]{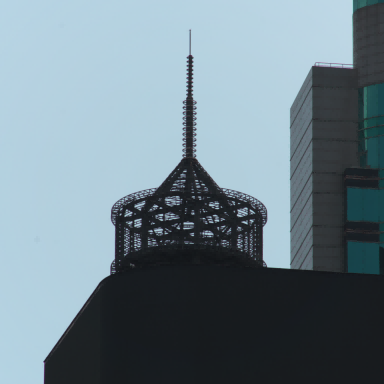}
	\\
	\hspace{-2.6mm}
	\includegraphics[width=0.195\linewidth]{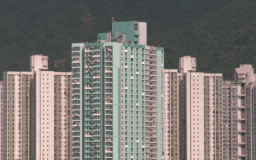} &\hspace{-4mm}
	\includegraphics[width=0.195\linewidth]{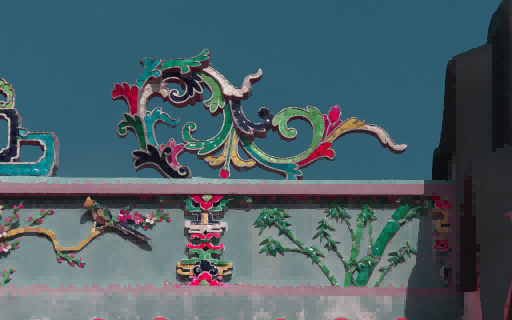} &\hspace{-4mm}
	\includegraphics[width=0.195\linewidth]{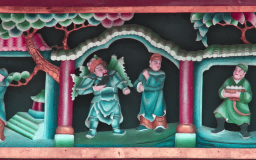} &\hspace{-4mm}
	\includegraphics[width=0.195\linewidth]{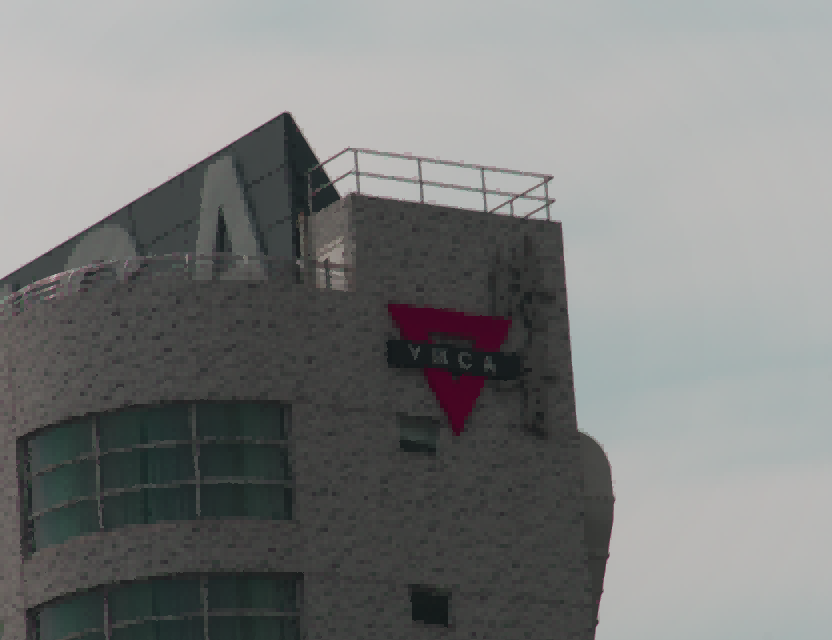} &\hspace{-4mm}
	\includegraphics[width=0.195\linewidth]{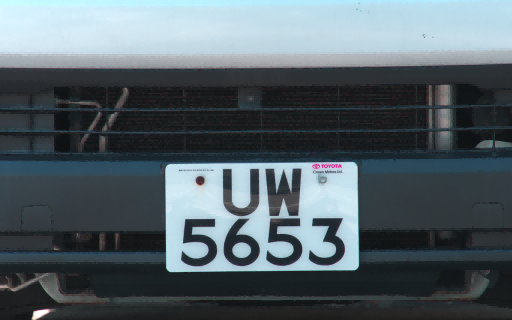}
	\\
	\hspace{-2.6mm}
	\includegraphics[width=0.195\linewidth]{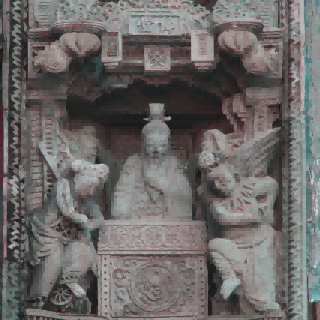} &\hspace{-4mm}
	\includegraphics[width=0.195\linewidth]{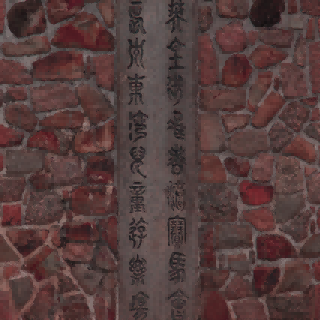} &\hspace{-4mm}
	\includegraphics[width=0.195\linewidth]{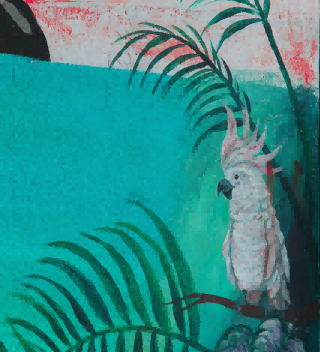} &\hspace{-4mm}
	\includegraphics[width=0.195\linewidth]{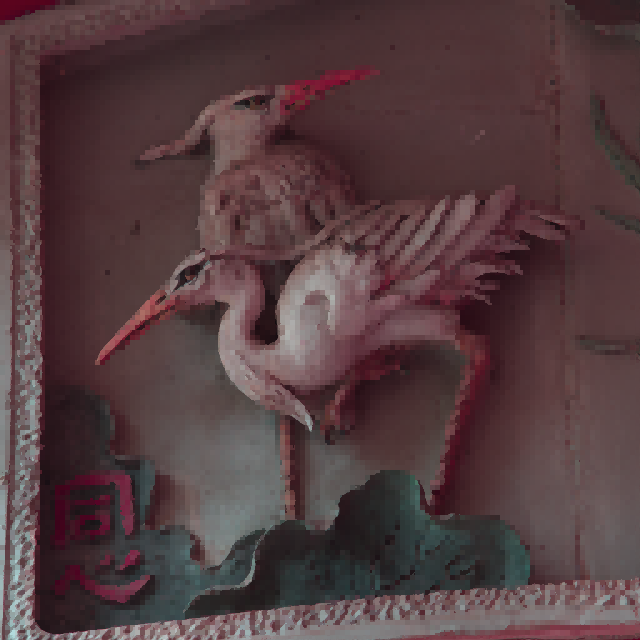} &\hspace{-4mm}
	\includegraphics[width=0.195\linewidth]{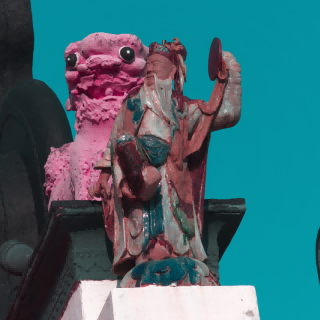}
	\\
	\hspace{-2.6mm}
	\includegraphics[width=0.195\linewidth]{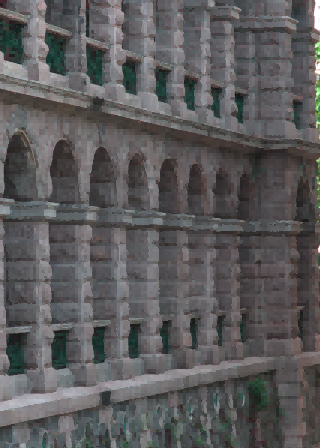} &\hspace{-4mm}
	\includegraphics[width=0.195\linewidth]{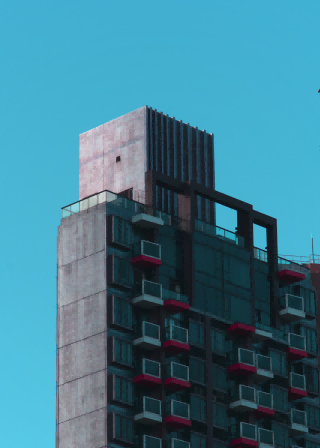} &\hspace{-4mm}
	\includegraphics[width=0.195\linewidth]{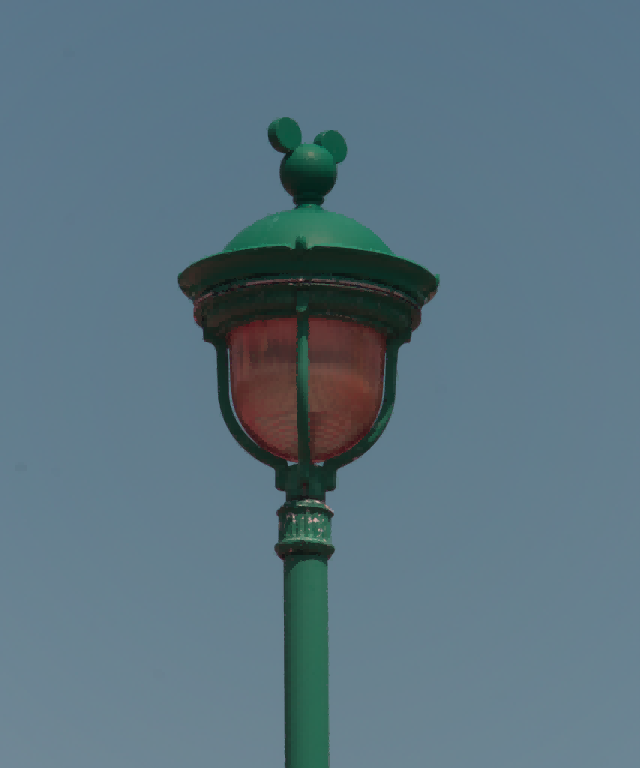} &\hspace{-4mm}
	\includegraphics[width=0.195\linewidth]{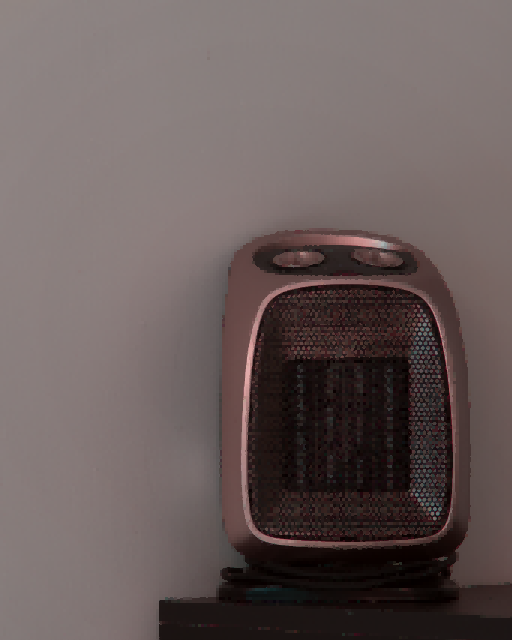} &\hspace{-4mm}
	\includegraphics[width=0.195\linewidth]{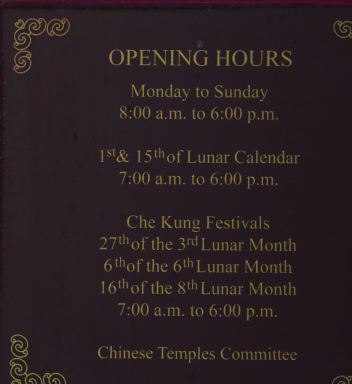}
	\\
	\end{tabular}
		%\end{center}
	%\vspace{0.5mm}
	\caption{Example normal-light scenes in the train set of our proposed SRRIIE dataset. The train set of our proposed SRRIIE dataset consists of 300 different image sequences captured from 300 non-overlapping outdoor and indoor scenes.
	}
	\vspace{-2mm}
	\label{fig: train set}
\end{figure*}

\begin{figure*}[!tp]
	\centering
	\vspace{0mm}
	\begin{tabular}{cccccc}
	\hspace{-2.6mm}
	\includegraphics[width=0.195\linewidth]{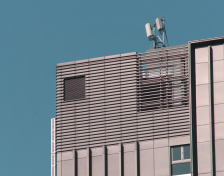} &\hspace{-4mm}
	\includegraphics[width=0.195\linewidth]{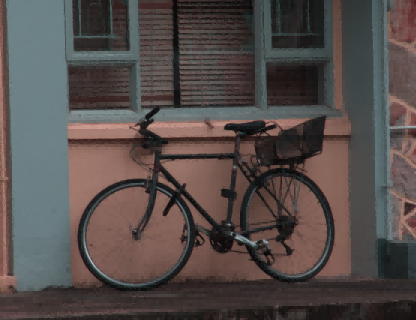} &\hspace{-4mm}
	\includegraphics[width=0.195\linewidth]{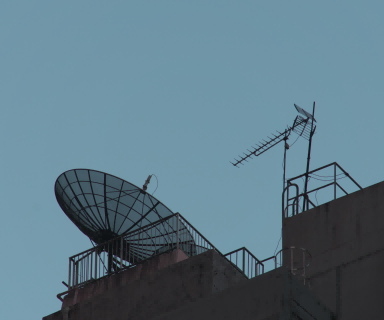} &\hspace{-4mm}
	\includegraphics[width=0.195\linewidth]{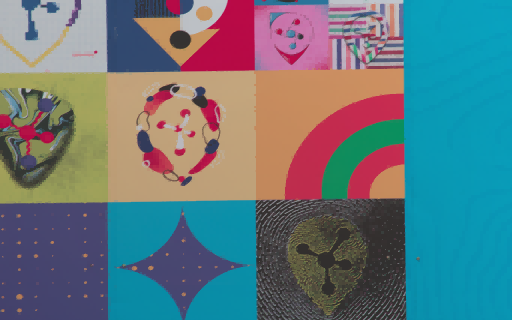} &\hspace{-4mm}
	\includegraphics[width=0.195\linewidth]{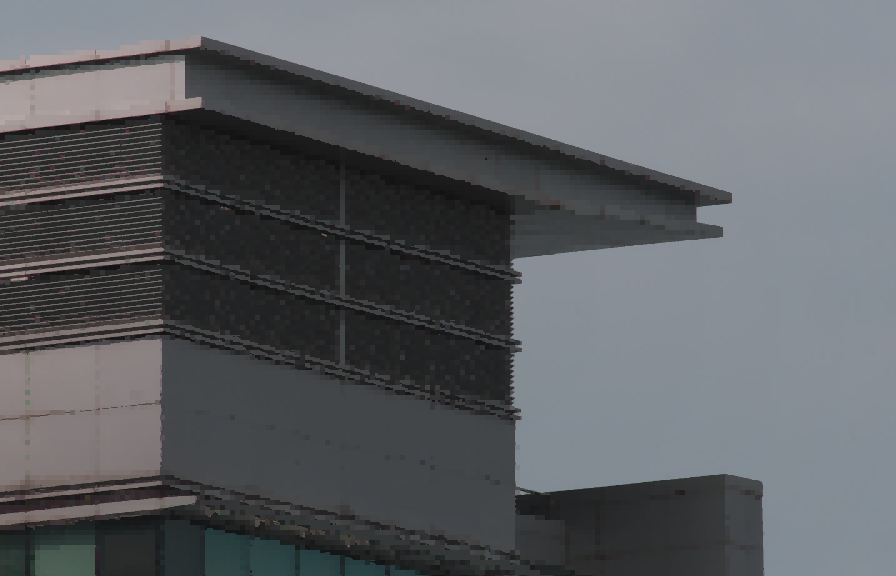}
	\\
	\hspace{-2.6mm}
	\includegraphics[width=0.195\linewidth]{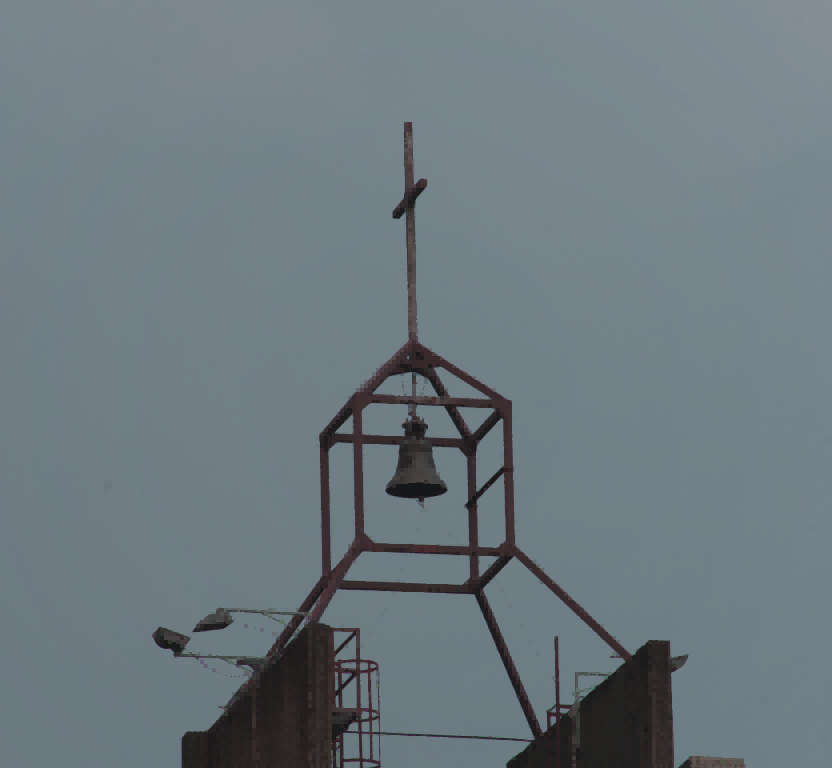} &\hspace{-4mm}
	\includegraphics[width=0.195\linewidth]{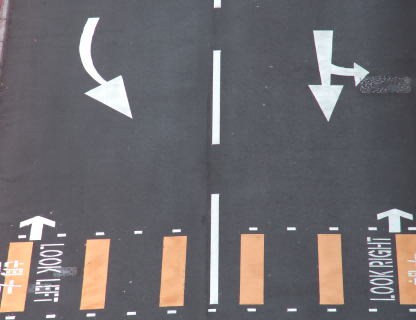} &\hspace{-4mm}
	\includegraphics[width=0.195\linewidth]{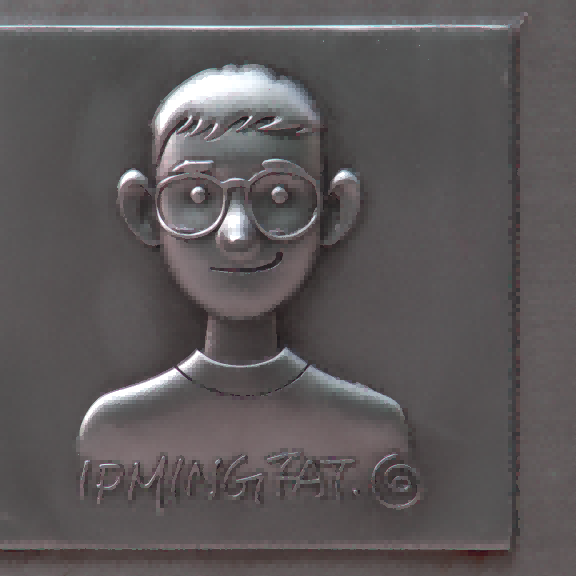} &\hspace{-4mm}
	\includegraphics[width=0.195\linewidth]{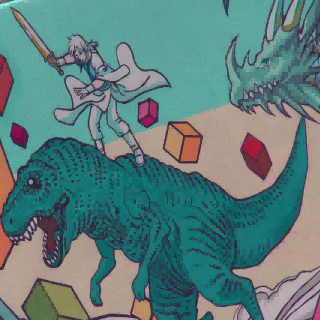} &\hspace{-4mm}
	\includegraphics[width=0.195\linewidth]{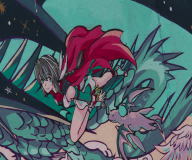}
	\\
	\hspace{-2.6mm}
	\includegraphics[width=0.195\linewidth]{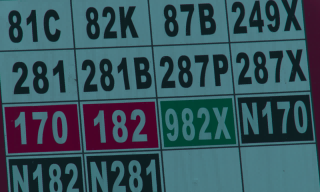} &\hspace{-4mm}
	\includegraphics[width=0.195\linewidth]{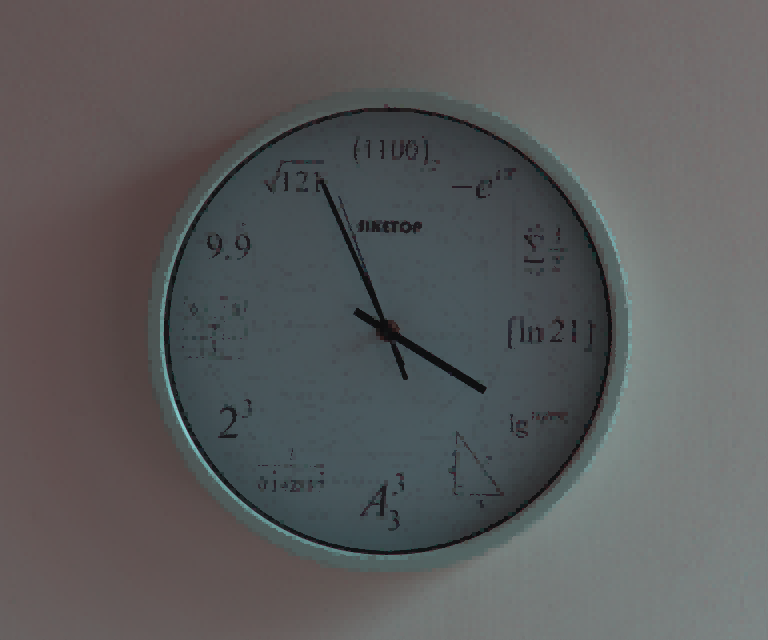} &\hspace{-4mm}
	\includegraphics[width=0.195\linewidth]{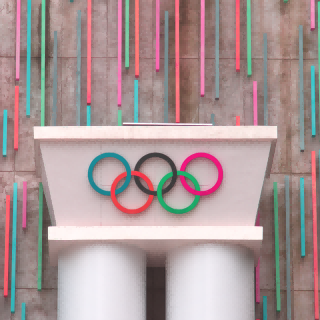} &\hspace{-4mm}
	\includegraphics[width=0.195\linewidth]{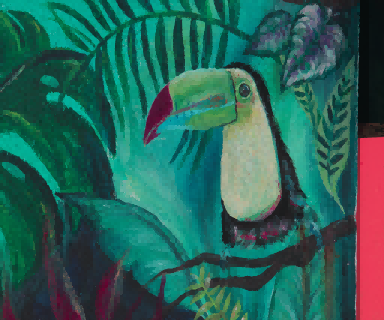} &\hspace{-4mm}
	\includegraphics[width=0.195\linewidth]{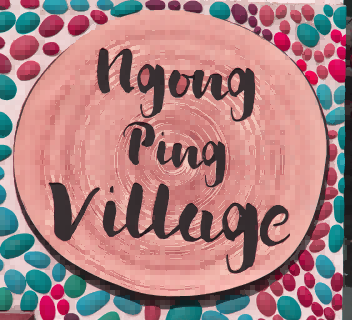}
	\\
	\hspace{-2.6mm}
	\includegraphics[width=0.195\linewidth]{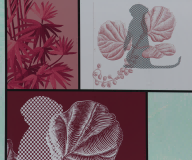} &\hspace{-4mm}
	\includegraphics[width=0.195\linewidth]{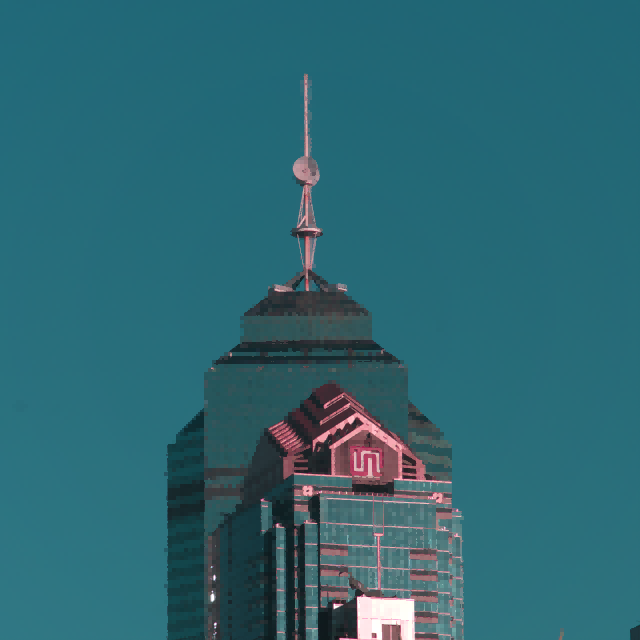} &\hspace{-4mm}
	\includegraphics[width=0.195\linewidth]{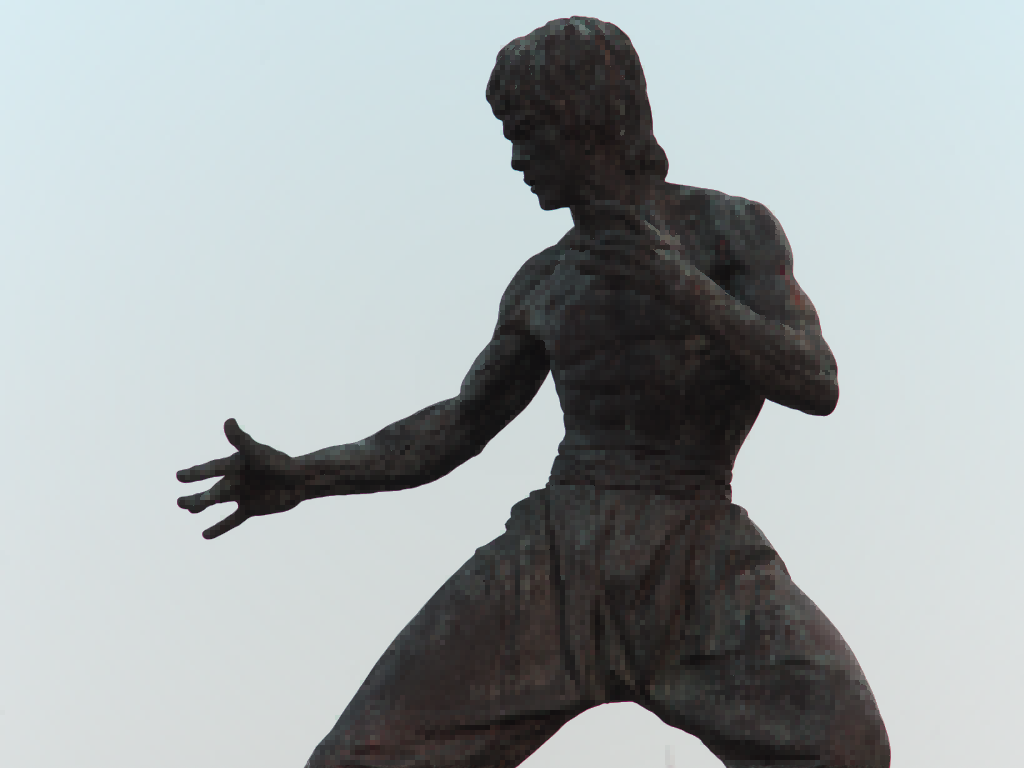} &\hspace{-4mm}
	\includegraphics[width=0.195\linewidth]{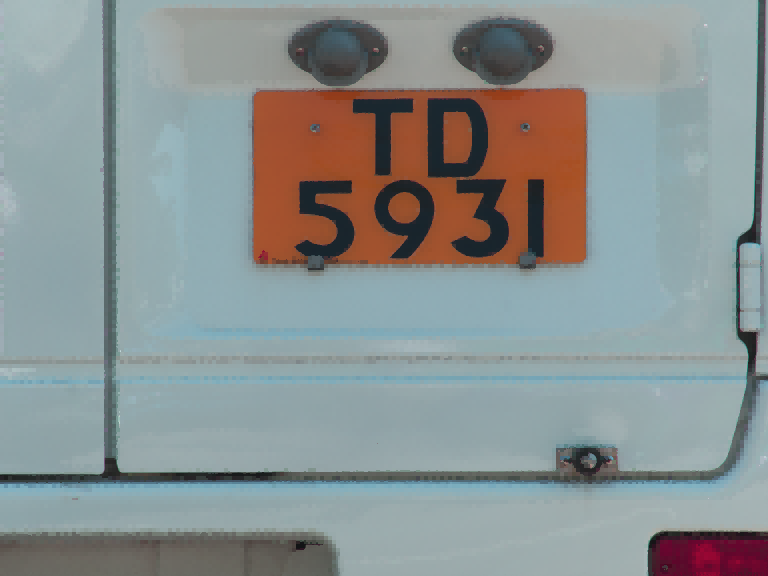} &\hspace{-4mm}
	\includegraphics[width=0.195\linewidth]{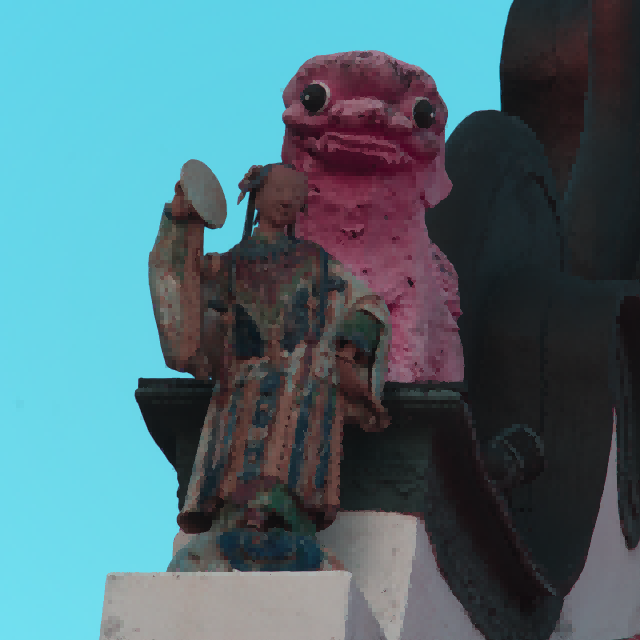}
	\\
	\hspace{-2.6mm}
	\includegraphics[width=0.195\linewidth]{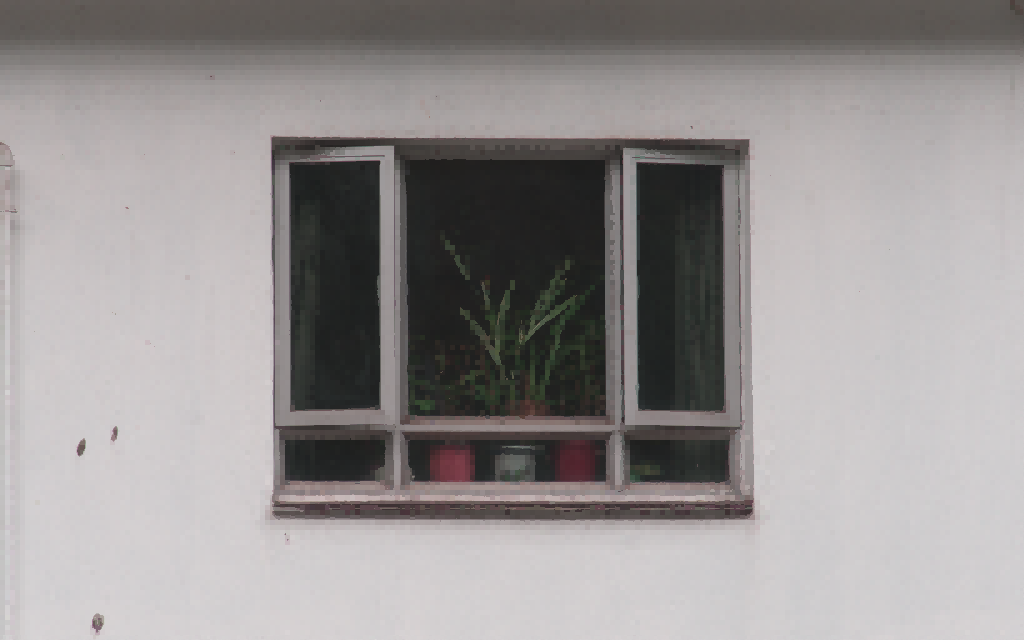} &\hspace{-4mm}
	\includegraphics[width=0.195\linewidth]{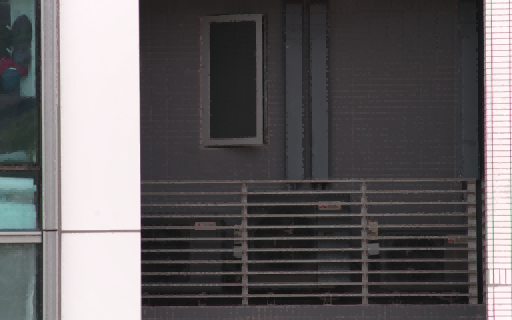} &\hspace{-4mm}
	\includegraphics[width=0.195\linewidth]{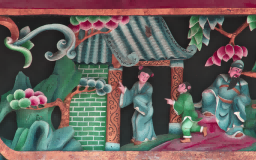} &\hspace{-4mm}
	\includegraphics[width=0.195\linewidth]{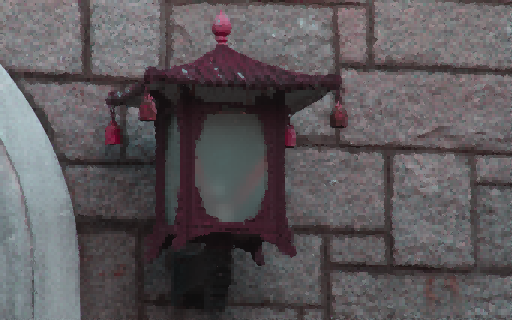} &\hspace{-4mm}
	\includegraphics[width=0.195\linewidth]{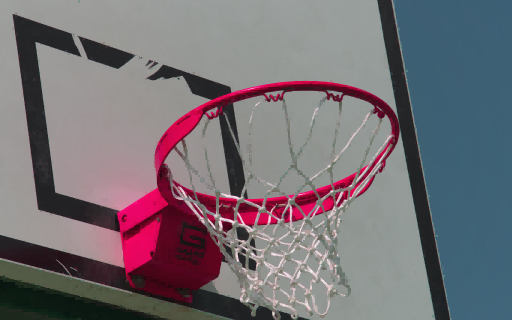}
	\\
	\hspace{-2.6mm}
	\includegraphics[width=0.195\linewidth]{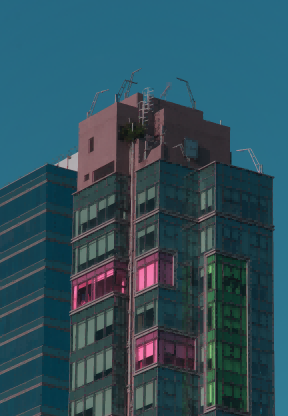} &\hspace{-4mm}
	\includegraphics[width=0.195\linewidth]{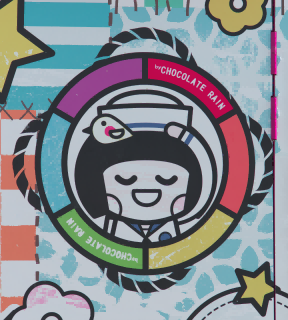} &\hspace{-4mm}
	\includegraphics[width=0.195\linewidth]{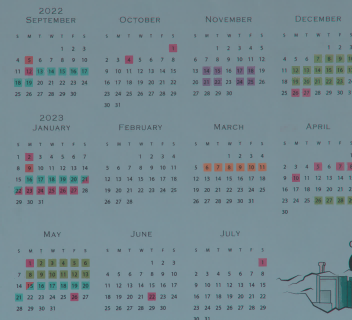} &\hspace{-4mm}
	\includegraphics[width=0.195\linewidth]{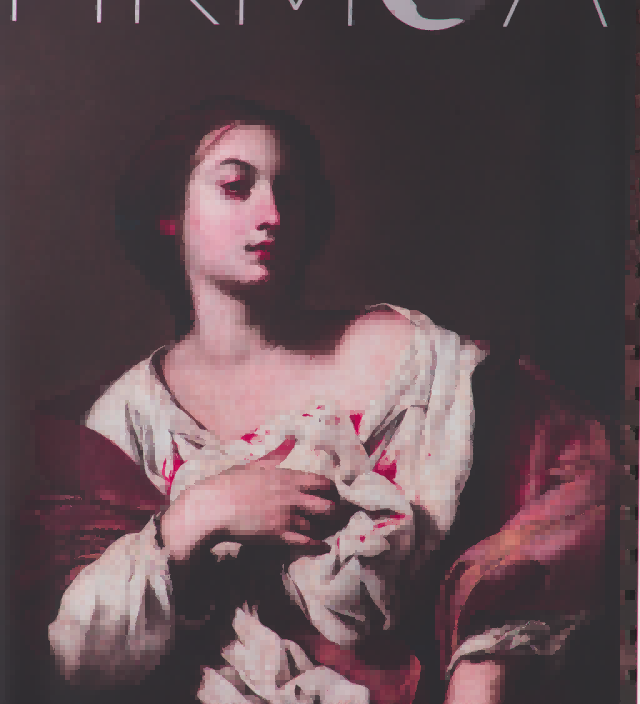} &\hspace{-4mm}
	\includegraphics[width=0.195\linewidth]{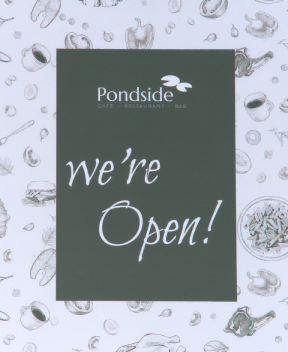}
	\\
	\end{tabular}
		%\end{center}
	%\vspace{0.5mm}
	\caption{Example normal-light scenes in the test set of our proposed SRRIIE dataset. The test set of our proposed SRRIIE dataset consists of 100 different image sequences captured from 100 non-overlapping outdoor and indoor scenes.
	}
	\vspace{-2mm}
	\label{fig: test set}
\end{figure*}

\begin{figure*}[!tp]\footnotesize
	\centering
	\vspace{0mm}
	\begin{tabular}{cccccc}
	\hspace{-2.6mm}
	\includegraphics[width=0.245\linewidth]{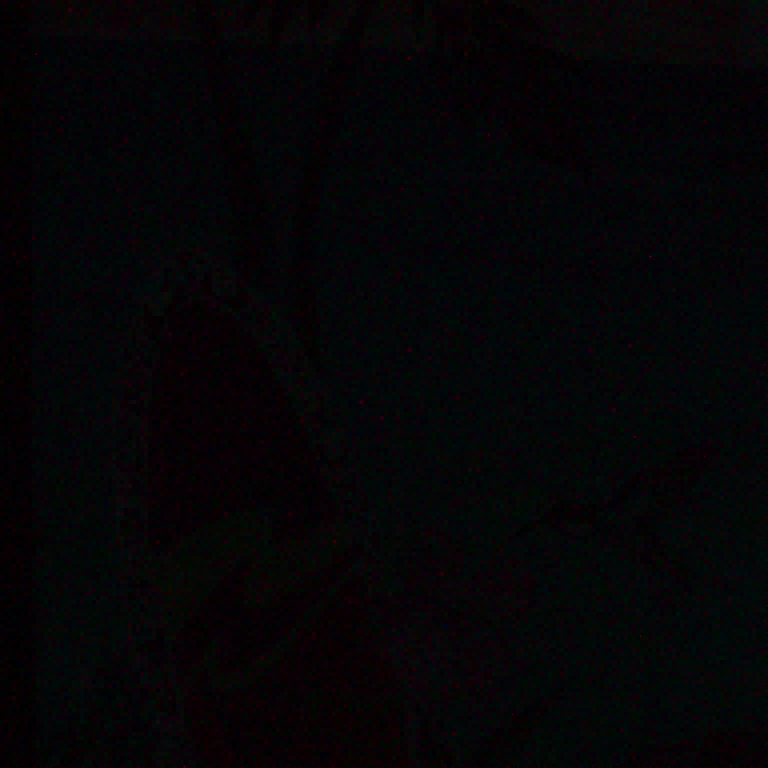} &\hspace{-4mm}
	\includegraphics[width=0.245\linewidth]{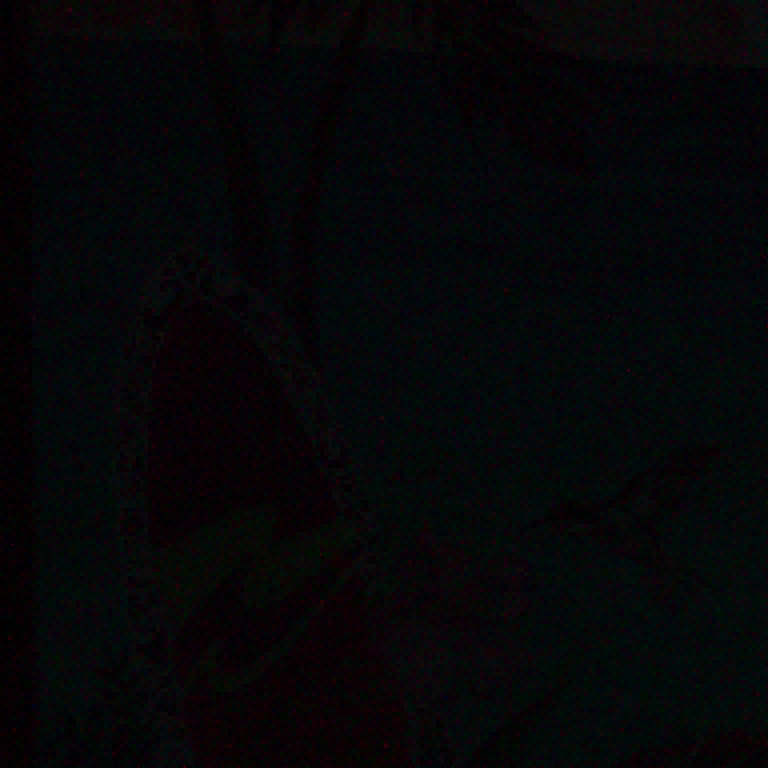} &\hspace{-4mm}
	\includegraphics[width=0.245\linewidth]{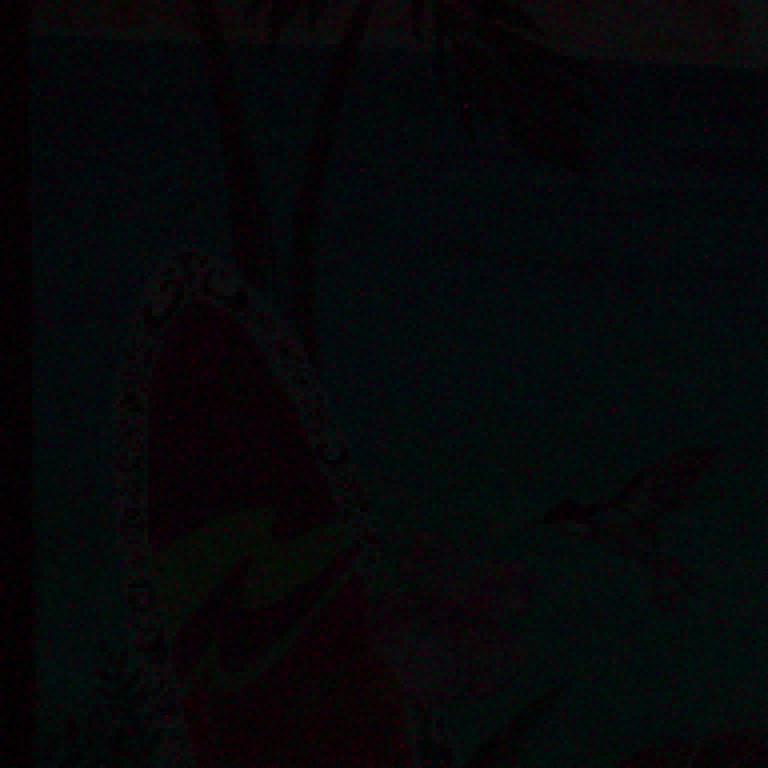} &\hspace{-4mm}
	\includegraphics[width=0.245\linewidth]{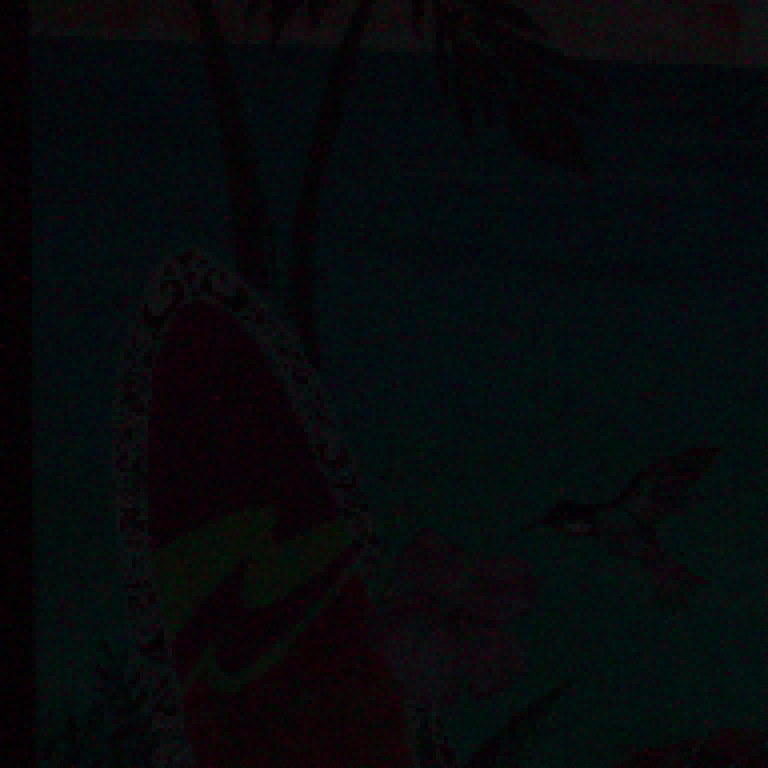}
	\\
	\hspace{-2.6mm}
	(a) -6.0 EV ($\times$1) &\hspace{-4mm} (b) -5.5 EV ($\times$1) &\hspace{-4mm} (c) -5.0 EV ($\times$1) &\hspace{-4mm} (d) -4.5 EV ($\times$1)
	\\
	\hspace{-2.6mm}
	\includegraphics[width=0.245\linewidth]{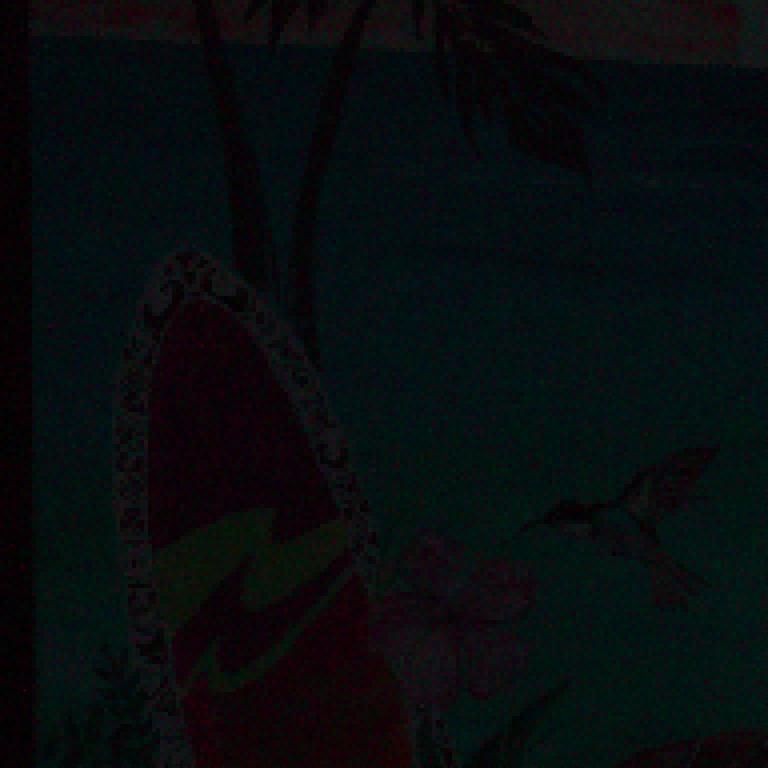} &\hspace{-4mm}
	\includegraphics[width=0.245\linewidth]{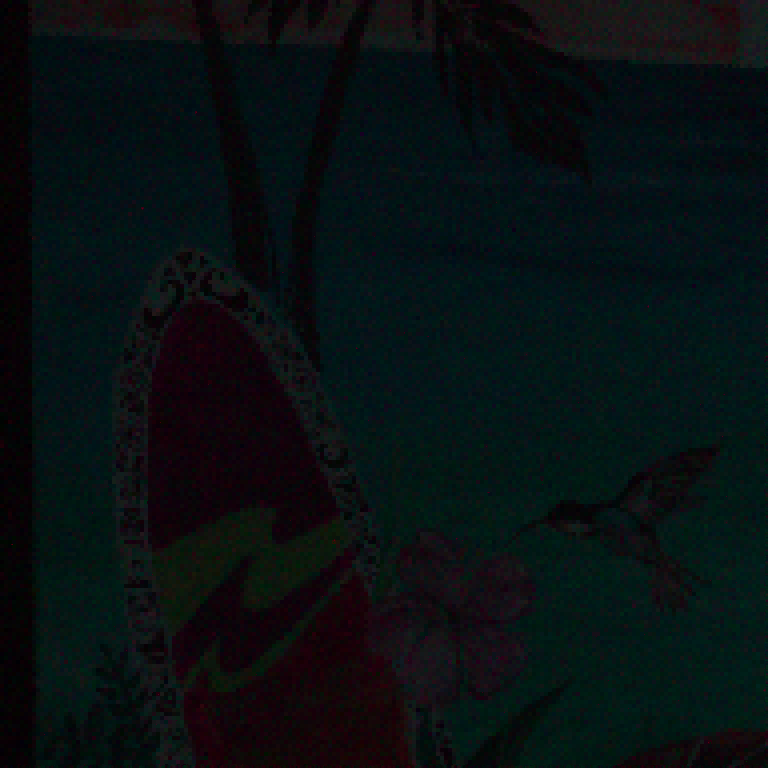} &\hspace{-4mm}
	\includegraphics[width=0.245\linewidth]{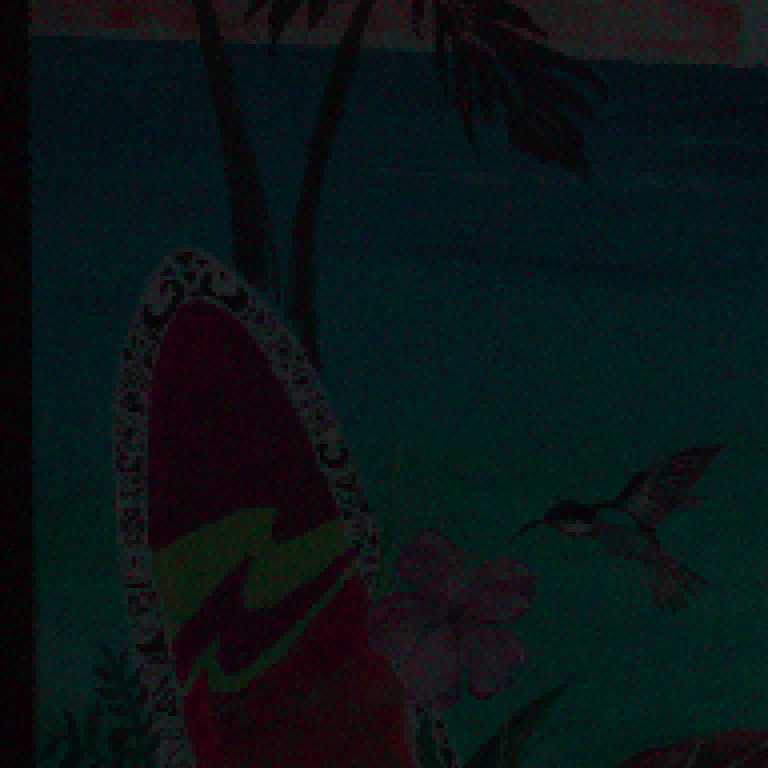} &\hspace{-4mm}
	\includegraphics[width=0.245\linewidth]{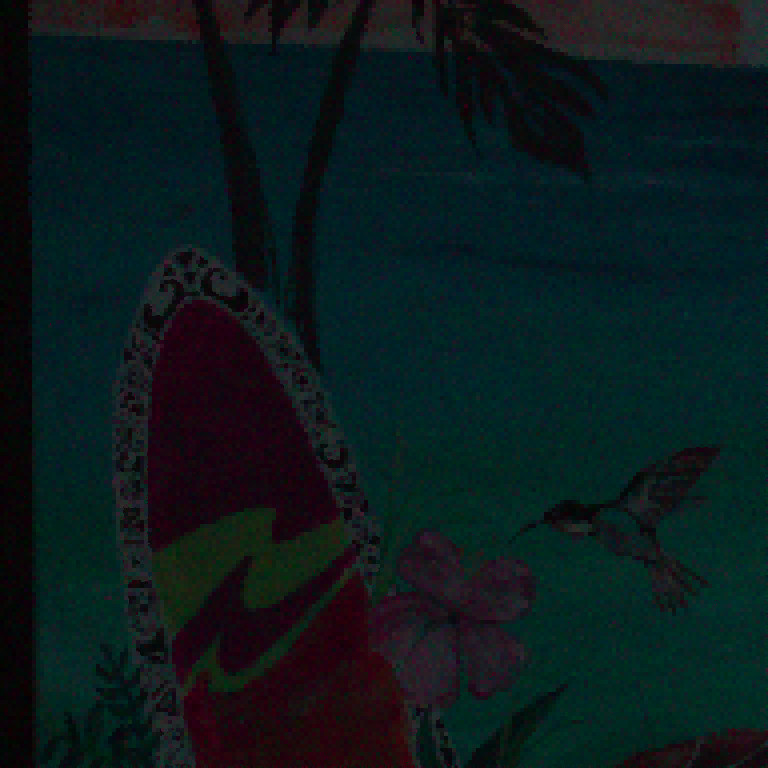}
	\\
	\hspace{-2.6mm}
	(e) -4.0 EV ($\times$1) &\hspace{-4mm} (f) -3.5 EV ($\times$1) &\hspace{-4mm} (g) -3.0 EV ($\times$1) &\hspace{-4mm} (h) -2.5 EV ($\times$1)
	\\
	\hspace{-2.6mm}
	\includegraphics[width=0.245\linewidth]{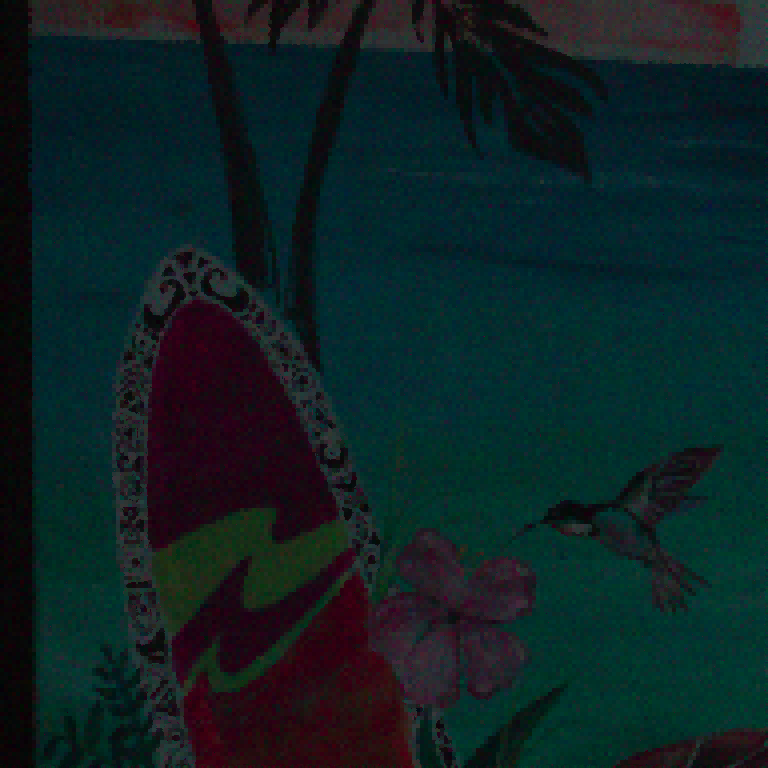} &\hspace{-4mm}
	\includegraphics[width=0.245\linewidth]{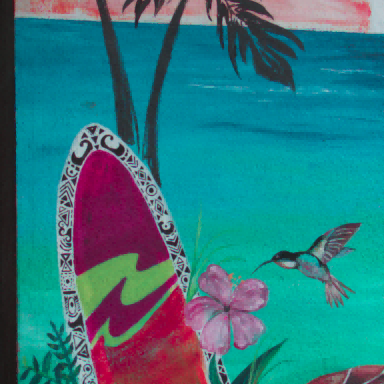} &\hspace{-4mm}
	\includegraphics[width=0.245\linewidth]{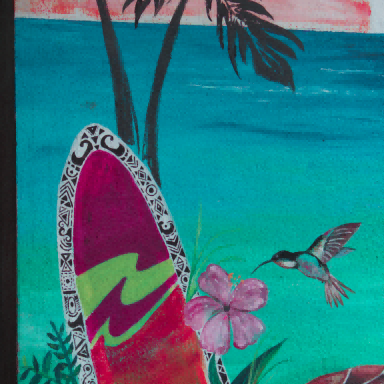} &\hspace{-4mm}
	\includegraphics[width=0.245\linewidth]{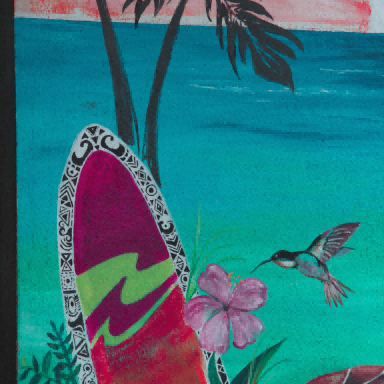}
	\\
	\hspace{-2.6mm}
	(i) -2.0 EV ($\times$1) &\hspace{-4mm} (j) 0 EV ($\times$1) &\hspace{-4mm} (k) 0 EV ($\times$2) &\hspace{-4mm} (l) 0 EV ($\times$4)
	\\
	\end{tabular}
		%\end{center}
	%\vspace{0.5mm}
	\caption{Low-light low-resolution images for each scene are captured with varying under-exposure levels ranging from -6 EV to -2 EV, while normal-light high-resolution images are captured with a fixed 0 EV. The captured low-light images are labeled as (a)-(i), and the normal-light images are labeled as (j)-(l).
		}
	\vspace{-2mm}
	\label{fig: an image sequence}
\end{figure*}

In Section 3.1 of the manuscript, we present a comprehensive data collection and processing pipeline for the SRRIIE dataset, which includes 400 paired image sequences of outdoor and indoor scenes captured with a Sony A7 IV camera and Tamron 28-200mm zoom lens. To ensure that the dataset is diverse and representative of real-world scenarios, we selectively split the image sequences, based on categories of image contents and camera ISO levels, into non-overlapping training and test sets of 300 and 100, respectively. The dataset contains various scenes, such as buildings, sculptures, wall paintings, traffic signs, license plates, and furniture. Examples of 30 normal-light scenes from the training and test sets are shown in Figure \ref{fig: train set} and \ref{fig: test set}, respectively. Each image sequence includes nine low-light low-resolution images (Raw/sRGB) captured with exposure levels ranging from -6 EV to -2 EV and ISO levels ranging from 800 to 12800, and three normal-light high-resolution images (sRGB) captured with 0 EV and ISO 50. A sample image sequence is shown in Figure \ref{fig: an image sequence}. Both Raw sensor data and sRGB data are provided in the SRRIIE dataset, and the images shown in the manuscript and supplemental material are rendered to sRGB for ease of visualization.
%
%We use different ISO values ranging from 800 to 12800 for different scenes to cover a wide range of illumination variations as shown in Figure \ref{fig: different ISO}. As the ISO value increases, the real noises become more severe. Existing algorithms performs less effectively in these cases as shown in Sections 5 and 6 of the manuscript, and Section 2 in this document. 
%
%
Note that the original camera resolution is 4672 $\times$ 7008 and we preprocess each image sequence with different center crop sizes.

\clearpage

\clearpage

\begin{table*}[!htp]\small
\centering
\caption{We compared the mean ($\mu$) and standard deviation ($\sigma$) of our SRRIIE dataset to those of the RELLISUR \cite{aakerberg2021rellisur} dataset. Our dataset exhibits a wider range of under-exposure levels and lower signal deviations, making it more challenging than the RELLISUR dataset.
}
\vspace{1mm}
\label{tab:mu-sigma}
\begin{tabular}{|ll|c|c|c|c|c|c|c|c|c|}
\hline
\multicolumn{2}{|c|}{Dataset}                           & -2.0 EV & -2.5 EV & -3.0 EV & -3.5 EV & -4.0 EV & -4.5 EV & -5.0 EV & -5.5 EV & -6.0 EV \\ 
\hline
\hline
\multicolumn{1}{|l|}{\multirow{2}{*}{RELLISUR}} & $\mu$    & -       & 22.59   & 17.15   & 12.20   & 8.60    & 6.03    & 4.29    & -       & -       \\
\multicolumn{1}{|l|}{}                          & $\sigma$ & -       & 16.83   & 12.76   & 9.36    & 6.79    & 4.91    & 3.55    & -       & -       \\ \hline
\multicolumn{1}{|l|}{\multirow{2}{*}{SRRIIE}}     & $\mu$    & 26.15   & 20.69   & 16.38   & 13.05   & 10.44   & 8.45    & 6.88    & 5.77    & 4.98    \\
\multicolumn{1}{|l|}{}                          & $\sigma$ & 10.90   & 8.68    & 6.92    & 5.56    & 4.50    & 3.70    & 3.07    & 2.67    & 2.41    \\ \hline
\end{tabular}
\end{table*}

The RELLISUR dataset is the most related dataset to our proposed SRRIIE dataset \cite{aakerberg2021rellisur}. However, it is captured in well-lighted conditions with low ISO levels and does not consider the noise model. Consequently, the low-light images in RELLISUR contain less real-world noise. In contrast, our proposed SRRIIE dataset is captured with varying exposure levels ranging from -6 EV to 0 EV and ISO levels ranging from 50 to 12800, to model real-world image degradation in low-illumination environments. As a result, the image restoration problems in our dataset are highly ill-posed for existing restoration and enhancement methods. We have illustrated the diverse image gradient distributions of different noisy images in Figure 3 of the manuscript and conducted an ablation study experiment in Section 7.3 of the manuscript to demonstrate cross-dataset generalization using our proposed method. We further summarized the average mean $\mu$ and standard deviation value $\sigma$ of our SRRIIE dataset and RELLISUR dataset in Table \ref{tab:mu-sigma} in the supplementary material. Our proposed SRRIIE dataset contains a wider range of under-exposure levels and lower signal deviations than the RELLISUR dataset. The low contrast and small spread characteristics of our SRRIIE dataset make it challenging to learn with existing deep learning-based methods, and these significant differences distinguish our real-world SRRIIE dataset from existing datasets such as the RELLISUR dataset.

\section{More Experimental Results}
\label{sec: More Experimental Results}

\begin{figure*}[!tp]\small
	\centering
	\vspace{0mm}
	\begin{tabular}{cccccc}
	\hspace{-2.6mm}
	\includegraphics[width=0.245\linewidth]{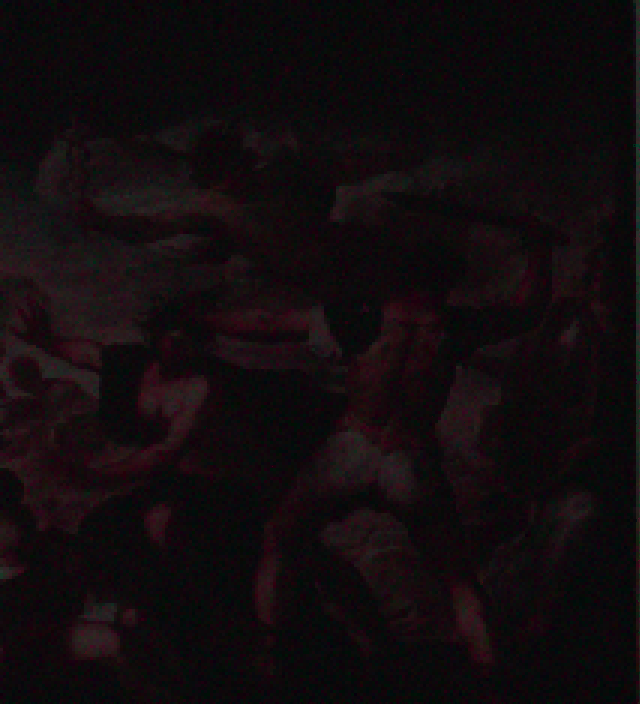} &\hspace{-4mm}
	\includegraphics[width=0.245\linewidth]{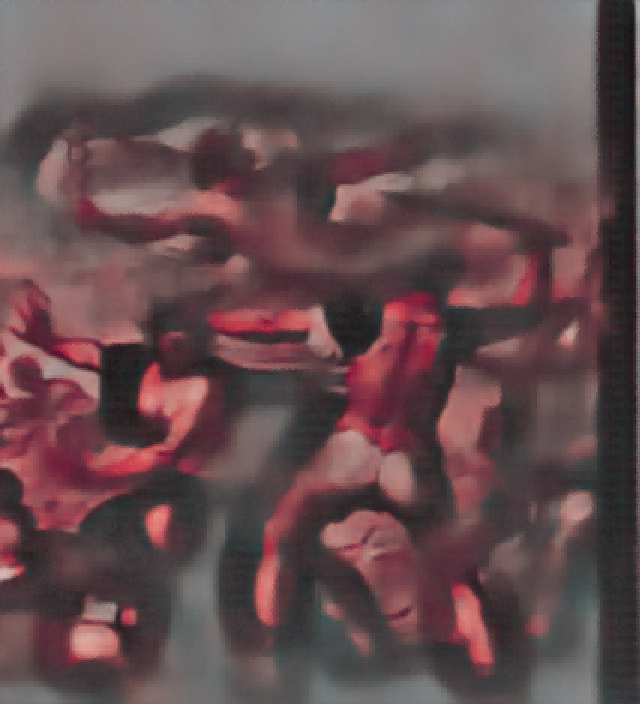} &\hspace{-4mm}
	\includegraphics[width=0.245\linewidth]{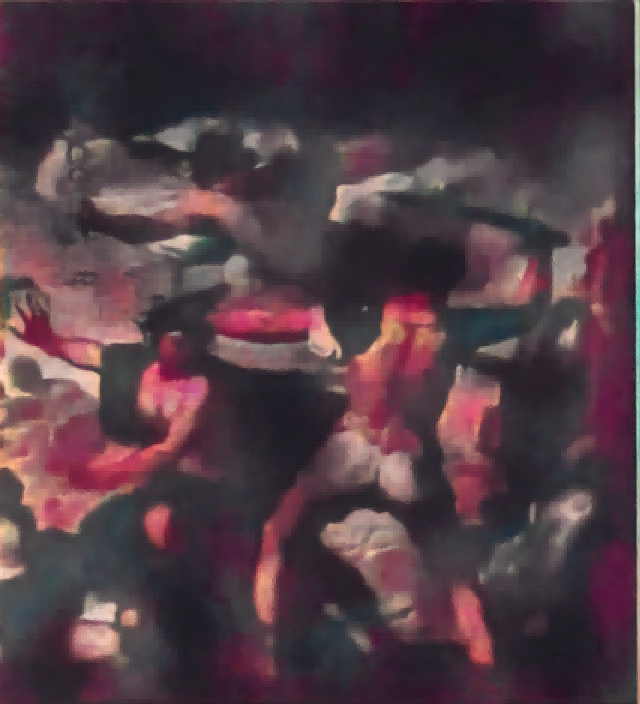} &\hspace{-4mm}
	\includegraphics[width=0.245\linewidth]{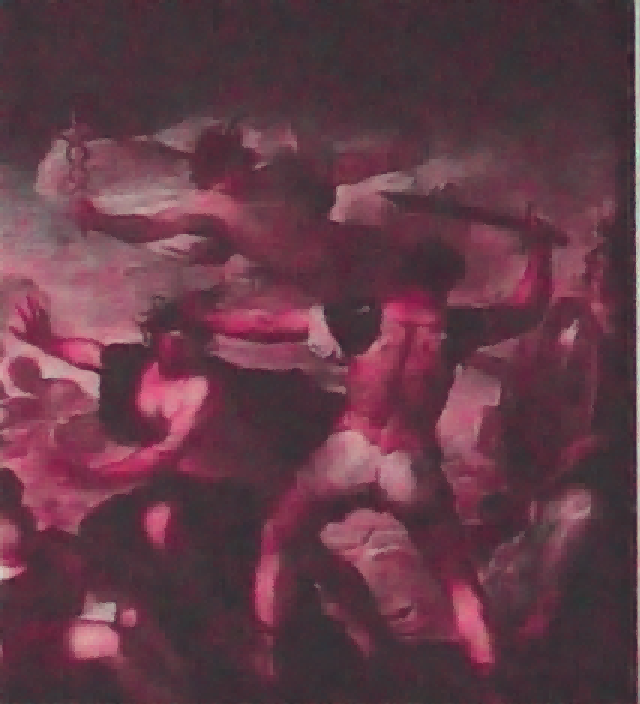}
	\\
	\hspace{-2.6mm}
	(a) Input image &\hspace{-4.5mm} (b) USRNet \cite{zhang2020deep} &\hspace{-4.5mm} (c) MIRNet \cite{Zamir2022MIRNetv2} $\rightarrow$ SwinIR \cite{liang2021swinir} &\hspace{-4.5mm} (d) MIRNet \cite{Zamir2022MIRNetv2}
	\\
	\hspace{-2.6mm}
	\includegraphics[width=0.245\linewidth]{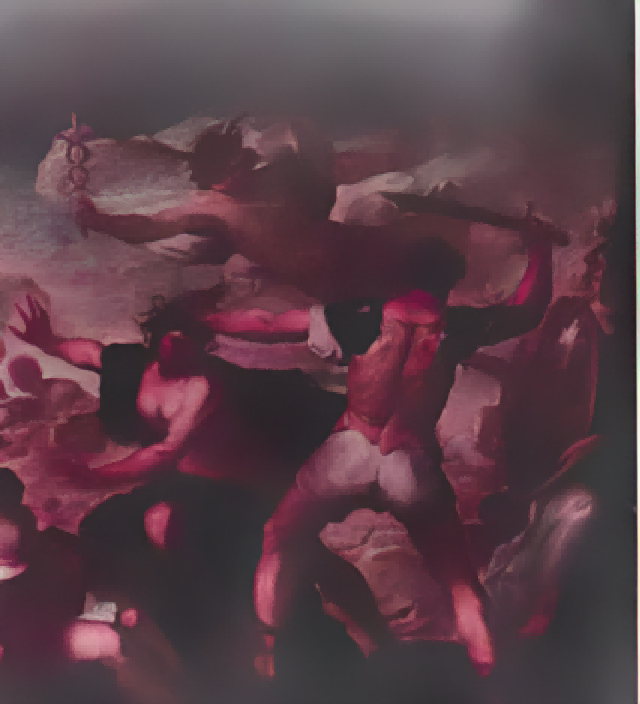} &\hspace{-4mm}
	\includegraphics[width=0.245\linewidth]{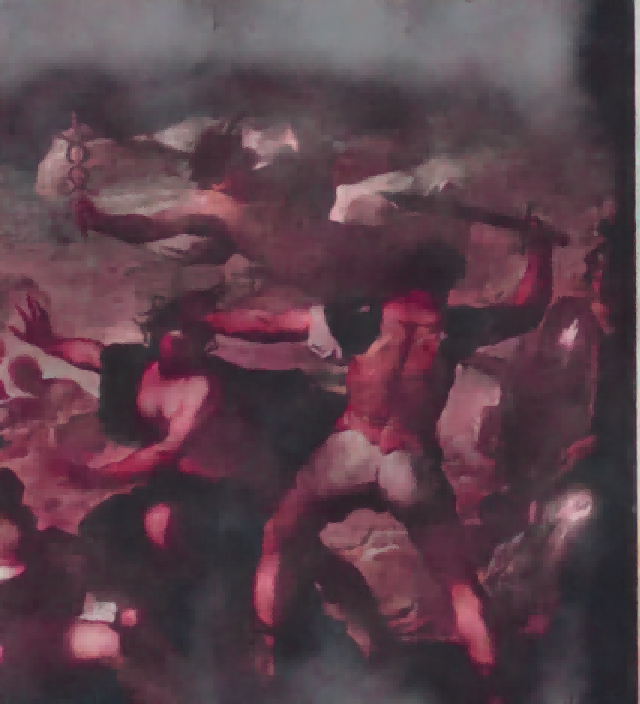} &\hspace{-4mm}
	\includegraphics[width=0.245\linewidth]{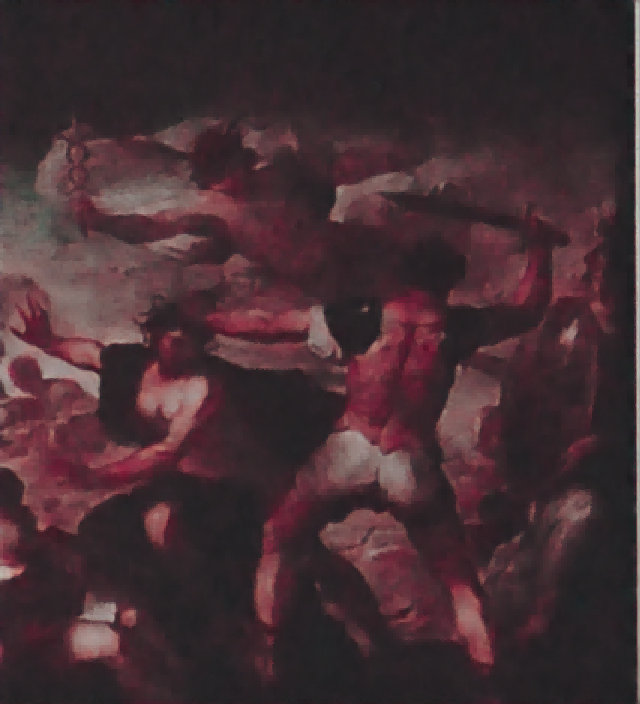} &\hspace{-4mm}
	\includegraphics[width=0.245\linewidth]{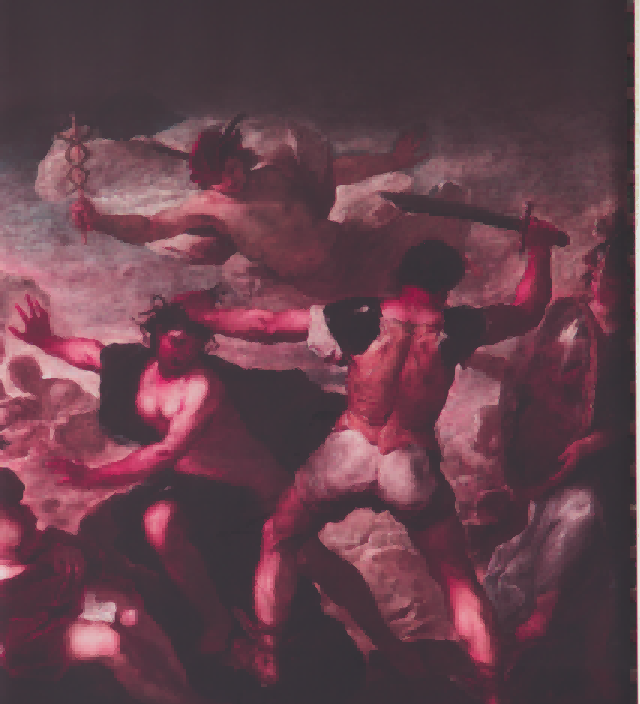}
	\\
	\hspace{-2.6mm}
	(e) Real-ESRGAN \cite{wang2021realesrgan} &\hspace{-4.5mm} (f) SwinIR \cite{liang2021swinir} &\hspace{-4.5mm} (g) Ours &\hspace{-4.5mm} (h) Ground Truth
	\\
	\end{tabular}
		%\end{center}
	%\vspace{0.5mm}
	\caption{Qualitative visual evaluations ($\times$2 SR) for the compared methods on the proposed SRRIIE dataset. The low-quality image is captured with the camera settings of -2.5 EV and ISO 4000. While other methods generate severe artifacts, our proposed method restores more natural and visually pleasing results.
	}
	\vspace{-2mm}
	\label{fig: sota-1}
\end{figure*}

\begin{figure*}[!tp]\small
	\centering
	\vspace{0mm}
	\begin{tabular}{cccccc}
	\hspace{-2.6mm}
	\includegraphics[width=0.245\linewidth]{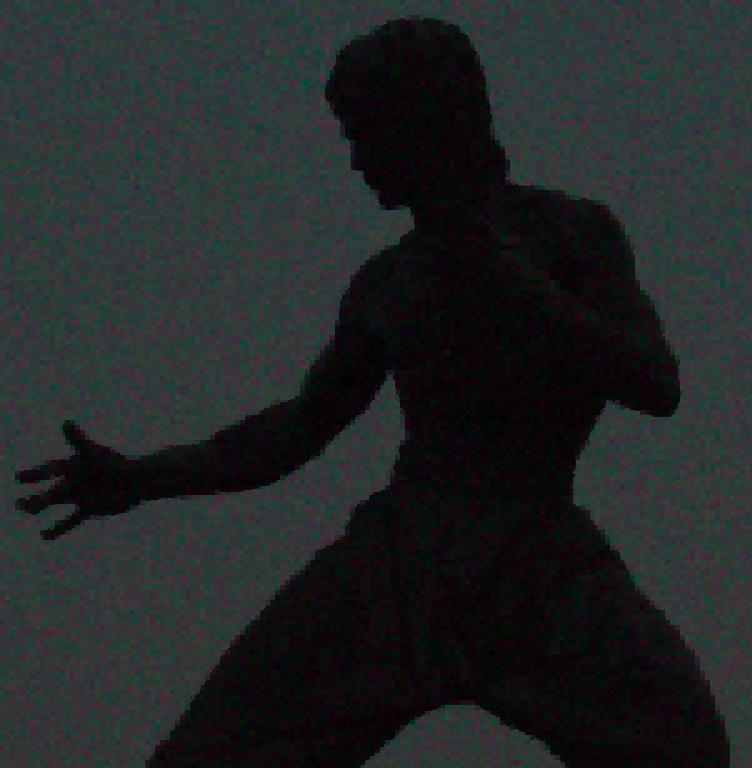} &\hspace{-4mm}
	\includegraphics[width=0.245\linewidth]{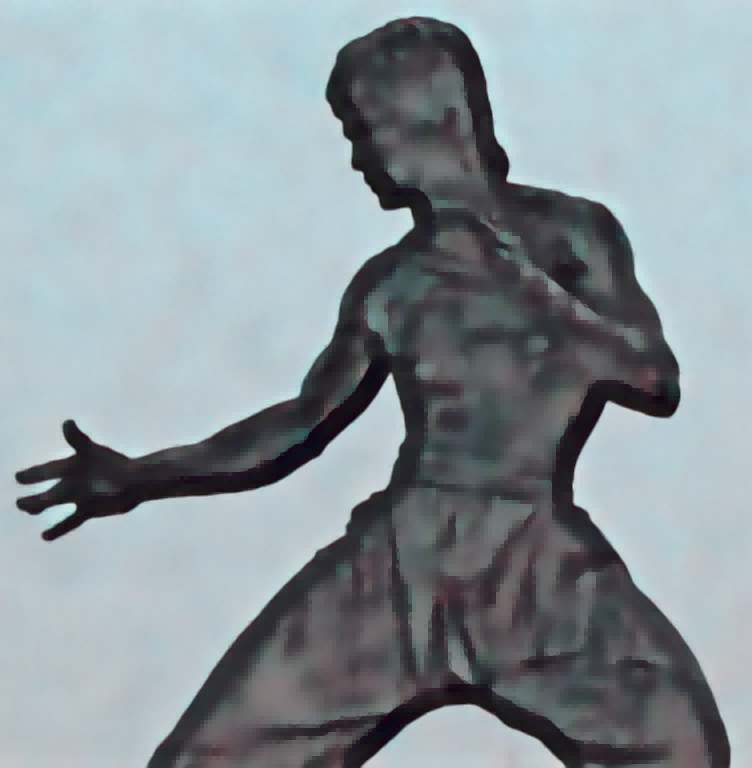} &\hspace{-4mm}
	\includegraphics[width=0.245\linewidth]{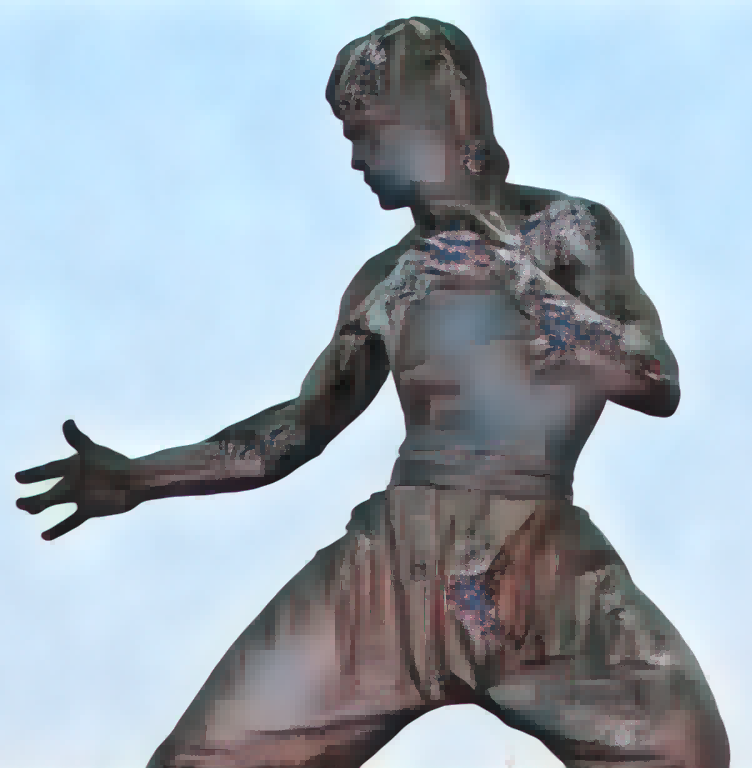} &\hspace{-4mm}
	\includegraphics[width=0.245\linewidth]{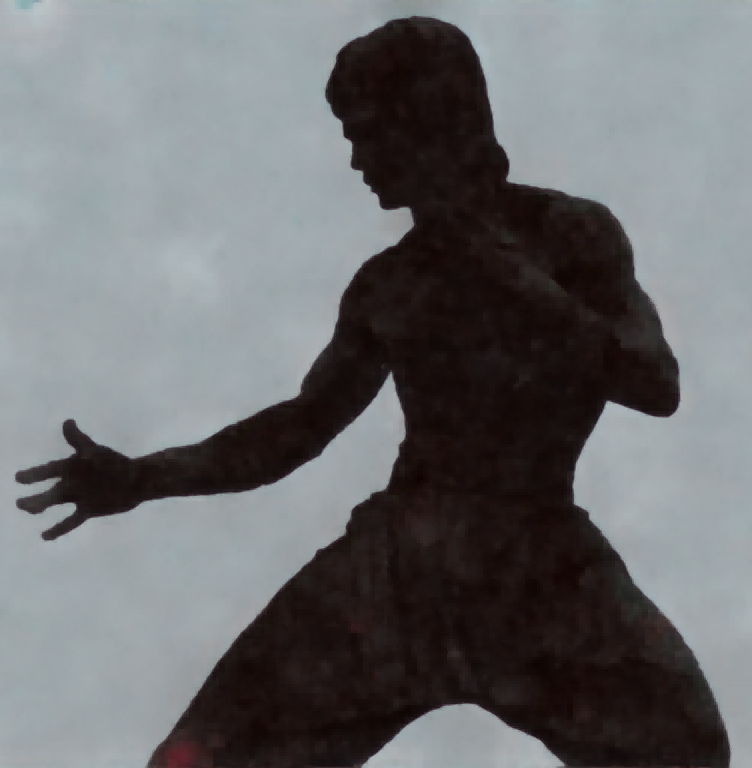}
	\\
	\hspace{-2.6mm}
	(a) Input image &\hspace{-4.5mm} (b) USRNet \cite{zhang2020deep} &\hspace{-4.5mm} (c) RealSR \cite{Ji_2020_CVPR_Workshops} &\hspace{-4.5mm} (d) MIRNet \cite{Zamir2022MIRNetv2}
	\\
	\hspace{-2.6mm}
	\includegraphics[width=0.245\linewidth]{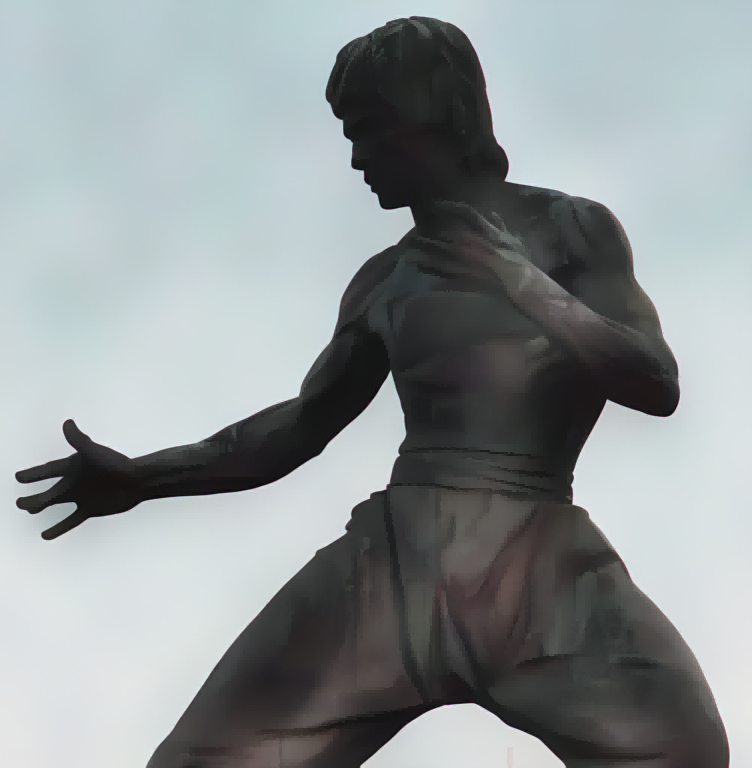} &\hspace{-4mm}
	\includegraphics[width=0.245\linewidth]{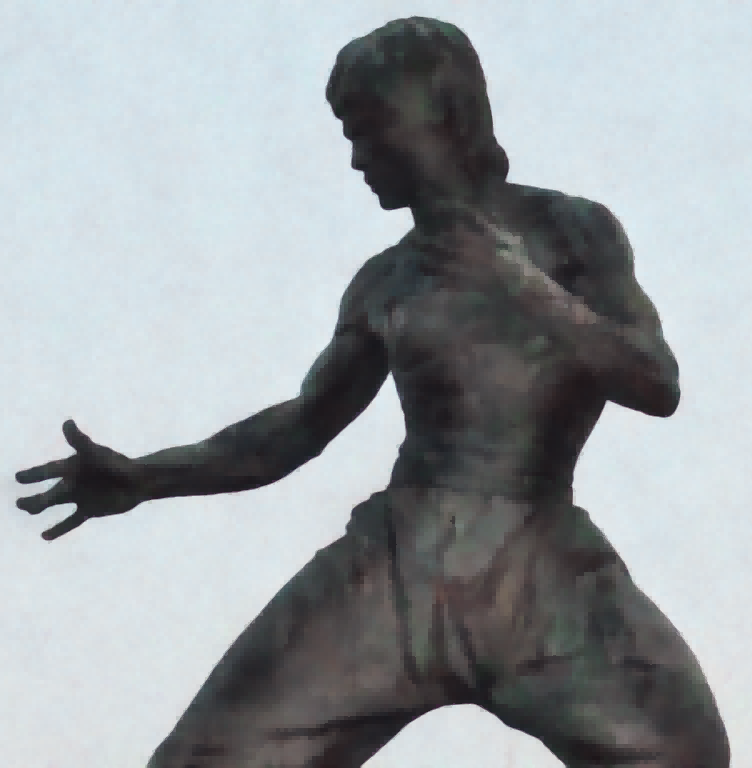} &\hspace{-4mm}
	\includegraphics[width=0.245\linewidth]{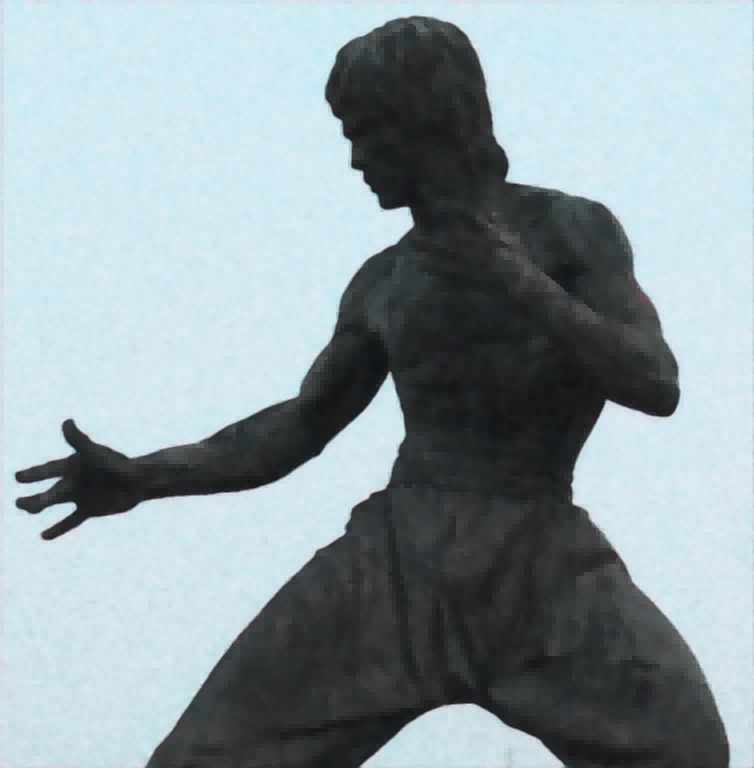} &\hspace{-4mm}
	\includegraphics[width=0.245\linewidth]{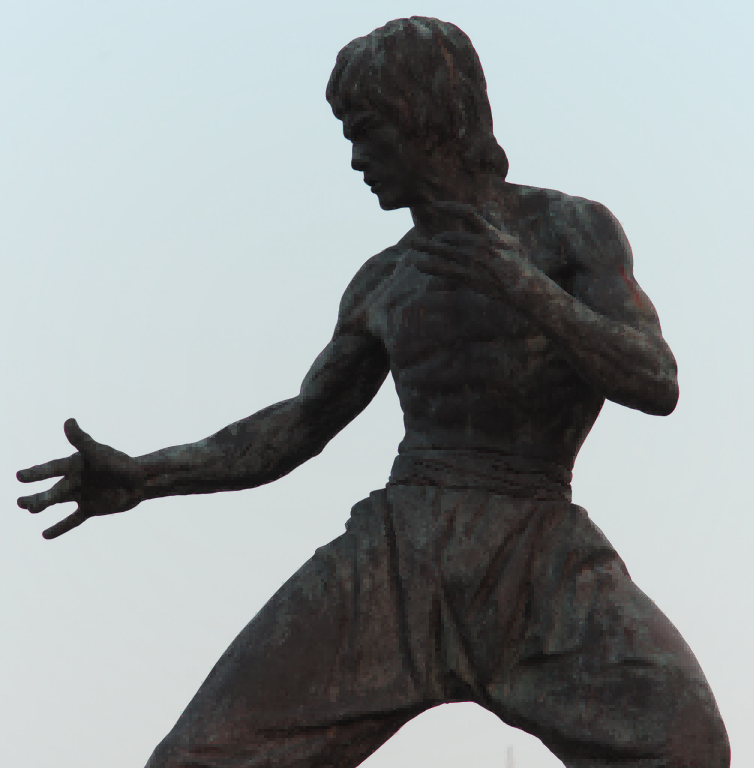}
	\\
	\hspace{-2.6mm}
	(e) Real-ESRGAN \cite{wang2021realesrgan} &\hspace{-4.5mm} (f) SwinIR \cite{liang2021swinir} &\hspace{-4.5mm} (g) Ours &\hspace{-4.5mm} (h) Ground Truth
	\\
	\end{tabular}
		%\end{center}
	%\vspace{0.5mm}
	\caption{Qualitative visual evaluations ($\times$4 SR) for the compared methods on the proposed SRRIIE dataset. The low-quality image is captured with the camera settings of -4 EV and ISO 5000. While other methods generated severe artifacts, our proposed method performed favorably as demonstrated by the resulting images.
	}
	\vspace{-2mm}
	\label{fig: sota-2}
\end{figure*}

In Section 6.2 of the manuscript, we compare our proposed method against state-of-the-art algorithms on our proposed SRRIIE dataset. We employ multiple reconstruction and perceptual metrics including PSNR, SSIM, LPIPS, DISTS, and FID to benchmark the compared methods. Our comprehensive quantitative and qualitative experimental results demonstrate that our proposed conditional diffusion model-based method outperforms other methods. We present more resulting images on our SRRIIE dataset in Figures \ref{fig: sota-1} and \ref{fig: sota-2}. We use the publicly available codes of state-of-the-art methods and fine-tune their pre-trained models on our dataset instead of training from scratch to ensure fair comparisons. Our proposed method effectively generates natural results. In contrast, state-of-the-art methods often fail to restore images and generate severe artifacts in the high ill-posed problem of super-resolving real-world image illumination enhancement.

\end{appendix}

\end{document}

\title{Super-resolving Real-world Illumination Enhancement: \\
A Raw Image Dataset and Baseline}

\author{First Author\\
Institution1\\
Institution1 address\\
{\tt\small firstauthor@i1.org}
\and
Second Author\\
Institution2\\
First line of institution2 address\\
{\tt\small secondauthor@i2.org}
}
\maketitle

%%%%%%%%% ABSTRACT
\section*{\centering Overview}
\label{sec: Overview}
We thank the reviewers for viewing the supplemental material. 
In this supplementary document, we further show more real-world scenes during the image collection of our IESR-RAW dataset in Section \ref{sec: iesr-raw dataset}. And we show more and larger qualitative visual results on a variety of scenes and compare our method against state-of-the-art algorithms in Section \ref{sec: More Experimental Results}.
The results shown in this supplementary material are best viewed on a high-resolution display.

\section{Real-world Scenes in the proposed IESR-RAW Dataset}
\label{sec: iesr-raw dataset}

\begin{figure*}[!tp]
	\centering
	\vspace{0mm}
	\begin{tabular}{cccccc}
	\hspace{-2.6mm}
	\includegraphics[width=0.195\linewidth]{figures/Dataset/train/297_x4_10.png} &\hspace{-4mm}
	\includegraphics[width=0.195\linewidth]{figures/Dataset/train/256_x4_10.png} &\hspace{-4mm}
	\includegraphics[width=0.195\linewidth]{figures/Dataset/train/336_x4_10.png} &\hspace{-4mm}
	\includegraphics[width=0.195\linewidth]{figures/Dataset/train/87_x4_10.png} &\hspace{-4mm}
	\includegraphics[width=0.195\linewidth]{figures/Dataset/train/98_x4_10.png}
	\\
	\includegraphics[width=0.195\linewidth]{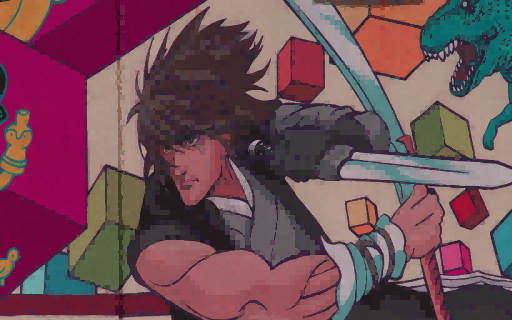} &\hspace{-4mm}
	\includegraphics[width=0.195\linewidth]{figures/Dataset/train/122_x4_10.png} &\hspace{-4mm}
	\includegraphics[width=0.195\linewidth]{figures/Dataset/train/181_x4_10.png} &\hspace{-4mm}
	\includegraphics[width=0.195\linewidth]{figures/Dataset/train/221_x4_10.png} &\hspace{-4mm} 
	\includegraphics[width=0.195\linewidth]{figures/Dataset/train/287_x4_10.png}
	\\
	\includegraphics[width=0.195\linewidth]{figures/Dataset/train/75_x4_10.png} &\hspace{-4mm}
	\includegraphics[width=0.195\linewidth]{figures/Dataset/train/265_x4_10.png} &\hspace{-4mm}
	\includegraphics[width=0.195\linewidth]{figures/Dataset/train/183_x4_10.png} &\hspace{-4mm}
	\includegraphics[width=0.195\linewidth]{figures/Dataset/train/217_x4_10.png} &\hspace{-4mm}
	\includegraphics[width=0.195\linewidth]{figures/Dataset/train/246_x4_10.png}
	\\
	\includegraphics[width=0.195\linewidth]{figures/Dataset/train/59_x4_10.png} &\hspace{-4mm}
	\includegraphics[width=0.195\linewidth]{figures/Dataset/train/301_x4_10.png} &\hspace{-4mm}
	\includegraphics[width=0.195\linewidth]{figures/Dataset/train/323_x4_10.png} &\hspace{-4mm}
	\includegraphics[width=0.195\linewidth]{figures/Dataset/train/84_x4_10.png} &\hspace{-4mm}
	\includegraphics[width=0.195\linewidth]{figures/Dataset/train/384_x4_10.png}
	
	\\
	\includegraphics[width=0.195\linewidth]{figures/Dataset/train/359_x4_10.png} &\hspace{-4mm}
	\includegraphics[width=0.195\linewidth]{figures/Dataset/train/362_x4_10.png} &\hspace{-4mm}
	\includegraphics[width=0.195\linewidth]{figures/Dataset/train/365_x4_10.png} &\hspace{-4mm}
	\includegraphics[width=0.195\linewidth]{figures/Dataset/train/370_x4_10.png} &\hspace{-4mm}
	\includegraphics[width=0.195\linewidth]{figures/Dataset/train/348_x4_10.png}
	\\
	\includegraphics[width=0.195\linewidth]{figures/Dataset/train/214_x4_10.png} &\hspace{-4mm}
	\includegraphics[width=0.195\linewidth]{figures/Dataset/train/274_x4_10.png} &\hspace{-4mm}
	\includegraphics[width=0.195\linewidth]{figures/Dataset/train/282_x4_10.png} &\hspace{-4mm}
	\includegraphics[width=0.195\linewidth]{figures/Dataset/train/158_x4_10.png} &\hspace{-4mm}
	\includegraphics[width=0.195\linewidth]{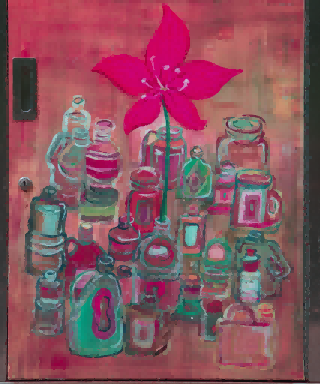}
	\\
	\end{tabular}
		%\end{center}
	%\vspace{0.5mm}
	\caption{Example normal-light scenes in the training set of our proposed IESR-RAW Dataset. There are 300 non-overlapping outdoor and indoor scenes for the data collection of 300 different image sequences in the training set.
	}
	\vspace{-4mm}
	\label{fig: train set}
\end{figure*}

\begin{figure*}[!tp]
	\centering
	\vspace{0mm}
	\begin{tabular}{cccccc}
	\hspace{-2.6mm}
	\includegraphics[width=0.195\linewidth]{figures/Dataset/test/7_x4_10.png} &\hspace{-4mm}
	\includegraphics[width=0.195\linewidth]{figures/Dataset/test/20_x4_10.png} &\hspace{-4mm}
	\includegraphics[width=0.195\linewidth]{figures/Dataset/test/275_x4_10.png} &\hspace{-4mm}
	\includegraphics[width=0.195\linewidth]{figures/Dataset/test/55_x4_10.png} &\hspace{-4mm}
	\includegraphics[width=0.195\linewidth]{figures/Dataset/test/57_x4_10.png}
	\\
	\includegraphics[width=0.195\linewidth]{figures/Dataset/test/76_x4_10.png} &\hspace{-4mm}
	\includegraphics[width=0.195\linewidth]{figures/Dataset/test/393_x4_10.png} &\hspace{-4mm}
	\includegraphics[width=0.195\linewidth]{figures/Dataset/test/97_x4_10.png} &\hspace{-4mm}
	\includegraphics[width=0.195\linewidth]{figures/Dataset/test/117_x4_10.png} &\hspace{-4mm}
	\includegraphics[width=0.195\linewidth]{figures/Dataset/test/119_x4_10.png}
	\\
	\includegraphics[width=0.195\linewidth]{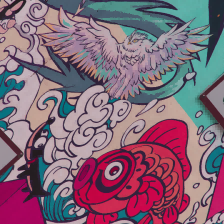} &\hspace{-4mm}
	\includegraphics[width=0.195\linewidth]{figures/Dataset/test/160_x4_10.png} &\hspace{-4mm}
	\includegraphics[width=0.195\linewidth]{figures/Dataset/test/167_x4_10.png} &\hspace{-4mm}
	\includegraphics[width=0.195\linewidth]{figures/Dataset/test/366_x4_10.png} &\hspace{-4mm}
	\includegraphics[width=0.195\linewidth]{figures/Dataset/test/286_x4_10.png}
	\\
	\includegraphics[width=0.195\linewidth]{figures/Dataset/test/188_x4_10.png} &\hspace{-4mm}
	\includegraphics[width=0.195\linewidth]{figures/Dataset/test/203_x4_10.png} &\hspace{-4mm}
	\includegraphics[width=0.195\linewidth]{figures/Dataset/test/211_x4_10.png} &\hspace{-4mm}
	\includegraphics[width=0.195\linewidth]{figures/Dataset/test/385_x4_10.png} &\hspace{-4mm}
	\includegraphics[width=0.195\linewidth]{figures/Dataset/test/357_x4_10.png}
	\\
	\includegraphics[width=0.195\linewidth]{figures/Dataset/test/179_x4_10.png} &\hspace{-4mm}
	\includegraphics[width=0.195\linewidth]{figures/Dataset/test/50_x4_10.png} &\hspace{-4mm}
	\includegraphics[width=0.195\linewidth]{figures/Dataset/test/309_x4_10.png} &\hspace{-4mm}
	\includegraphics[width=0.195\linewidth]{figures/Dataset/test/291_x4_10.png} &\hspace{-4mm}
	\includegraphics[width=0.195\linewidth]{figures/Dataset/test/298_x4_10.png}
	\\
	\includegraphics[width=0.195\linewidth]{figures/Dataset/test/273_x4_10.png} &\hspace{-4mm}
	\includegraphics[width=0.195\linewidth]{figures/Dataset/test/270_x4_10.png} &\hspace{-4mm}
	\includegraphics[width=0.195\linewidth]{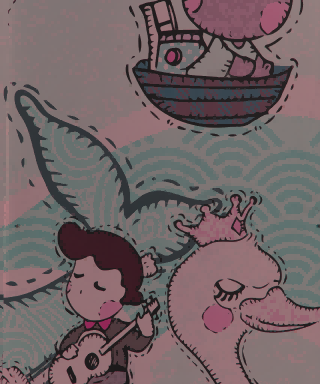} &\hspace{-4mm}
	\includegraphics[width=0.195\linewidth]{figures/Dataset/test/86_x4_10.png} &\hspace{-4mm}
	\includegraphics[width=0.195\linewidth]{figures/Dataset/test/175_x4_10.png}
	\\
	\end{tabular}
		%\end{center}
	%\vspace{0.5mm}
	\caption{Example normal-light scenes in the test set of our proposed IESR-RAW Dataset. There are 100 non-overlapping outdoor and indoor scenes for the data collection of 100 different image sequences in the test set.
	}
	\vspace{-4mm}
	\label{fig: test set}
\end{figure*}

\begin{figure*}[!tp]\footnotesize
	\centering
	\vspace{0mm}
	\begin{tabular}{cccccc}
	\hspace{-2.6mm}
	\includegraphics[width=0.245\linewidth]{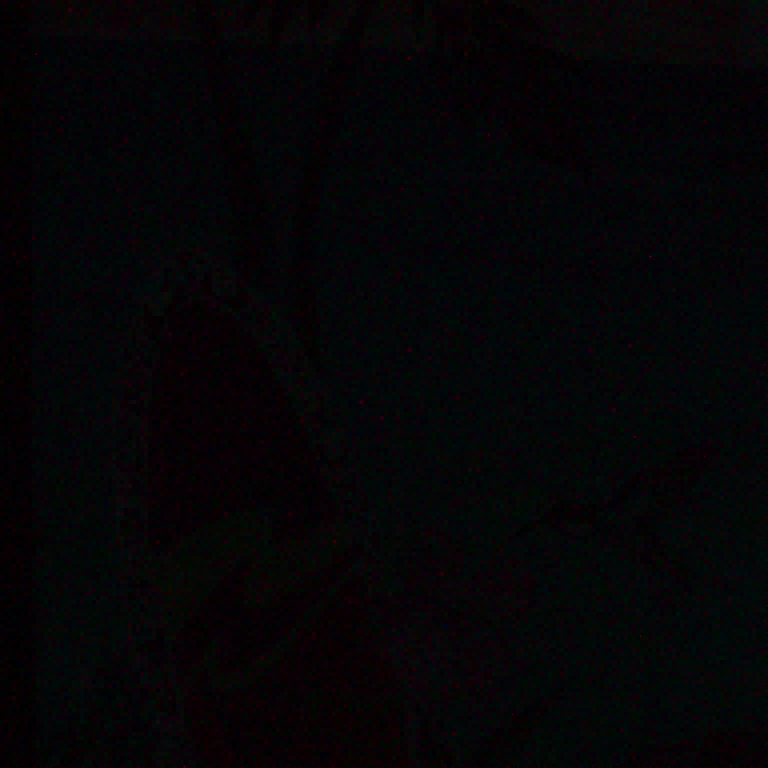} &\hspace{-4mm}
	\includegraphics[width=0.245\linewidth]{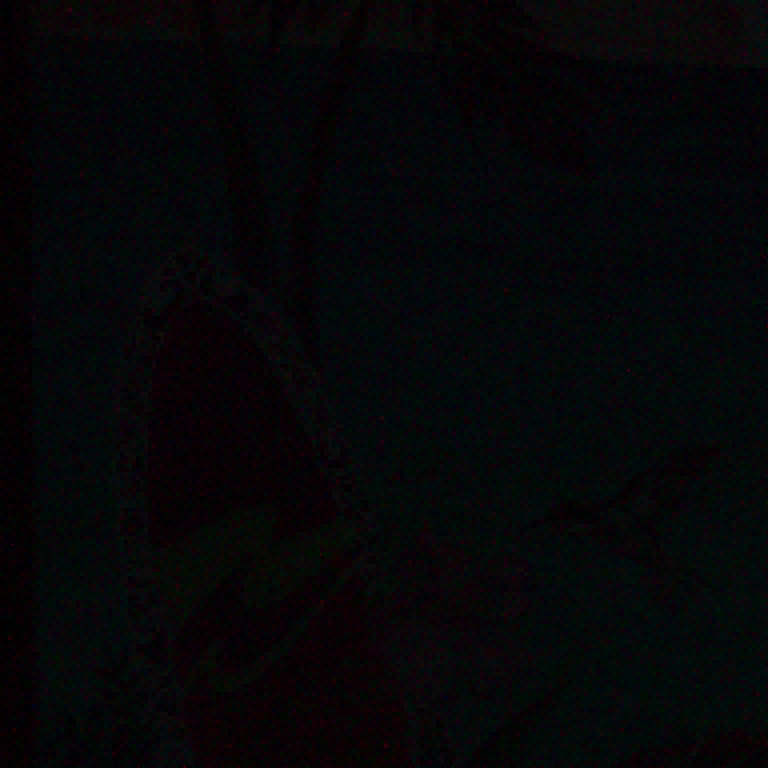} &\hspace{-4mm}
	\includegraphics[width=0.245\linewidth]{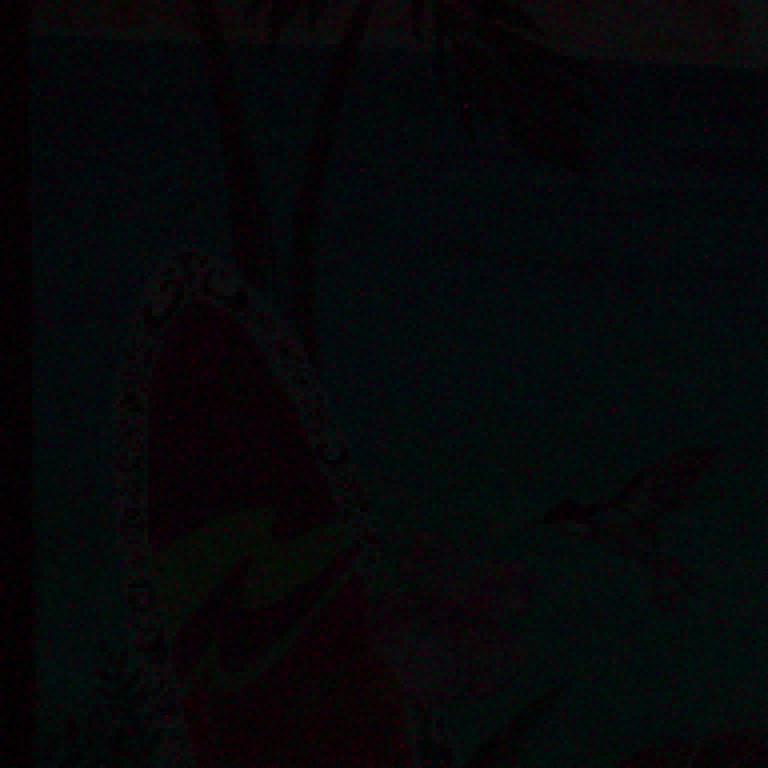} &\hspace{-4mm}
	\includegraphics[width=0.245\linewidth]{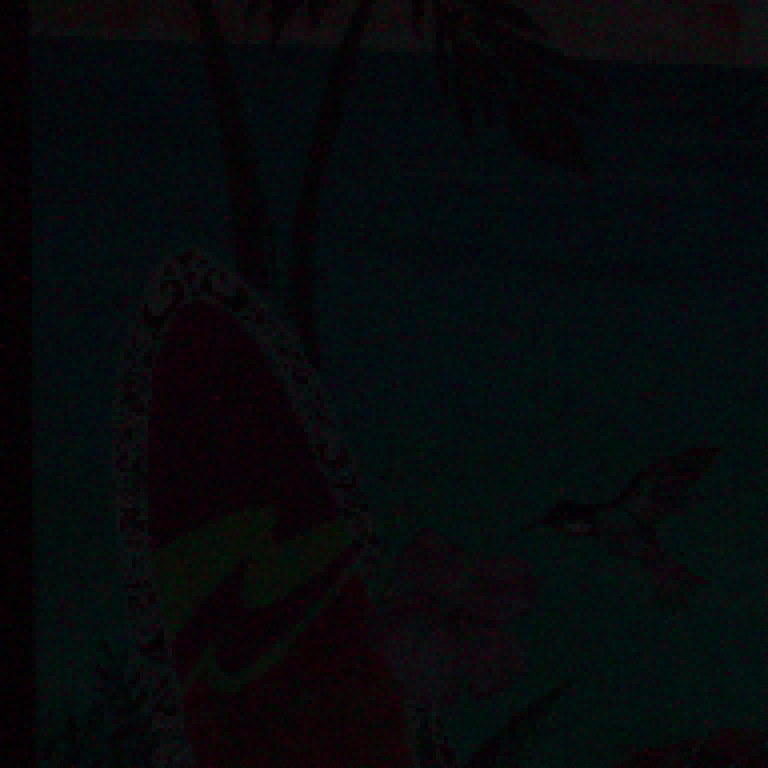}
	\\
	\hspace{-2.6mm}
	(a) -6.0 EV ($\times$1) &\hspace{-4mm} (b) -5.5 EV ($\times$1) &\hspace{-4mm} (c) -5.0 EV ($\times$1) &\hspace{-4mm} (d) -4.5 EV ($\times$1)
	\\
	\hspace{-2.6mm}
	\includegraphics[width=0.245\linewidth]{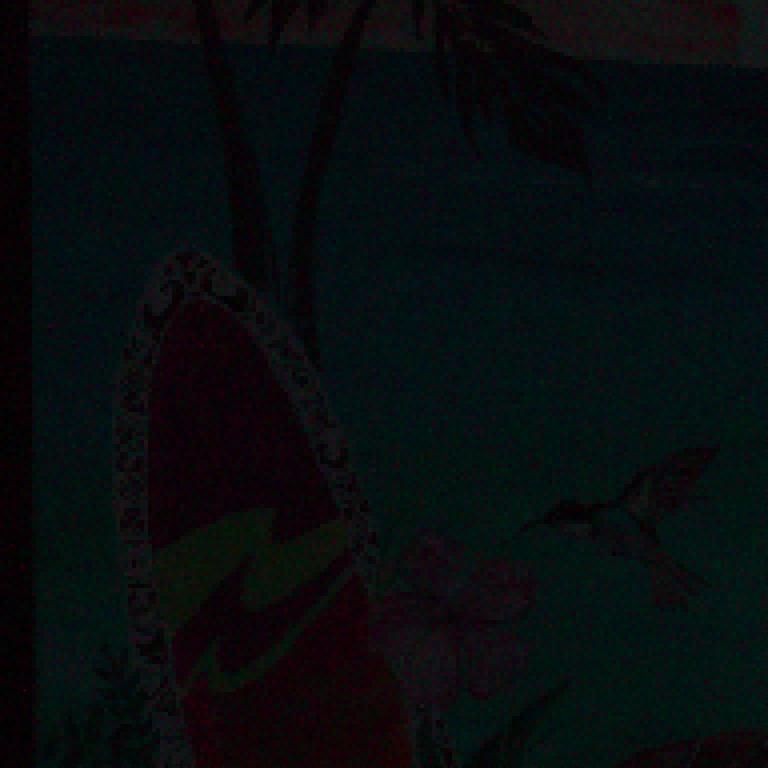} &\hspace{-4mm}
	\includegraphics[width=0.245\linewidth]{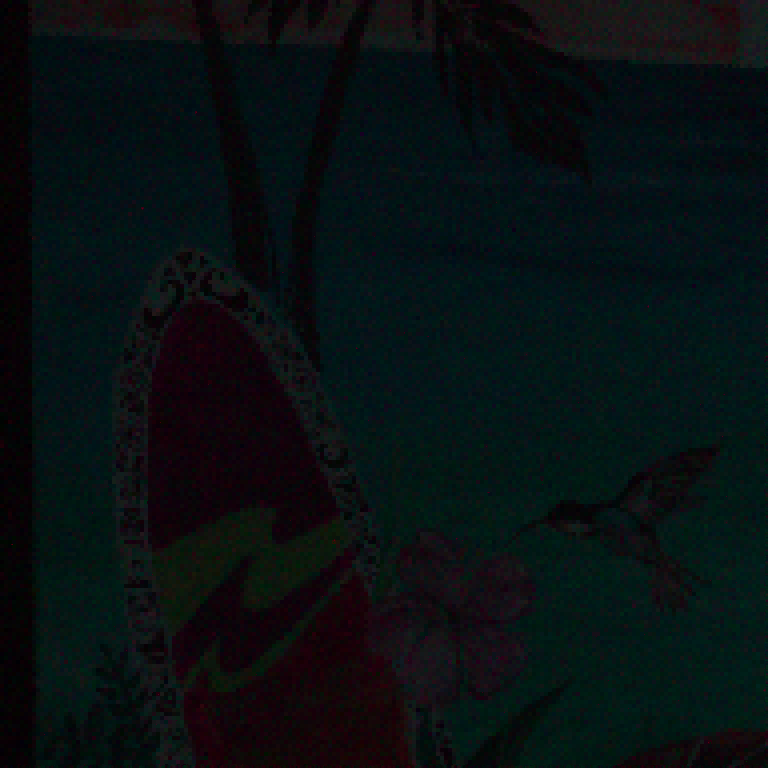} &\hspace{-4mm}
	\includegraphics[width=0.245\linewidth]{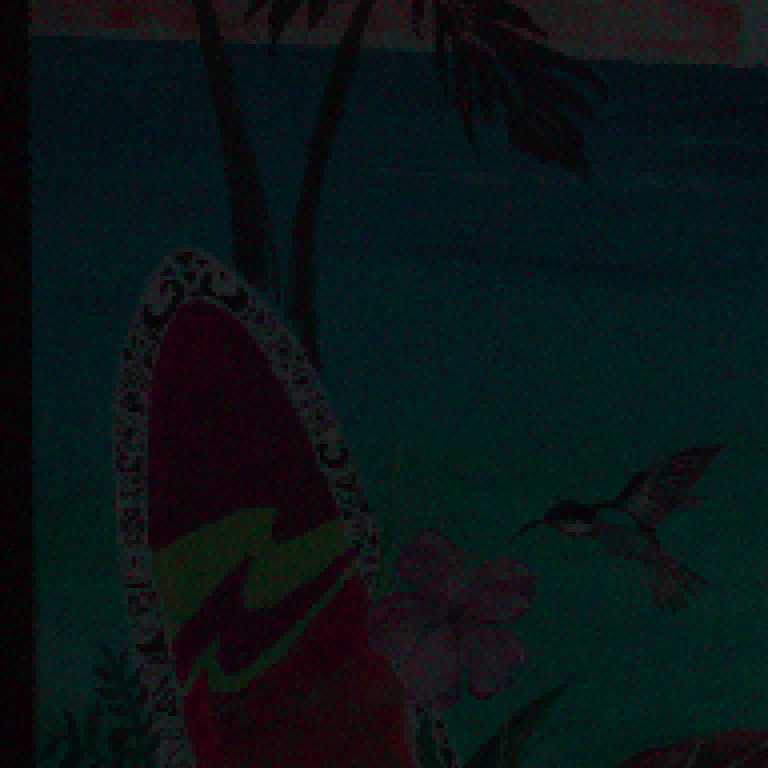} &\hspace{-4mm}
	\includegraphics[width=0.245\linewidth]{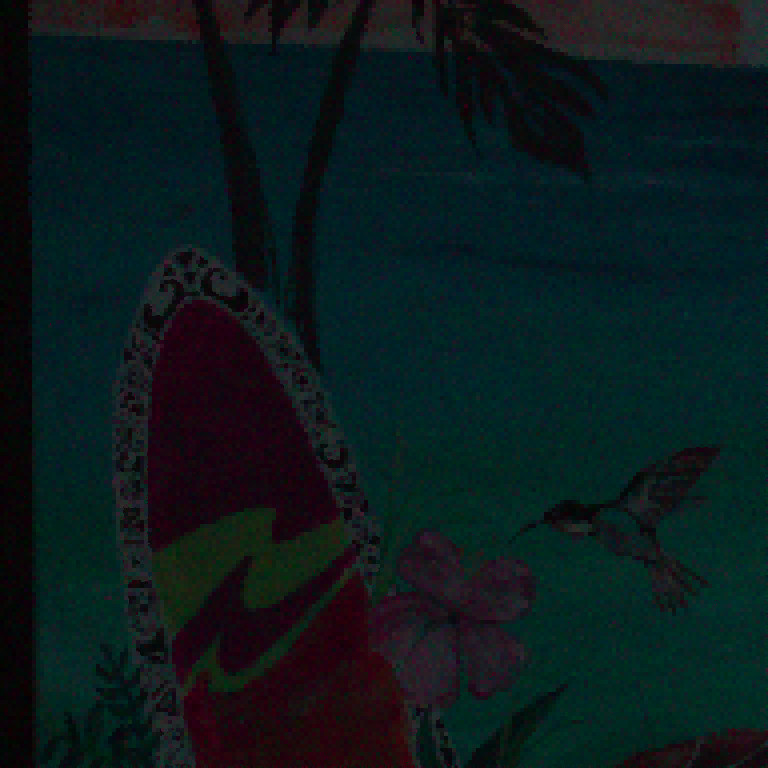}
	\\
	\hspace{-2.6mm}
	(e) -4.0 EV ($\times$1) &\hspace{-4mm} (f) -3.5 EV ($\times$1) &\hspace{-4mm} (g) -3.0 EV ($\times$1) &\hspace{-4mm} (h) -2.5 EV ($\times$1)
	\\
	\hspace{-2.6mm}
	\includegraphics[width=0.245\linewidth]{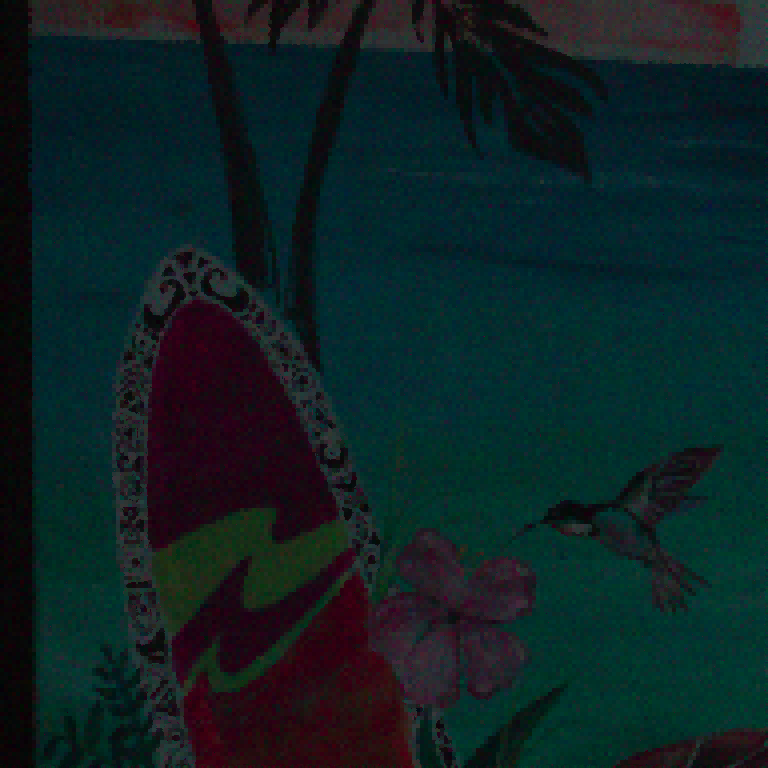} &\hspace{-4mm}
	\includegraphics[width=0.245\linewidth]{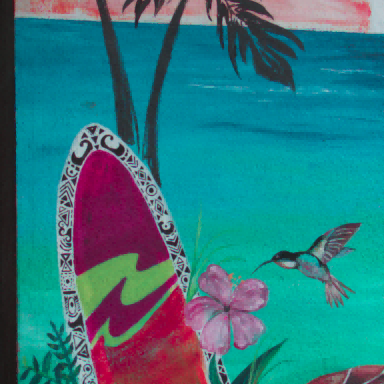} &\hspace{-4mm}
	\includegraphics[width=0.245\linewidth]{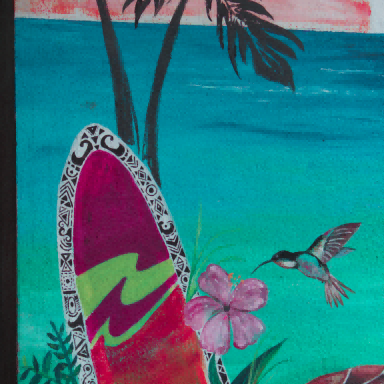} &\hspace{-4mm}
	\includegraphics[width=0.245\linewidth]{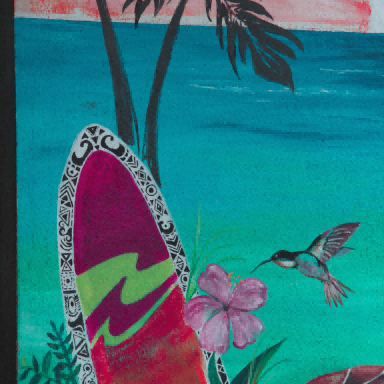}
	\\
	\hspace{-2.6mm}
	(i) -2.0 EV ($\times$1) &\hspace{-4mm} (j) 0 EV ($\times$1) &\hspace{-4mm} (k) 0 EV ($\times$2) &\hspace{-4mm} (l) 0 EV ($\times$4)
	\\
	\end{tabular}
		%\end{center}
	%\vspace{0.5mm}
	\caption{We capture the low-light images with varying under-exposure levels for each scene ranging from -6 EV to -2 EV, and fixed 0 EV for the normal-light images. We show 9 out of 27 low-light low-resolution images and corresponding 3 normal-light high-resolution images in this Figure.
	}
	\vspace{-4mm}
	\label{fig: an image sequence}
\end{figure*}

\begin{figure*}[!tp]\footnotesize
	\centering
	\vspace{0mm}
	\begin{tabular}{cccccc}
	\hspace{-2.6mm}
	\includegraphics[width=0.245\linewidth]{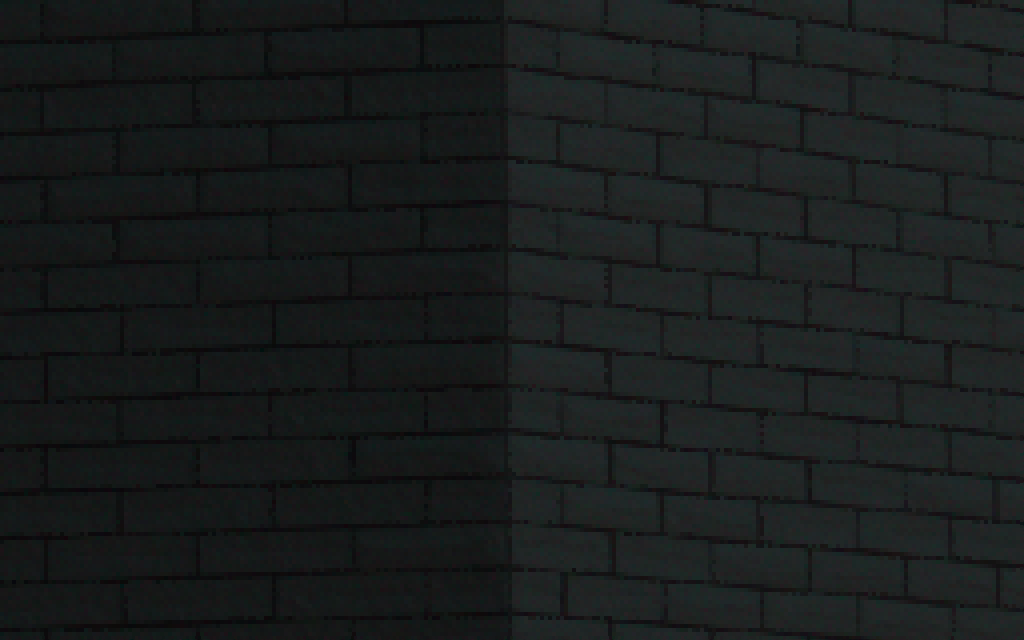} &\hspace{-4mm}
	\includegraphics[width=0.245\linewidth]{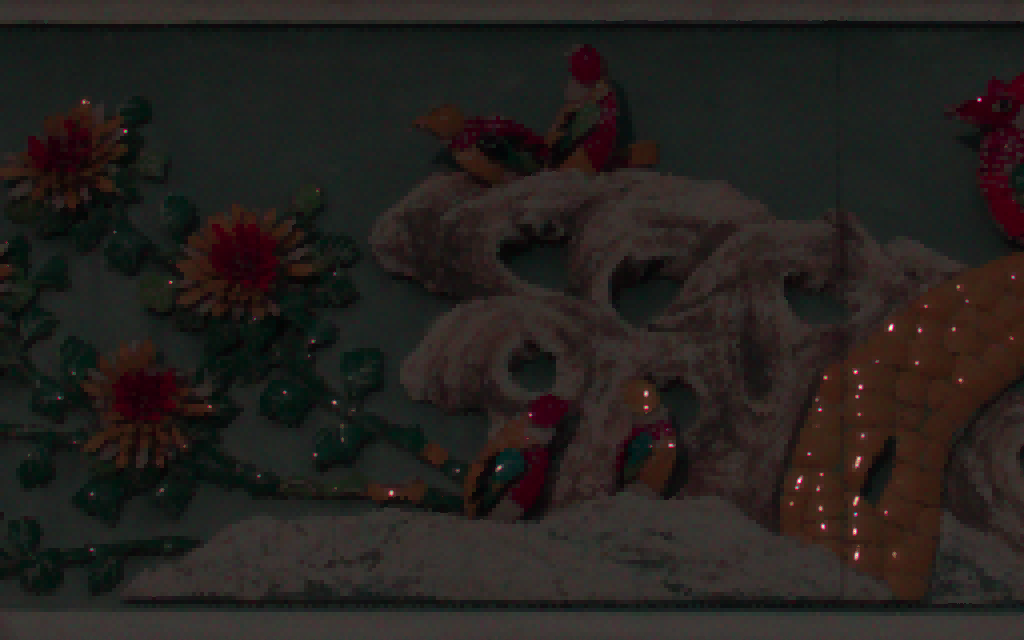} &\hspace{-4mm}
	\includegraphics[width=0.245\linewidth]{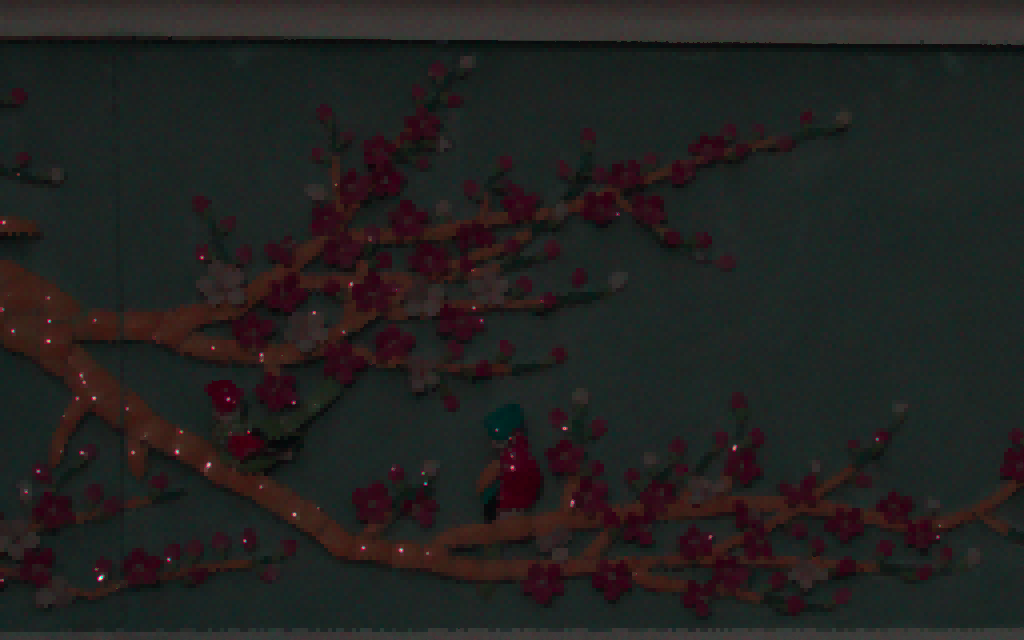} &\hspace{-4mm}
	\includegraphics[width=0.245\linewidth]{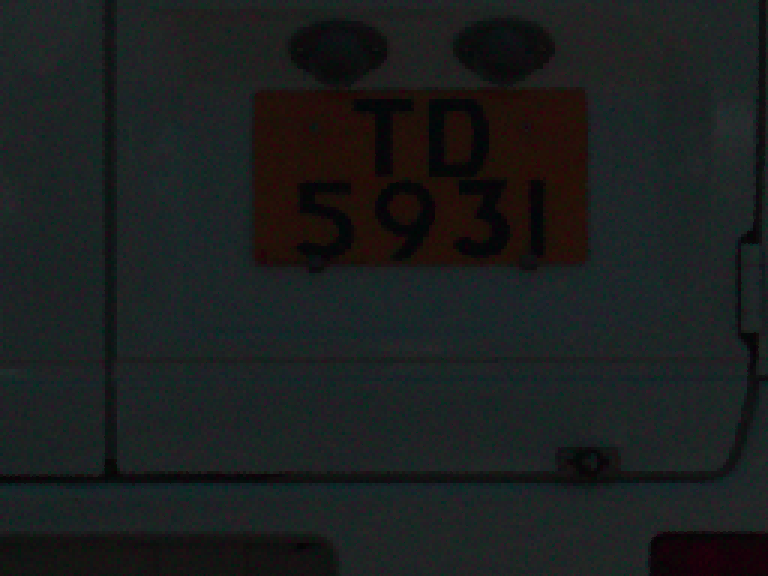}
	\\
	\hspace{-2.6mm}
	(a) ISO 800 &\hspace{-4mm} (b) ISO 1000 &\hspace{-4mm} (c) ISO 1250
 &\hspace{-4mm} (d) ISO 1600
	\\
	\hspace{-2.6mm}
	\includegraphics[width=0.245\linewidth]{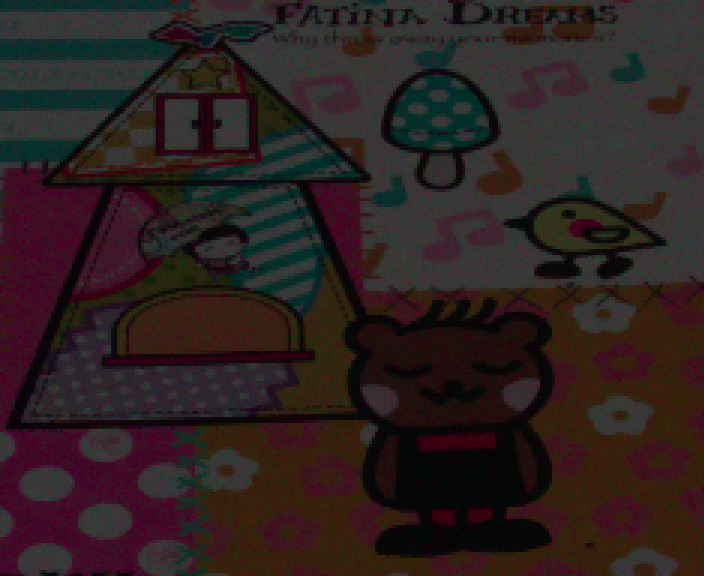} &\hspace{-4mm}
	\includegraphics[width=0.245\linewidth]{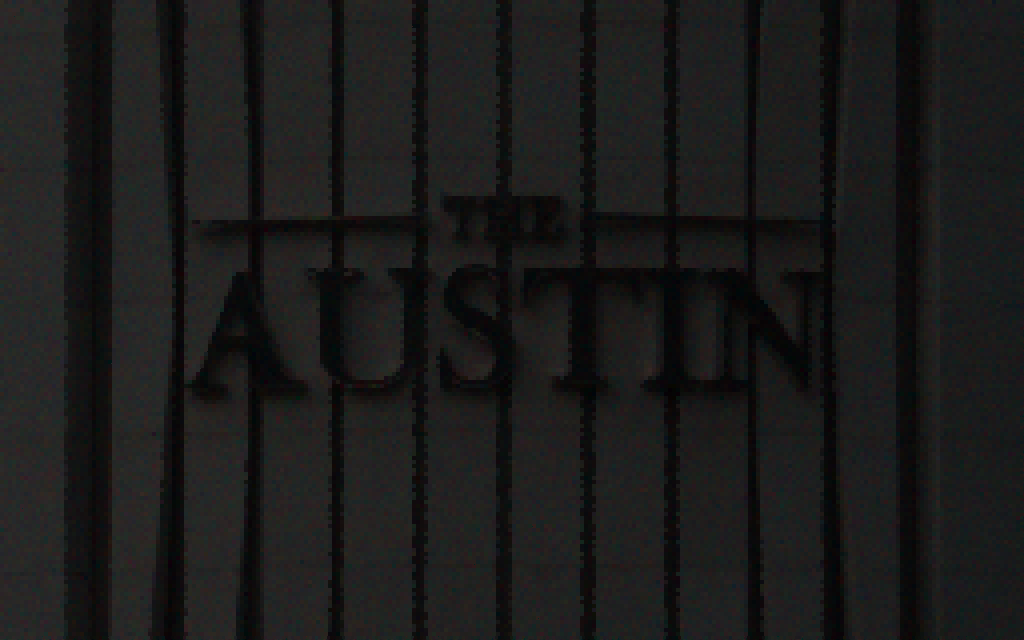} &\hspace{-4mm}
	\includegraphics[width=0.245\linewidth]{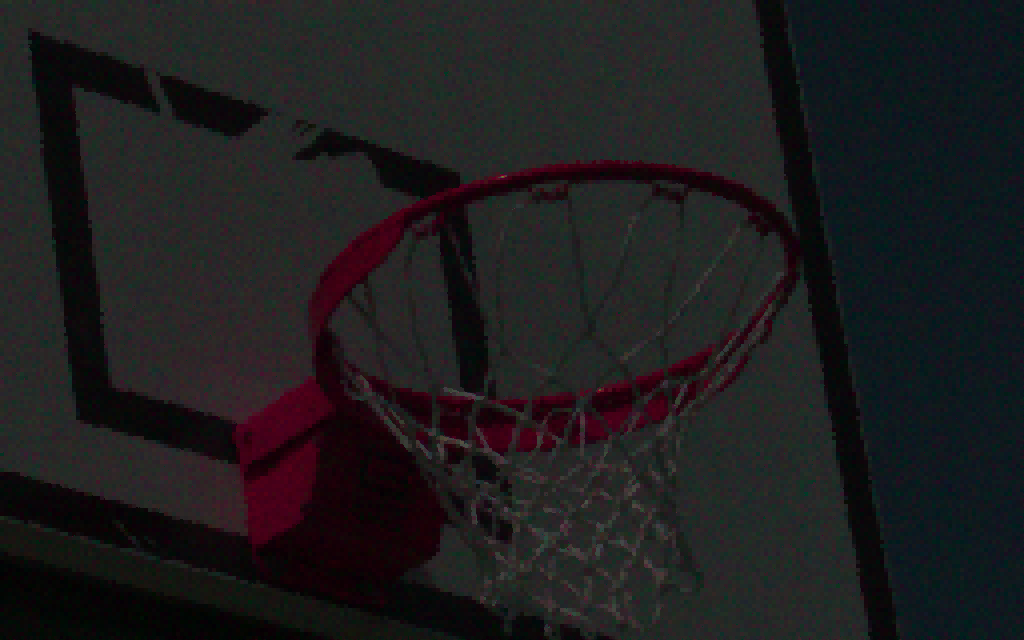} &\hspace{-4mm}
	\includegraphics[width=0.245\linewidth]{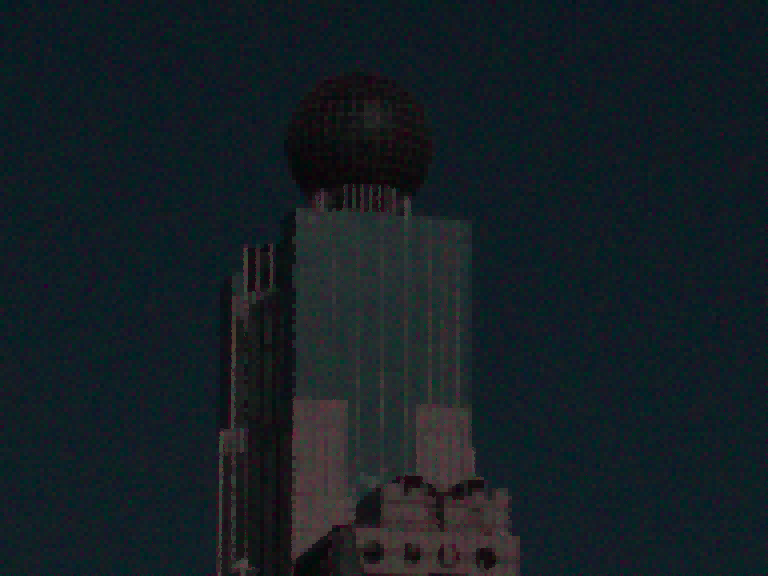}
	\\
	\hspace{-2.6mm}
	(e) ISO 2000 &\hspace{-4mm} (f) ISO 2500 &\hspace{-4mm} (g) ISO 3200 &\hspace{-4mm} (h) ISO 4000
	\\
	\hspace{-2.6mm}
	\includegraphics[width=0.245\linewidth]{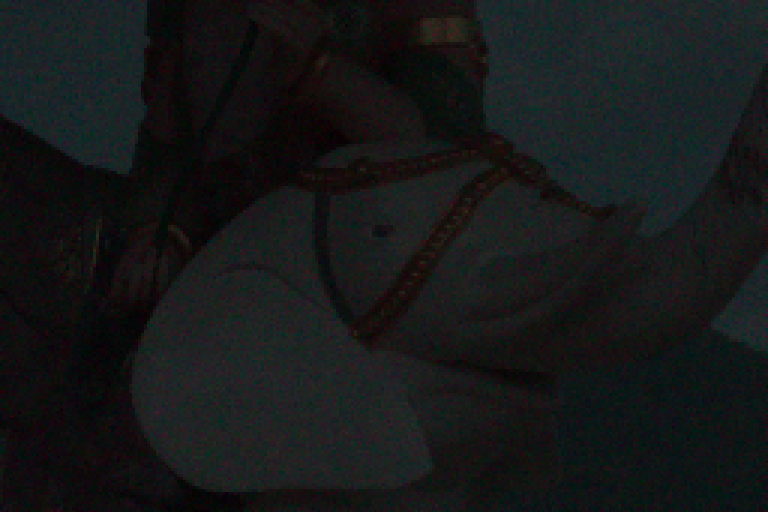} &\hspace{-4mm}
	\includegraphics[width=0.245\linewidth]{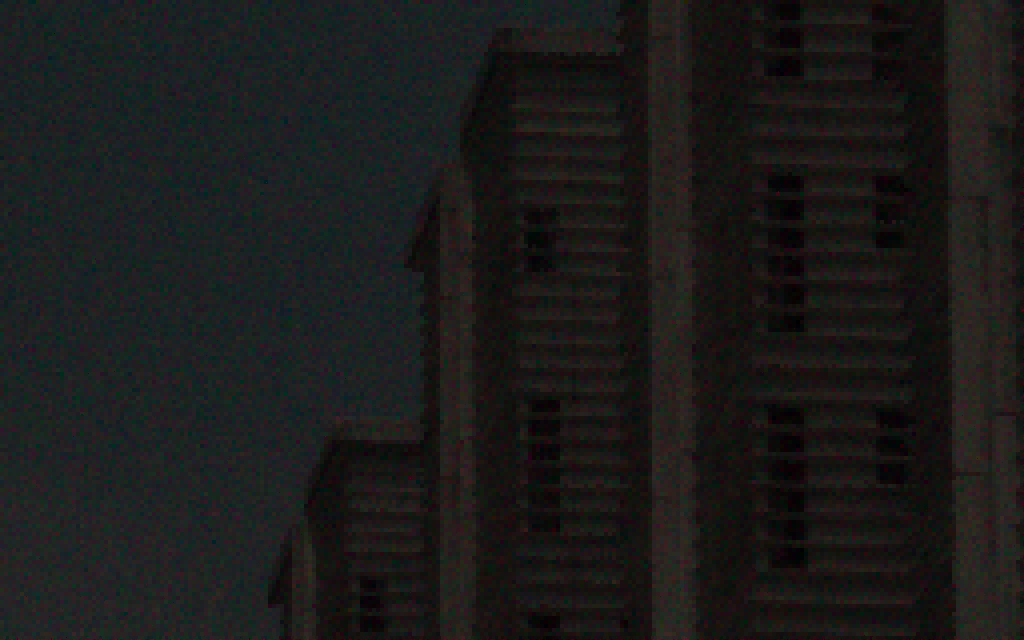} &\hspace{-4mm}
	\includegraphics[width=0.245\linewidth]{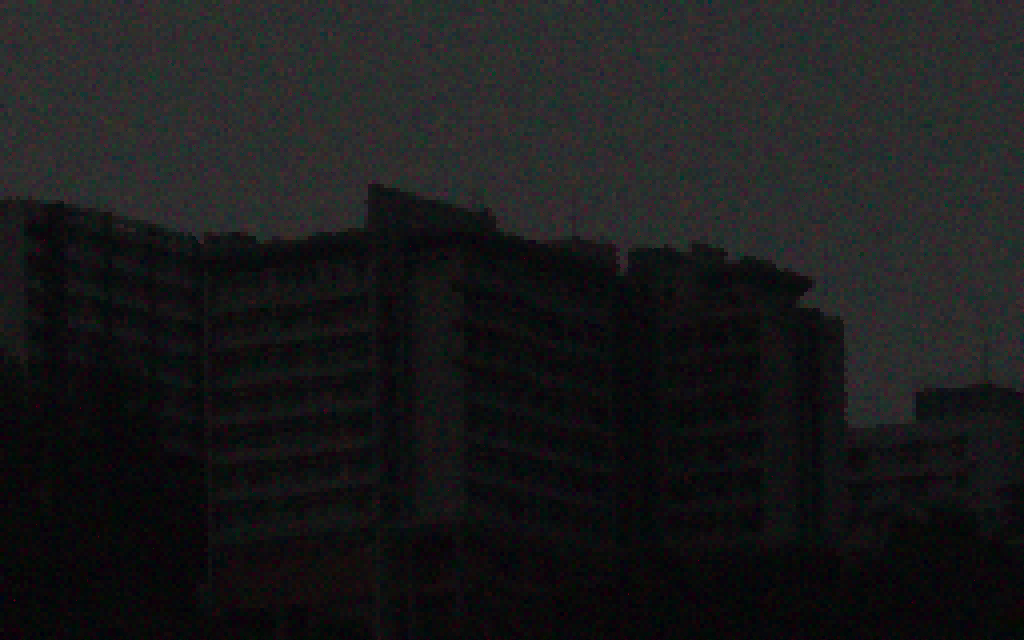} &\hspace{-4mm}
	\includegraphics[width=0.245\linewidth]{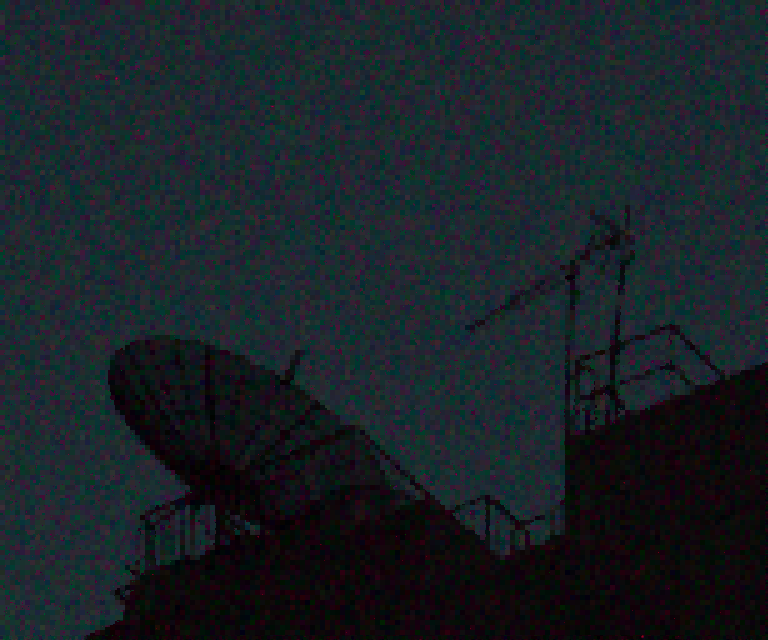}
	\\
	\hspace{-2.6mm}
	(i) ISO 5000 &\hspace{-4mm} (j) ISO 6400 &\hspace{-4mm} (k) ISO 8000 &\hspace{-4mm} (l) ISO 10000
	\\
	\hspace{-2.6mm}
	\includegraphics[width=0.245\linewidth]{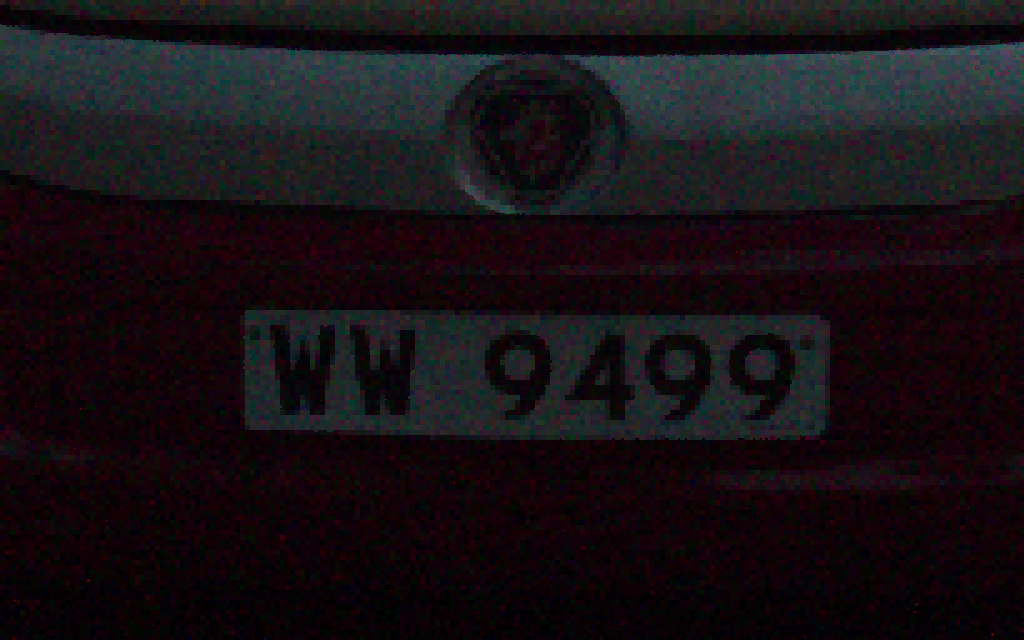} &\hspace{-4mm}
	\\
	(m) ISO 12800
	\\
	\end{tabular}
		%\end{center}
	%\vspace{0.5mm}
	\caption{We capture the low-light images with varying ISO values per scene ranging from 800 to 12800. As the ISO value increases, the real-world noises become more severe. The under-exposure level is -2 EV in this Figure.
	}
	\vspace{-4mm}
	\label{fig: different ISO}
\end{figure*}

As shown in Section 3.1 of the manuscript, we have illustrated the detailed data pipeline for the proposed IESR-RAW Dataset, which contains 500 image sequences with a total of 15000 images for 500 different indoor and outdoor scenes. 
Considering the categories of image contents and data distributions of camera ISO values within 500 scenes, we selectively split these 500 image sequences into 300, 100, and 100 sets for training, validation, and test, respectively.
Our dataset covers a variety of scenes including buildings, sculptures, wall paintings, traffic signs, license plates, etc., which do not overlap.
Figure \ref{fig: train set}, \ref{fig: val set} and \ref{fig: test set} shows example normal-light scenes for the training set, validation set, and test set, respectively.
Each image sequence contains 30 images with 3 optical zoom levels ($\times$1 and $\times$1 and $\times$4) per scene to form the low-high resolution image pair, including 27 low-light images in raw data space and 3 normal-light images in RGB color space. 
For simplicity, we only show 9 low-light low-resolution images with the under-exposure levels ranging from -6 EV to -2 EV, and the corresponding 3 normal-light high-resolution images in Figure \ref{fig: an image sequence}.
We use different ISO values ranging from 800 to 12800 for different scenes to cover a wide range of illumination variations as shown in Figure \ref{fig: different ISO}. As the ISO value increases, the real noises become more severe. Existing algorithms performs less effectively in these cases as shown in Sections 5 and 6 of the manuscript, and Section 2 in this document. 
Note that the original camera resolution is 4672 $\times$ 7008 and we preprocess each image sequence with different center crop sizes.

\begin{figure*}[!tp]
	\centering
	\vspace{0mm}
	\begin{tabular}{cccccc}
	\hspace{-2.6mm}
	\includegraphics[width=0.195\linewidth]{figures/Dataset/val/343_x4_10.png}&\hspace{-4mm}
	\includegraphics[width=0.195\linewidth]{figures/Dataset/val/89_x4_10.png} &\hspace{-4mm}
	\includegraphics[width=0.195\linewidth]{figures/Dataset/val/163_x4_10.png} &\hspace{-4mm}
	\includegraphics[width=0.195\linewidth]{figures/Dataset/val/190_x4_10.png} &\hspace{-4mm}
	\includegraphics[width=0.195\linewidth]{figures/Dataset/val/100_x4_10.png}
	\\
	\includegraphics[width=0.195\linewidth]{figures/Dataset/val/115_x4_10.png} &\hspace{-4mm}
	\includegraphics[width=0.195\linewidth]{figures/Dataset/val/123_x4_10.png} &\hspace{-4mm}
	\includegraphics[width=0.195\linewidth]{figures/Dataset/val/145_x4_10.png} &\hspace{-4mm}
	\includegraphics[width=0.195\linewidth]{figures/Dataset/val/148_x4_10.png} &\hspace{-4mm}
	\includegraphics[width=0.195\linewidth]{figures/Dataset/val/154_x4_10.png}
	\\
	\includegraphics[width=0.195\linewidth]{figures/Dataset/val/364_x4_10.png} &\hspace{-4mm}
	\includegraphics[width=0.195\linewidth]{figures/Dataset/val/384_x4_10.png} &\hspace{-4mm}
	\includegraphics[width=0.195\linewidth]{figures/Dataset/val/156_x4_10.png} &\hspace{-4mm}
	\includegraphics[width=0.195\linewidth]{figures/Dataset/val/212_x4_10.png} &\hspace{-4mm}
	\includegraphics[width=0.195\linewidth]{figures/Dataset/val/213_x4_10.png}
	\\
	\includegraphics[width=0.195\linewidth]{figures/Dataset/val/4_x4_10.png} &\hspace{-4mm}
	\includegraphics[width=0.195\linewidth]{figures/Dataset/val/99_x4_10.png} &\hspace{-4mm}
	\includegraphics[width=0.195\linewidth]{figures/Dataset/val/262_x4_10.png} &\hspace{-4mm}
	\includegraphics[width=0.195\linewidth]{figures/Dataset/val/276_x4_10.png} &\hspace{-4mm}
	\includegraphics[width=0.195\linewidth]{figures/Dataset/val/279_x4_10.png}
	\\
	\includegraphics[width=0.195\linewidth]{figures/Dataset/val/288_x4_10.png} &\hspace{-4mm}
	\includegraphics[width=0.195\linewidth]{figures/Dataset/val/289_x4_10.png} &\hspace{-4mm}
	\includegraphics[width=0.195\linewidth]{figures/Dataset/val/302_x4_10.png} &\hspace{-4mm}
	\includegraphics[width=0.195\linewidth]{figures/Dataset/val/308_x4_10.png} &\hspace{-4mm}
	\includegraphics[width=0.195\linewidth]{figures/Dataset/val/51_x4_10.png}
	\\
	\includegraphics[width=0.195\linewidth]{figures/Dataset/val/255_x4_10.png} &\hspace{-4mm}
	\includegraphics[width=0.195\linewidth]{figures/Dataset/val/257_x4_10.png} &\hspace{-4mm}
	\includegraphics[width=0.195\linewidth]{figures/Dataset/val/9_x4_10.png} &\hspace{-4mm}
	\includegraphics[width=0.195\linewidth]{figures/Dataset/val/360_x4_10.png} &\hspace{-4mm}
	\includegraphics[width=0.195\linewidth]{figures/Dataset/val/390_x4_10.png}
	\\
	\end{tabular}
		%\end{center}
	%\vspace{0.5mm}
	\caption{Example normal-light scenes in the validation set of our proposed IESR-RAW Dataset. There are 100 non-overlapping outdoor and indoor scenes for the data collection of 100 different image sequences in the validation set.
	}
	\vspace{-4mm}
	\label{fig: val set}
\end{figure*}

\section{More Experimental Results}
\label{sec: More Experimental Results}

\begin{figure*}[!tp]\small
	\centering
	\vspace{0mm}
	\begin{tabular}{cccccc}
	\hspace{-2.6mm}
	\includegraphics[width=0.245\linewidth]{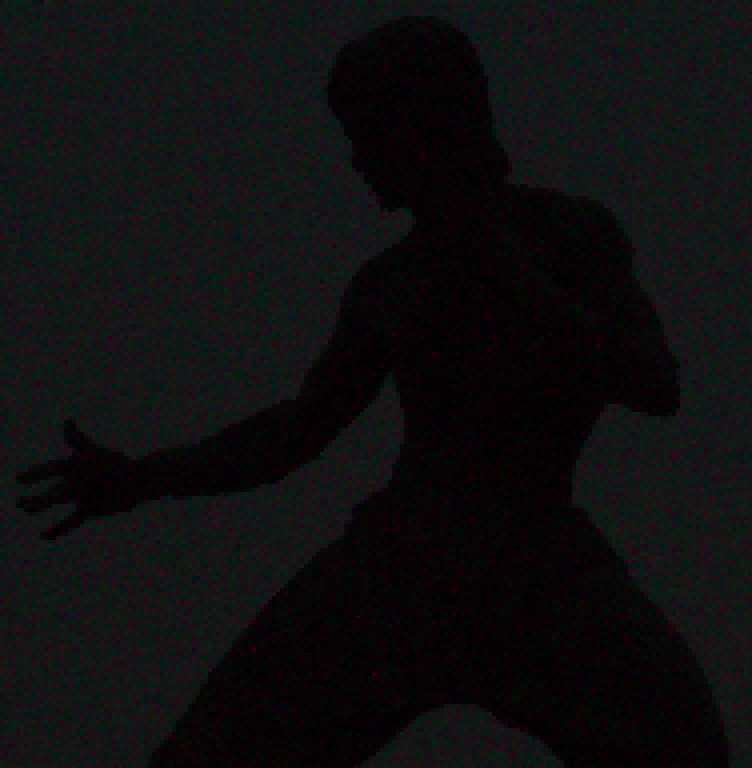} &\hspace{-4mm}
	\includegraphics[width=0.245\linewidth]{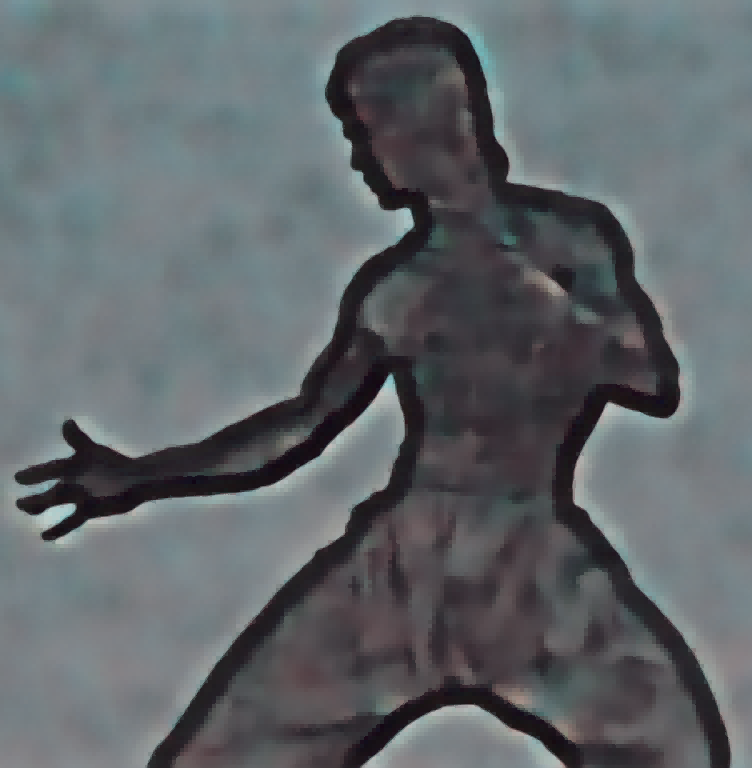} &\hspace{-4mm}
	\includegraphics[width=0.245\linewidth]{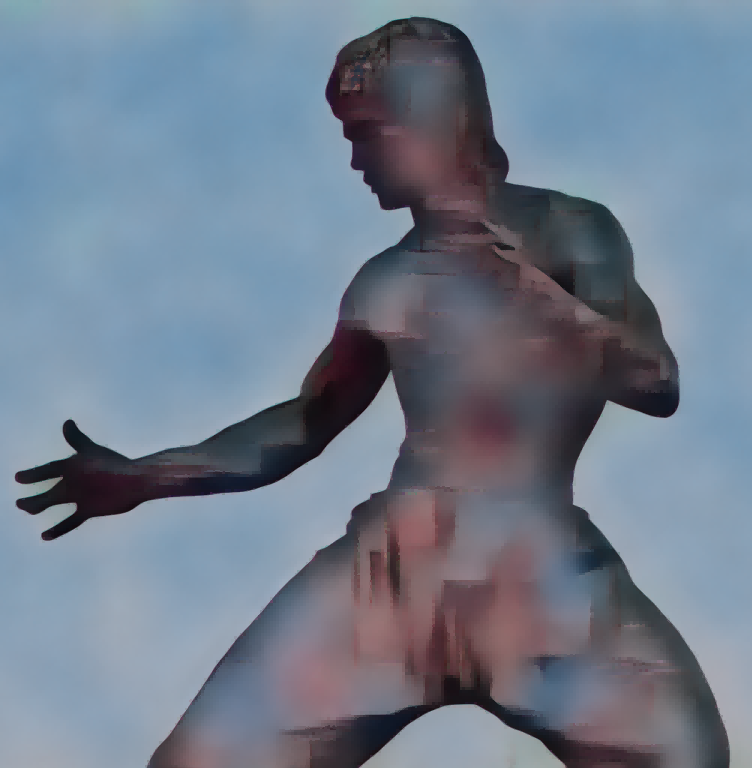} &\hspace{-4mm}
	\includegraphics[width=0.245\linewidth]{figures/sota/211/0455_mirnet.png}
	\\
	\hspace{-2.6mm}
	(a) Input image &\hspace{-4.5mm} (b) USRNet \cite{zhang2020deep} &\hspace{-4.5mm} (c) RealSR \cite{Ji_2020_CVPR_Workshops} &\hspace{-4.5mm} (d) MIRNet \cite{Zamir2022MIRNetv2}
	\\
	\hspace{-2.6mm}
	\includegraphics[width=0.245\linewidth]{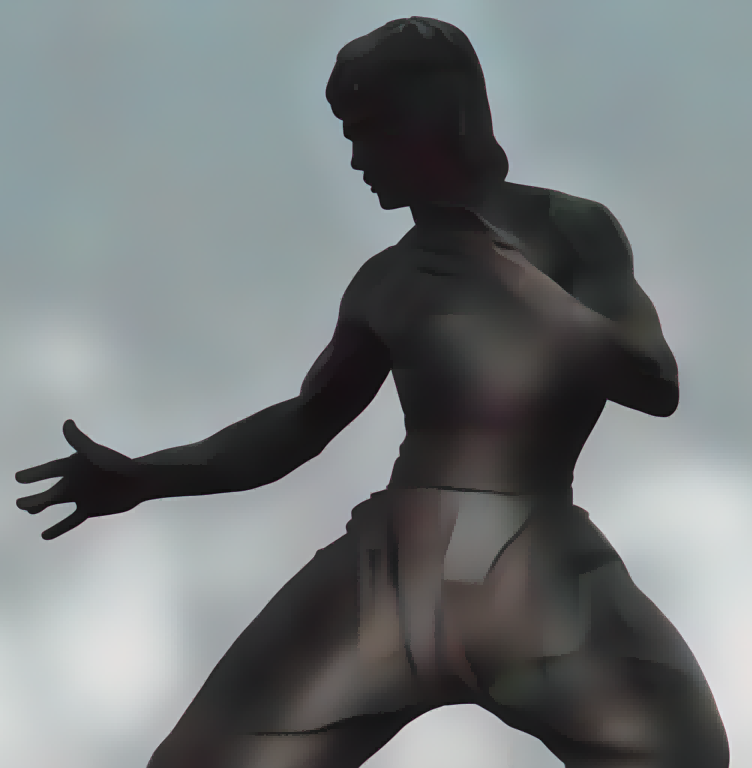} &\hspace{-4mm}
	\includegraphics[width=0.245\linewidth]{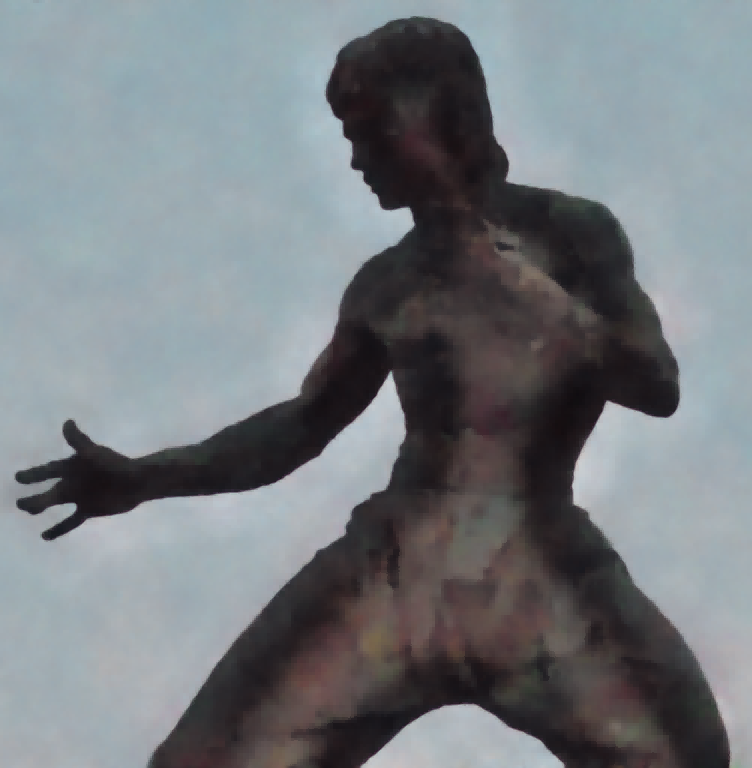} &\hspace{-4mm}
	\includegraphics[width=0.245\linewidth]{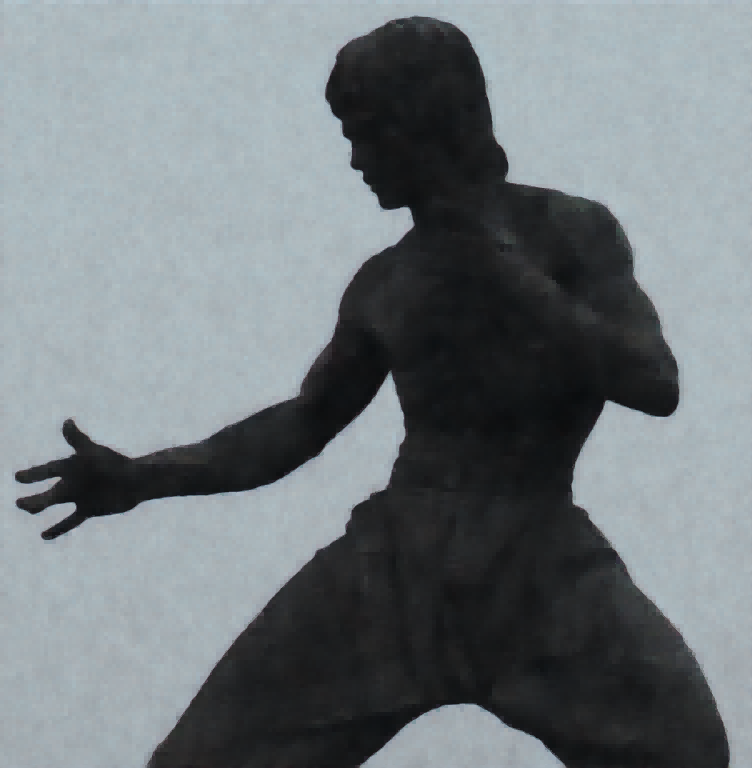} &\hspace{-4mm}
	\includegraphics[width=0.245\linewidth]{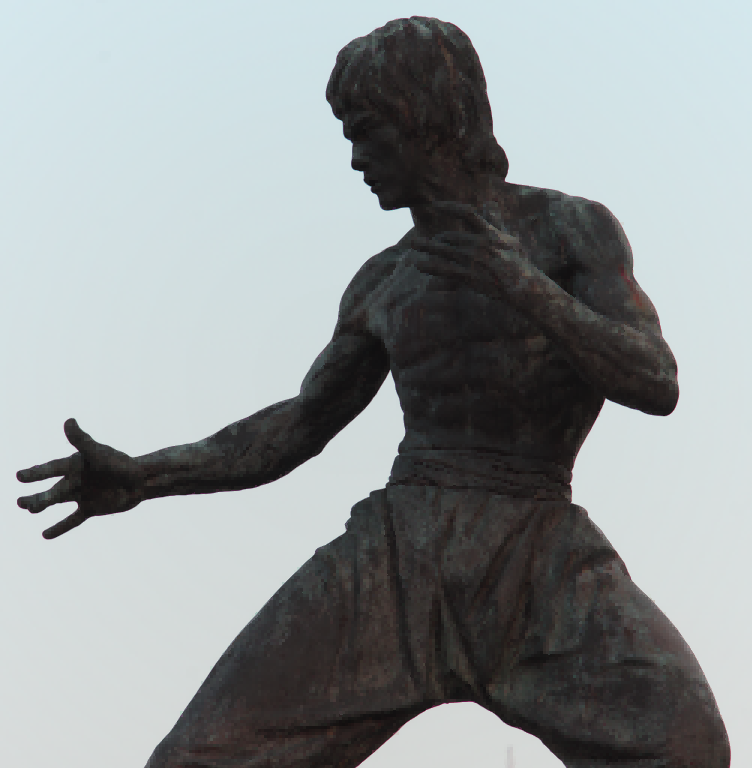}
	\\
	\hspace{-2.6mm}
	(e) Real-ESRGAN \cite{wang2021realesrgan} &\hspace{-4.5mm} (f) SwinIR \cite{liang2021swinir} &\hspace{-4.5mm} (g) Ours &\hspace{-4.5mm} (h) Ground Truth
	\\
	\end{tabular}
		%\end{center}
	%\vspace{0.5mm}
	\caption{Qualitative visual evaluations (-4 EV, ISO 5000) for the compared methods on x4 SR of the proposed IESR-RAW dataset.
	}
	\vspace{-4mm}
	\label{fig: sota-1}
\end{figure*}

\begin{figure*}[!tp]\small
	\centering
	\vspace{0mm}
	\begin{tabular}{cccccc}
	\hspace{-2.6mm}
	\includegraphics[width=0.245\linewidth]{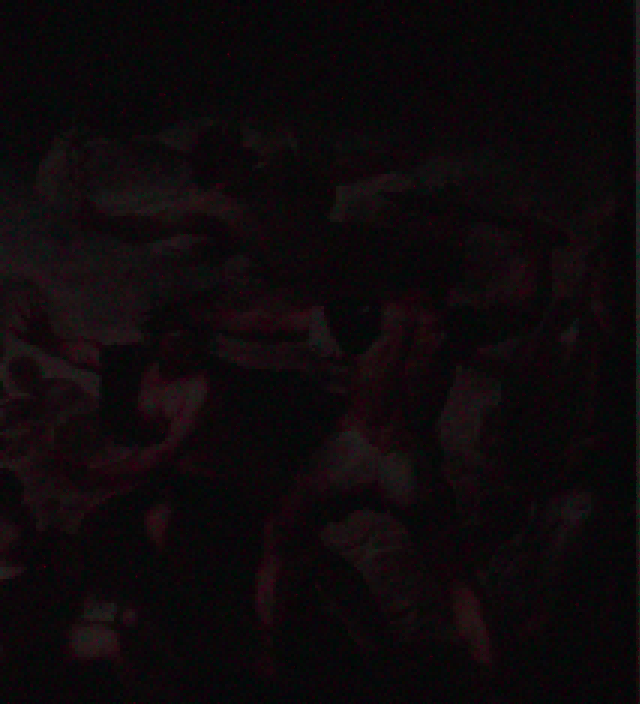} &\hspace{-4mm}
	\includegraphics[width=0.245\linewidth]{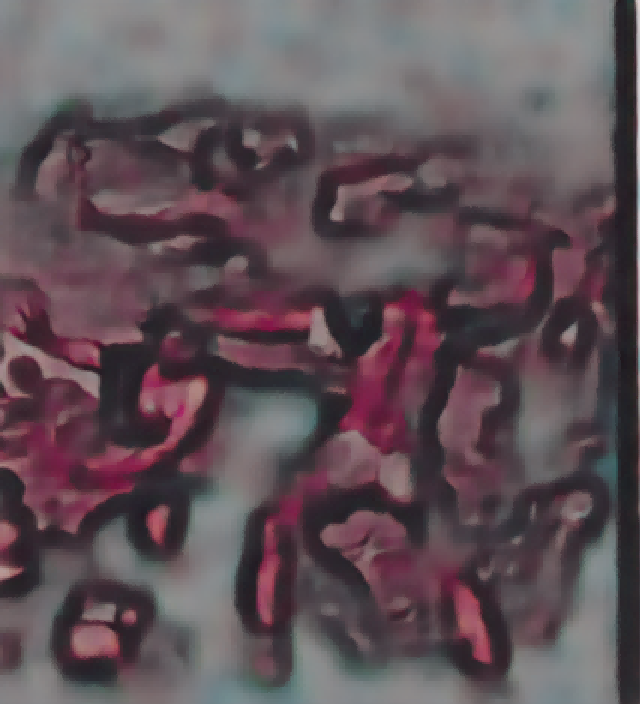} &\hspace{-4mm}
	\includegraphics[width=0.245\linewidth]{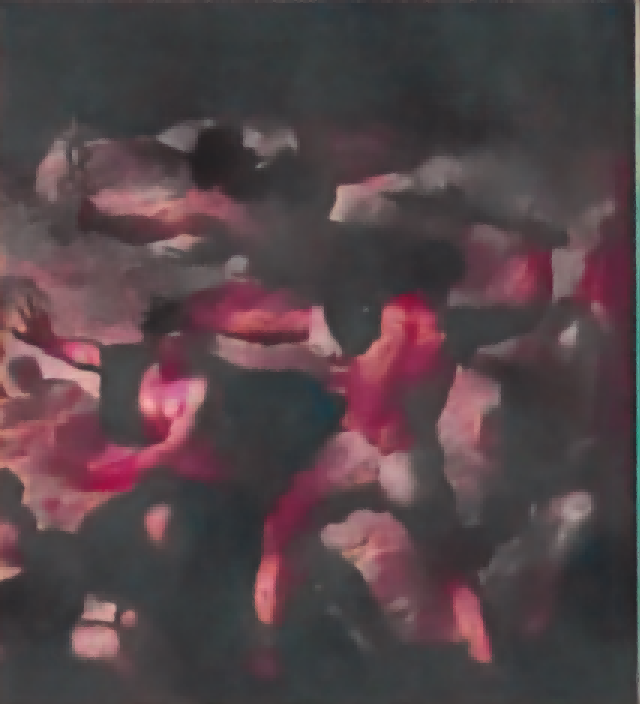} &\hspace{-4mm}
	\includegraphics[width=0.245\linewidth]{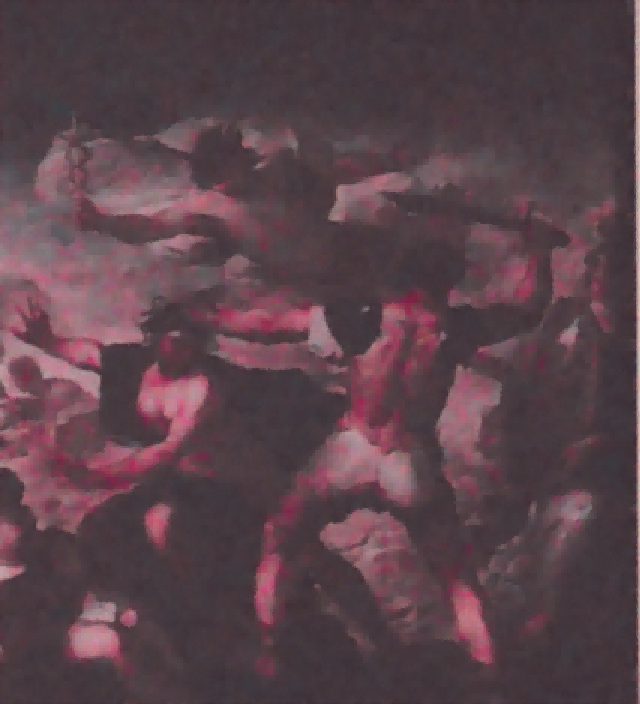}
	\\
	\hspace{-2.6mm}
	(a) Input image &\hspace{-4.5mm} (b) USRNet \cite{zhang2020deep} &\hspace{-4.5mm} (c) MIRNet \cite{Zamir2022MIRNetv2} $\rightarrow$ SwinIR \cite{liang2021swinir} &\hspace{-4.5mm} (d) MIRNet \cite{Zamir2022MIRNetv2}
	\\
	\hspace{-2.6mm}
	\includegraphics[width=0.245\linewidth]{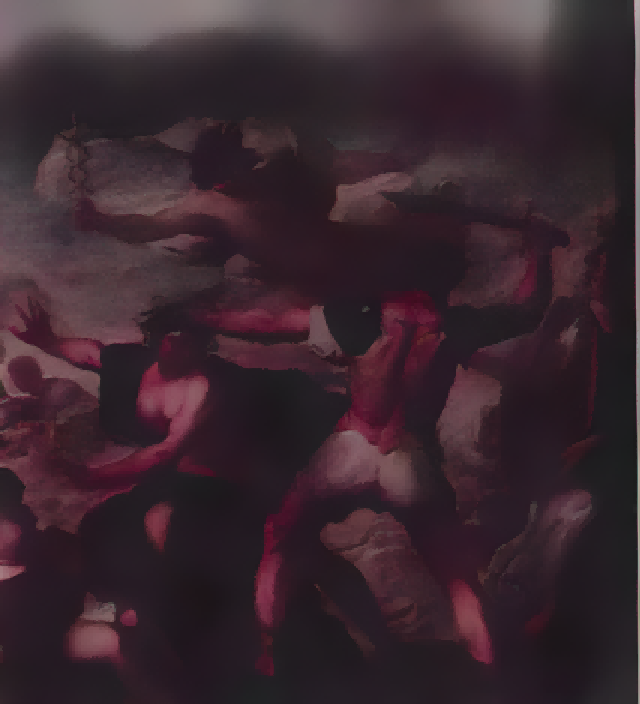} &\hspace{-4mm}
	\includegraphics[width=0.245\linewidth]{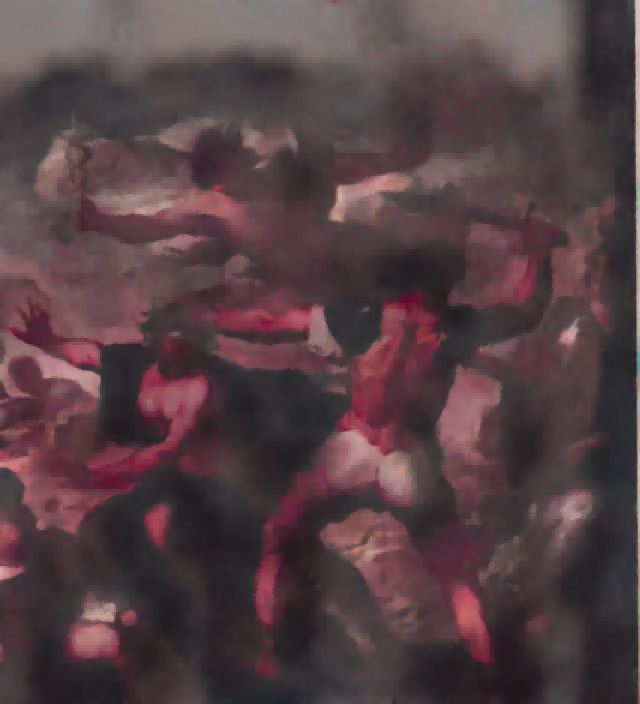} &\hspace{-4mm}
	\includegraphics[width=0.245\linewidth]{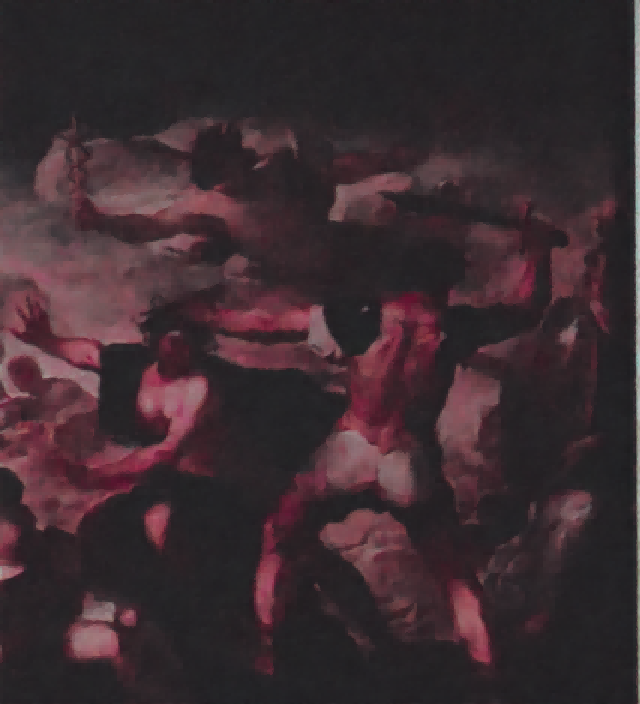} &\hspace{-4mm}
	\includegraphics[width=0.245\linewidth]{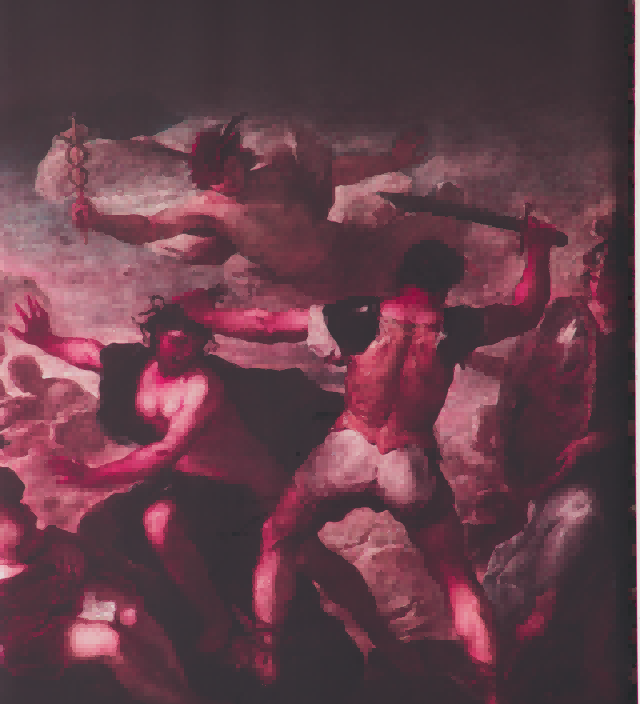}
	\\
	\hspace{-2.6mm}
	(e) Real-ESRGAN \cite{wang2021realesrgan} &\hspace{-4.5mm} (f) SwinIR \cite{liang2021swinir} &\hspace{-4.5mm} (g) Ours &\hspace{-4.5mm} (h) Ground Truth
	\\
	\end{tabular}
		%\end{center}
	%\vspace{0.5mm}
	\caption{Qualitative visual evaluations (-3 EV, ISO 4000) for the compared methods on x4 SR of the proposed IESR-RAW dataset.
	}
	\vspace{-4mm}
	\label{fig: sota-2}
\end{figure*}

As shown in Section 5.2 of the manuscript, we have compared the proposed method against the state-of-the-art algorithms on the proposed IESR-RAW dataset. To benchmark the compared methods, we have employed multiple reconstruction and perceptual metrics including PSNR, SSIM, LPIPS \cite{zhang2018perceptual}, DISTS \cite{ding2020iqa} and FID \cite{Seitzer2020FID}.
Demonstrated by both comprehensive quantitative and qualitative experimental results, we have shown that the proposed conditional diffusion model based method performs favorably against other ResNet-based methods, GAN-based methods, and Attention-based transformer methods. 
In Figures \ref{fig: sota-1} and \ref{fig: sota-2}, we show more and larger resulting images on the proposed IESR-RAW dataset, where the results of the state-of-the-art methods are obtained using the corresponding publicly available codes for fair comparisons. We fine-tune the pre-trained models of other methods the proposed IESR-RAW dataset instead of training from scratch and choose their best models for evaluation.

%%%%%%%%% REFERENCES
\clearpage
{\small
\bibliographystyle{ieee_fullname}
\bibliography{egbib}
}